\documentclass[11pt]{article}
\usepackage[utf8]{inputenc}

\usepackage{amssymb, amsmath, amsthm, graphicx, authblk, bm, bbm,  soul}
\usepackage{fullpage}
\usepackage{booktabs}
\usepackage[round]{natbib}
\usepackage{mathtools}
\usepackage{mathrsfs} 
\usepackage{stmaryrd}
\usepackage{multirow}
\usepackage{enumitem}
\usepackage[dvipsnames]{xcolor}

\usepackage{algorithm}
\usepackage{algpseudocode}
\algnewcommand\algorithmicinput{\textbf{INPUT:}}
\algnewcommand\INPUT{\item[\algorithmicinput]}
\algnewcommand\algorithmicoutput{\textbf{OUTPUT:}}
\algnewcommand\OUTPUT{\item[\algorithmicoutput]}

\usepackage{hyperref}[]
\hypersetup{
    colorlinks=true,
    linkcolor=blue,
    filecolor=magenta,      
    urlcolor=cyan,
      citecolor=blue,
    }
\usepackage{cleveref}

\newtheorem{theorem}{Theorem}
\newtheorem{lemma} {Lemma}

\newtheorem{corollary}{Corollary}

\newtheorem{remark}{Remark}

\allowdisplaybreaks

\DeclareMathOperator*{\rank}{rank}

\DeclareMathOperator*{\diag}{diag}
\DeclareMathOperator*{\tr}{tr}

\title{Optimal Bias-variance Tradeoff in Matrix and Tensor Estimation}
 
\date{\today}

\begin{document}

\author[1]{Shivam Kumar}
\author[2]{Xiaokai Luo}
\author[3]{Haotian Xu}
\author[4]{Carlos Misael Madrid Padilla}
\author[5]{Oscar Hernan Madrid Padilla}
\author[6]{Daren Wang}

\affil[1]{Booth School of Business, University of Chicago}
\affil[2]{Department of ACMS, University of Notre Dame}
\affil[3]{Department of Mathematics and Statistics, Auburn University}
\affil[4]{Department of Statistics and Data Science,  Washington University in Saint Louis}

\affil[5]{Department of Statistics, University of California, Los Angeles}
\affil[6]{Department of Mathematics, University of California, San Diego}
\maketitle

\begin{abstract}
We study matrix and tensor denoising when the underlying signal is \textbf{not} necessarily low-rank. In the tensor setting, we observe
\[
Y = X^\ast + Z \in \mathbb{R}^{p_1 \times p_2 \times p_3},
\]
where $X^\ast$ is an unknown signal tensor and $Z$ is a noise tensor. We propose a  one-step variant of the higher-order SVD (HOSVD) estimator, denoted $\widetilde X$, and    show that, uniformly over any user-specified Tucker ranks $(r_1,r_2,r_3)$, with high probability,
\[
\|\widetilde X - X^\ast\|_{\mathrm F}^2
= O\Big( \kappa^2\Big\{r_1r_2r_3 + \sum_{k=1}^3 p_k r_k\Big\} + \xi_{(r_1,r_2,r_3)}^2 \Big).
\]
Here, $\xi_{(r_1,r_2,r_3)}$ is the best achievable Tucker rank-$(r_1,r_2,r_3)$ approximation error of $X^\ast$ (bias), $\kappa^2$ quantifies the noise level, and $\kappa^2\{r_1r_2r_3+\sum_{k=1}^3 p_k r_k\}$ is the variance term scaling with the effective degrees of freedom of $\widetilde X$. This yields a rank-adaptive bias--variance tradeoff: increasing $(r_1,r_2,r_3)$ decreases the bias $\xi_{(r_1,r_2,r_3)}$ while increasing variance. In the matrix setting, we show that truncated SVD achieves an analogous bias--variance tradeoff for arbitrary signal matrices. Notably, our matrix result  requires \textbf{no}  assumptions on the signal matrix, such as finite rank or spectral gaps. Finally, we complement our upper bounds with matching information-theoretic lower bounds, showing that the resulting bias--variance tradeoff is minimax optimal up to universal constants in both the matrix and tensor settings.

\end{abstract}
  
\section{Introduction}
Low-rank matrix and tensor estimation is a cornerstone of modern machine learning. Such models capture essential structure in high-dimensional signals while offering substantial gains in computation and storage. Applications include recommender systems \citep{koren2009matrix}, topic modeling \citep{blei2003latent}, community detection \citep{abbe2018community, diakonikolas2024implicit}, and, more recently, latent variable learning \citep{anandkumar2014tensor, sherman2020estimating, zhang2023moment} and generative modeling \citep{hur2023generative, peng2023generative}. The success of these methods suggests that many real-world signals in high-dimensional ambient spaces can be well approximated by low-rank structure.

The theory of low-rank estimation is well developed in the matrix setting. Many efficient methods have been proposed for matrix completion and denoising, including nuclear-norm relaxation \citep{koltchinskii2011nuclear}, convex optimization \citep{candes2012exact}, nonconvex approaches \citep{chi2019nonconvex}, and refined SVD guarantees for highly rectangular matrices \citep{cai2018rate}. Nevertheless, most existing guarantees are tailored to exactly low-rank signals. When the signal is not exactly low-rank, accurate estimation is typically established under additional structural conditions, such as spectral gap assumptions.
 
Substantial efforts have advanced low-rank tensor estimation, including higher-order tensor singular value decompositions in deterministic settings \citep{de2000multilinear,DeLathauwer2000}, symmetric tensor decomposition in deterministic settings \citep{jin2024scalable}, and Riemannian optimization methods for low-rank Tucker tensor completion \citep{kressner2014low}. At the same time, the tensor setting poses additional challenges: many fundamental problems are computationally intractable, multiple inequivalent notions of rank (e.g., CP and Tucker) coexist, and best low-rank approximations need not be unique \citep[e.g.][]{hillar2013most}. Accordingly, most theoretical guarantees in tensor estimation \citep[e.g.][]{anandkumar2017homotopy,zhang2018tensor,han2022optimal}, like their matrix counterparts, assume that the underlying signal is exactly low-rank.
In practice, however, signals are rarely exactly low-rank: their singular values typically decay gradually rather than dropping to zero at a finite rank. This creates a gap between theory and practice, since many existing guarantees are tailored to idealized exactly low-rank models and do not directly characterize performance in realistic regimes.

We close this gap with a rank-adaptive analysis of classical spectral estimators in the approximately low-rank regime. Our contributions are as follows:
\begin{itemize}[leftmargin=*]
\item \textbf{Rank-adaptive bias--variance theory.} We establish an explicit bias--variance decomposition for the estimation error that holds uniformly over user-specified target ranks.
\item \textbf{Matrix setting.} We show that truncated SVD achieves the optimal bias--variance tradeoff without any structural assumptions on the signal matrix.

\item \textbf{Tensor setting.} We analyze a simple one-step variant of HOSVD and obtain analogous guarantees without assuming that the ground truth is exactly Tucker low-rank.

\item \textbf{Minimax optimality.} We complement our upper bounds with matching information-theoretic lower bounds, establishing minimax optimality up to universal constants.
\item \textbf{Methodology.} Our analysis combines classical linear algebra tools, including Mirsky's and Ky Fan's theorems, with modern concentration arguments. As a byproduct, we develop new matrix and tensor perturbation bounds that may be of independent interest.

\item \textbf{Numerical validation.} We validate the predicted tradeoffs through simulations and a real 3D brain MRI example.

\end{itemize}

\subsection{Notation}\label{sec:notation}
\textbf{Matrices}:
For positive integers $p, r$, let 
$
\mathbb{O}^{p \times r} = \{ V \in \mathbb{R}^{p \times r} : V^{\top} V = I_r \}
$, the set of   column-orthonormal matrices. 
Let $M \in \mathbb{R}^{p \times q}$ and suppose the  full singular value decomposition (SVD)  satisfies $M = U \Sigma V^{\top}$. Here  $s=\rank(M) $, $U \in \mathbb{O}^{p \times s}$, $V \in \mathbb{O}^{q \times s}$, and $\Sigma \in \mathbb{R}^{s \times s}$ is diagonal  with singular values
$  
\sigma_1(M) \geq \sigma_2(M) \geq \cdots \geq \sigma_{s}(M) \geq 0.
$  We   write the   smallest singular value of $M$ as   $\sigma_{\min}(M)$.
The operator norm and Frobenius norm of $M$ are defined as 
\[
\|M\| = \sigma_{1}(M), \qquad 
\|M\|_{\mathrm{F}} = \Big(\sum_{i=1}^p \sum_{j=1}^q M_{i,j}^2\Big)^{1/2}.
\]  
For $r \leq \rank(M)$, the rank-$r$ truncated SVD is 
\(
M_{(r)} = U_{(r)} \Sigma_{(r)} V_{(r)}^{\top},
\)
where $U_{(r)} \in \mathbb{O}^{p \times r}$ and $V_{(r)} \in \mathbb{O}^{q \times r}$ corresponds to the leading $r$ left and right singular vectors respectively, and $\Sigma_{(r)} = \diag\{\sigma_1(M), \ldots, \sigma_r(M)\}$. For brevity, we use $\text{SVD}_{r}(M)$ to denote the leading $r$ left singular vectors, i.e.
$$\text{SVD}_{r}(M) = U_{(r)}.$$ 
For two matrices $M_1 \in \mathbb{R}^{p_1 \times q_1}$ and $M_2 \in \mathbb{R}^{p_2 \times q_2}$, their Kronecker product is $M_1 \otimes M_2 \in \mathbb{R}^{(p_1p_2)\times(q_1q_2)}$.  
\
\\
Let $U, \widehat{U} \in \mathbb{O}^{p \times r}$ be two singular subspaces. We write the principal angles between $U$ and $\widehat{U}$ as \[
\Theta(U, \widehat{U}) = \diag\{\arccos(\sigma_1(U^{\top}\widehat{U})), \dots, \arccos(\sigma_r(U^{\top}\widehat{U}))\}.
\]
We use $\|\sin\Theta(U, \widehat{U})\|$ and $\|\sin\Theta(U, \widehat{U})\|_{\mathrm{F}}$ to measure the distance between the two singular subspaces.

\medskip
\noindent\textbf{Tensors}:
For any    tensor $B \in \mathbb{R}^{p_1 \times p_2 \times p_3}$, the Frobenius norm is
\[
\|B\|_{\mathrm{F}} = \left(\sum_{i_1=1}^{p_1} \sum_{i_2=1}^{p_2} \sum_{i_3=1}^{p_3 } B_{i_1, i_2,i_3}^2\right)^{1/2}.
\]  
  The mode-$1$ matricization $\mathcal{M}_1(B)$ is the unfolding of $B$ into a $p_1 \times p_{2}p_{3}$ matrix. The mode-$2$   and mode-$3$ matricization of $B$ are defined similarly.
The mode-$1$ product of $B$ with a matrix $M \in \mathbb{R}^{q\times p_1  }$  is defined as 
\[
(B \times_1 M) _{ j, i_2 i_3}
= \sum_{i_1=1}^{p_1} M _{i_1,j} B_{i_1, i_2,i_3} .
\]
We write the Tucker rank of $B$ as   $(r_1,r_2,r_3)$ if 
\[
\rank(\mathcal{M}_j(B)) = r_j, \text{ for } j=1,2,3.
\]

\medskip
\noindent\textbf{Random variables}: For a random variable $X \in \mathbb{R}$, we denote the sub-Gaussian and sub-Exponential norms as  
\[
\|X\|_{\psi_2} = \inf\{K > 0: \mathbb{E}\exp(X^2/K^2) \leq 2\} \quad \text{and} \quad \|X\|_{\psi_1} = \inf\{K > 0: \mathbb{E}\exp(|X|/K) \leq 2\}.
\]
We write $X \sim \mathrm{subGaussian}(0, \kappa^2)$  if $\mathbb{E}(X) = 0$ and $\|X\|_{\psi_2} \leq \kappa$.

\medskip
\noindent\textbf{Universal constants}: We use $C_1, C_2, \dots$ and $c_1, c_2, \dots$ to denote positive constants whose values may differ from place to place.

\section{Optimal bias–variance tradeoff in matrix estimation}\label{sec:approx_low-rank_mat}
We consider the  model
\begin{equation}\label{eq:approx_low-rank_matrix}
    Y = X^* + Z \in \mathbb{R}^{m \times n},
\end{equation}
where $X^*$ is an unknown population matrix with arbitrary rank, and $Z$ is a   noise  matrix. In particular, we do not assume that $X^*$ is low-rank, nor do we assume any spectral gap conditions on $X^*$.
Even when $\mathrm{rank}(X^*)$ is not small, it is common to approximate $X^*$ by a rank-$r$ estimator, because  (i) low-rank decomposition can dramatically increase the computational and memory  efficiency, see \Cref{remark:SVD computation cost} for more details; and (ii) the      approximation   bias  can be  small, so   balancing the bias and variance may lead to smaller estimation error. Let
\[
X^*=\sum_{i=1}^{\min\{m,n\}}\sigma_i(X^*)\,u_i^*{v_i^*}^{\!\top},
\qquad
X_{(r)}^*=\sum_{i=1}^{ r }\sigma_i(X^*)\,u_i^*{v_i^*}^{\!\top}.
\]
By the Eckart–Young–Mirsky theorem \citep{eckart1936approximation}, $X_{(r)}^*$ is the best rank-$r$ approximation to $X^*$ in Frobenius norm in the sense that 
$$ \| X^* -X^*_{(r)}\|_\mathrm{F} \le \| X^*- W\|_\mathrm{F} \text{ for any } W\in \mathbb R^{m\times n} \text{ with } \rank (W) \le r. $$ Suppose we observe $Y$ instead of $X^*$, and write the  SVD of $Y$ as
\begin{equation}\label{eq:Y and Y_r}
    Y=\sum_{i=1}^{\min\{m,n\}}\sigma_i(Y)\,u_i v_i^\top,
    \qquad
    Y_{(r)} =\sum_{i=1}^{r}\sigma_i(Y)\,u_i v_i^\top .
\end{equation}
That is, $Y_{(r)}$ is the rank-$r$ truncation of the SVD of $Y$. We now provide an upper bound between $Y_{(r)}$ and the unknown matrix $X^*$.

\begin{theorem}\label{thm:matrix}
Under \eqref{eq:approx_low-rank_matrix}, let $Y_{(r)}$ be defined as in \eqref{eq:Y and Y_r}. Then
\begin{equation}\label{eq:bias-var_mat}
\|Y _{(r)}-X^*\|_{\mathrm{F}}
\ \le\ (2+\sqrt{2})(\sqrt{r}\|Z\| + \xi_{(r)}),
\end{equation}
where $\xi_{(r)} =\|X_{(r)}^*-X^*\|_{\mathrm{F}} = \sqrt{ \sum_{i=r+1}^{\min\{m,n\}}\sigma_i^2(X^*)}$.
\end{theorem}

\medskip
\noindent\textbf{Assumption-free with respect to $X^*$.}
\Cref{thm:matrix} requires no structural conditions on $X^*$. In particular, $\rank(X^*)$ may be arbitrary, and no spectral gap assumptions are needed.

\medskip
\noindent\textbf{Adaptivity in \Cref{thm:matrix}.}
For any prescribed rank $r$, the rank-$r$ truncated SVD estimator $Y_{(r)}$ satisfies the guarantee in \Cref{thm:matrix}. In other words, the theorem holds uniformly over all $r\ge 1$, so it applies to the vanilla SVD at any user-specified target rank.

\medskip 
\noindent\textbf{Constant in \eqref{eq:bias-var_mat}.}
Tracing the proof of \Cref{thm:matrix} shows that one may take the explicit constant
\(C=2+\sqrt{2}\approx 3.414\) in \eqref{eq:bias-var_mat}. We did not attempt to optimize this further.

\medskip
\noindent\textbf{Bias–variance tradeoff.}
Theorem~\ref{thm:matrix} exhibits a clear bias–variance tradeoff for rank-$r$ estimation.
The bias term \(\xi_{(r)}=\|X^*-X_{(r)}^*\|_{\mathrm{F}}\) is the best possible
rank-$r$ approximation error of \(X^*\) and decreases as \(r\) increases.
  The standard deviation term, being  order \(( \sqrt  r\|Z\|)\), captures the cost of estimating
additional singular components under noise and increases with \(r\).

  \medskip

In what follows, we illustrate the applicability and optimality of \Cref{thm:matrix} in two popular matrix estimation settings.

\begin{corollary}\label{coro:subGaussian_mat}
    Suppose $Z \in \mathbb{R}^{m \times n}$ is a sub-Gaussian random matrix in the sense that for any $v \in \mathbb R^{m}$ with $\|v\|_2=1$,
    and $w \in \mathbb R^{n}$ with $\|w\|_2=1$, it holds that 
    $$  \| v^\top Z w \|_{\psi_2} \leq \kappa .$$  
Then with probability at least $1 - \exp(-(m+n))$, there exists a universal constant $C > 0$ such that
\[
\|Y_{(r)} - X^*\|^2_{\mathrm{F}} 
\le  C \big (    \kappa ^2    r (m+  n)   +\xi^2_{(r)}  \big),
\]
where $\xi_{(r)} =\|X_{(r)}^*-X^*\|_{\mathrm{F}} = \sqrt{ \sum_{i=r+1}^{\min\{m,n\}}\sigma_i^2(X^*)}$.
\end{corollary}

Next, we show that the error bound in \Cref{coro:subGaussian_mat} is minimax optimal.
 \begin{theorem}  \label{thm:lower bound  matrix}
Let $r,  m, n $ be arbitrary positive integers such that $r  \le  \min\{ m,n\}$.     For $\xi>0$,  let $ 
\mathcal{M}_{ r  } ^\xi $ be   such that  
$$ 
\mathcal{M}_{ r  } ^\xi    = \big \{B \in \mathbb{R}^{m\times n  } : \text{ there exits } A \text{ such that } \rank(A) \le r \text{ and }  
  \|B-A\|_\mathrm F \le \xi   \big \} . $$
Consider the model $Y = X^\ast + Z \in \mathbb{R}^{ m\times n}$ where the entries of $Z$ are i.i.d.\ $\mathrm{subGaussian}(0,\kappa^2)$.   Suppose in addition that 
$\xi^2  \le  (m-r) n \kappa^2 $. Then
  there exist a universal constant  $c >0$ such that 
\begin{align*}
    \inf_{\widehat X     }   \sup_{X^\ast\in \mathcal{M}_{r}^{\xi } }\ \mathbb{E}\bigl\|\widehat X - X^\ast\bigr\|_{\mathrm F}^2\ &\ge\ c \Bigl\{\kappa^2 r(  m+n)     + \xi^2\Bigr\}  .
\end{align*}
\end{theorem} 
\Cref{coro:subGaussian_mat} and \Cref{thm:lower bound matrix} together show that \eqref{eq:bias-var_mat} is minimax optimal when $Z$ is sub-Gaussian. Next, we illustrate the applicability of \Cref{thm:matrix} in covariance matrix estimation.

\begin{corollary}\label{coro:cov}
    Let $Z_1,\dots,Z_N \in \mathbb{R}^n$ be i.i.d.\ mean-zero sub-Gaussian random vectors with covariance $\Sigma$. Define the sample covariance matrix
\[
Y = \frac{1}{N}\sum_{k=1}^N Z_k Z_k^\top,
\]
and let $X^* = \mathbb{E}[Y]$ be the population covariance matrix. 
Suppose $Z_k$ is sub-Gaussian in the sense that 
\[
\| u^\top Z_k \|_{\psi_2}  \leq \kappa
\quad \text{for all } u\in \mathbb R^{n}, \ \|u\|_2=1.
\]
Then with probability at least $1 - 2\exp(-n)$, there exists a universal constant $C>0$ such that
\[
\| Y_{(r)} - X^*\|_{\mathrm{F}} 
 \le  C \Bigl(    \kappa \sqrt r \Bigl[  \sqrt{\frac{n}{N} }  + \frac{n}{N} \Bigr]  +\xi_{(r)}  \Bigr),
\]
where $\xi_{(r)} =\|X_{(r)}^*-X^*\|_{\mathrm{F}} = \sqrt{ \sum_{i=r+1}^{\min\{m,n\}}\sigma_i^2(X^*)}$.
\end{corollary}

\begin{remark}[Computational and storage savings from matrix low-rank factorization]\label{remark:SVD computation cost}
  Suppose \(Y\in\mathbb{R}^{m\times n}\) and let \(Y_{(r)} = U_{(r)}\,\Sigma_{(r)}\,V_{(r)}^\top\) be a rank-\(r\) factorization (e.g.~the truncated SVD), where
\(U_{(r)}\in\mathbb{R}^{m\times r}\), \(V_{(r)}\in\mathbb{R}^{n\times r}\) satisfy \(U_{(r)}^\top U_{(r)}=I_r\), \(V_{(r)}^\top V_{(r)}=I_r\), and \(\Sigma_{(r)}=\mathrm{diag}(\sigma_1,\dots,\sigma_r)\in\mathbb{R}^{r\times r}\).
For a dense multiplication with \(v\in\mathbb{R}^n\), computing $Yv$ costs   
 $ O(mn)
$ operations. 
In contrast, using the factorization we compute successively
\[
t  = V_{(r)}^\top v \in \mathbb{R}^r \  (\text{cost} \approx 2nr),\quad
s  = \Sigma_{(r)} t \in \mathbb{R}^r \  (\text{cost} \approx r),\quad 
w  = U_{(r)} s \in \mathbb{R}^m \  (\text{cost} \approx 2mr).
\]
Hence
 $ 
Y_{(r)} v \;=\; U_{(r)}\,\Sigma_{(r)}\,V_{(r)}^\top v  $ can be computed in \(O\big(mr + nr\big)\) operations.

The SVD factorization also reduces storage costs. Storing the full matrix \(Y\) requires \(O(mn)\) memory, whereas storing the factors \(U_{(r)} \in \mathbb{R}^{m \times r}\),   \(\Sigma_{(r)} \in \mathbb{R}^{r \times r}\), and   \(V_{(r)} \in \mathbb{R}^{n \times r}\) requires only \(O(mr + nr)\) scalars.

Therefore, both computation and storage using the truncated SVD are significantly smaller than in the full matrix case, especially when \(r \ll \min\{m,n\}\).
\end{remark}

\section{Optimal bias–variance tradeoff in tensor estimation}\label{sec:tensor}
We consider the noisy tensor model
\begin{equation}\label{eq:approx_low-rank_model_3}
    Y = X^* + Z \in \mathbb{R}^{p_1 \times p_2 \times p_3},
\end{equation}
where $X^*$ is an unknown signal tensor and $Z$ is a noise tensor. Importantly, we note that  the results in this section do not assume that $X^*$ is exactly Tucker low-rank.
Given a user-specified Tucker rank $(r_1,r_2,r_3)$, we consider the following simple variant of the HOSVD algorithm \citep{DeLathauwer2000}.

\begin{algorithm}[H]
    \caption{One-step HOSVD} 
    \label{alg:approximate_low_rank_estimatiion}
    \begin{algorithmic}
        \INPUT   Tensor \(Y\); target Tucker rank \((r_1,r_2,r_3)\).
        \For{$k = 1,2,3$}
            \State $U_k^{(0)} \gets \text{SVD}_{r_k}\!\left(\mathcal{M}_k(Y)\right) \in \mathbb{O}^{p_k\times r_k}$.
        \EndFor
        \State $U_1^{(1)} \gets \text{SVD}_{r_1}\!\Big(\mathcal{M}_1(Y)\cdot \{U_{2}^{(0)} \otimes U_{3}^{(0)}\}\Big)\in \mathbb{O}^{p_1\times r_1}$, 
        \State $U_2^{(1)} \gets \text{SVD}_{r_2}\!\Big(\mathcal{M}_2(Y)\cdot \{U_{1}^{(0)} \otimes U_{3}^{(0)}\}\Big)\in \mathbb{O}^{p_2\times r_2}$,
        \State $U_3^{(1)} \gets \text{SVD}_{r_3}\!\Big(\mathcal{M}_3(Y)\cdot \{U_{1}^{(0)} \otimes U_{2}^{(0)}\}\Big)\in \mathbb{O}^{p_3\times r_3}$.
        \OUTPUT $\widetilde{X} \gets Y \times_1 U_1^{(1)}U_1^{(1)\top}\times_2 U_2^{(1)}U_2^{(1)\top}\times_3 U_3^{(1)}U_3^{(1)\top}$.
    \end{algorithmic}
\end{algorithm}

\medskip
Let $\mathcal{T}_{(r_1,r_2,r_3)}$ denote the class of tensors in $\mathbb{R}^{p_1\times p_2\times p_3}$ with Tucker rank at most $(r_1,r_2,r_3)$, i.e.
\begin{equation}\label{eq:tucker}
    \mathcal{T}_{(r_1,r_2,r_3)}=\big\{A \in \mathbb{R}^{p_1\times p_2\times p_3}:\ \rank(\mathcal{M}_k(A)) \le r_k,\ k=1,2,3\big\}.    
\end{equation}
For any tensor $X^* \in \mathbb{R}^{p_1\times p_2 \times p_3}$, let $\xi_{(r_1,r_2,r_3)}$ denote the best
Tucker rank-$(r_1,r_2,r_3)$ approximation error:
\[
\xi_{(r_1,r_2,r_3)} \;=\; \inf_{A \in \mathcal{T}_{(r_1,r_2,r_3)}} \|A - X^*\|_{\mathrm{F}}.
\]

\begin{theorem}\label{thm:main}
Suppose $Y$ follows \eqref{eq:approx_low-rank_model_3} with 
 $Z_{\mu_1,\mu_2,\mu_3} \overset{\mathrm{i.i.d.}}{\sim} \mathrm{subGaussian}(0,\kappa^2)$.   Let $\widetilde{X}$ be the output of \Cref{alg:approximate_low_rank_estimatiion} with target Tucker rank $(r_1,r_2,r_3)$. Define $r_{\max}=\max_{k} r_k$ and $p_{\min}=\min_{k} p_k$. Assume the singular gaps satisfy
\begin{equation}\label{eq:SNR}
\big\{\sigma_{r_k}(\mathcal{M}_k(X^*)) - \sigma_{r_k+1}(\mathcal{M}_k(X^*))\big\}^2 
\;\ge\; C_{\mathrm{gap}}\kappa^2\Big(\sqrt{p_1p_2p_3\, r_{\max}} + \sum_{k=1}^3 p_k r_{\max}\Big),
\end{equation}
for 
{all $k = 1,2,3$ and for} a sufficiently large constant $C_{\mathrm{gap}}>0$. Then, with probability at least $1-C_1p_1p_2p_3\exp(-C_2 p_{\min})$,
\begin{equation}\label{eq:tensor error bound}
\|\widetilde{X} - X^*\|_{\mathrm{F}} 
\;\le\; C_3\Big\{\sqrt{\kappa^2\big(\textstyle\sum_{k=1}^3 p_k r_k + r_1r_2r_3\big)} + \xi_{(r_1,r_2,r_3)}\Big\}.
\end{equation}
\end{theorem}

\medskip
\noindent\textbf{Adaptivity in \Cref{thm:main}.}
For any user specified  target Tucker rank \((r_1,r_2,r_3)\), the estimator \(\widetilde{X}\) satisfies \eqref{eq:tensor error bound}. In other words, the result holds uniformly over all target Tucker   ranks.

\medskip
\noindent\textbf{Bias–variance tradeoff.}
The error bound \eqref{eq:tensor error bound} exhibits a clear bias–variance decomposition for Tucker rank-\((r_1,r_2,r_3)\) estimators.
The bias term \(\xi_{(r_1,r_2,r_3)}\) is the best achievable approximation error of \(X^*\) at the chosen ranks and decreases as the \(r_k\) increase.
Therefore, as the ranks grow, variance increases while bias shrinks. Balancing these two leads to improved accuracy.

\medskip
\noindent\textbf{Proof techniques.} For the bias, we leverage classical linear algebra tools, such as  Mirsky’s and Ky Fan’s theorems, to characterize the optimal approximation error appearing in \eqref{eq:tensor error bound}. To handle the variance of the estimator $\widetilde X$, we adopt existing techniques from \cite{zhang2018tensor}.

\begin{remark}[Assumptions in \Cref{thm:main}]
  In \Cref{thm:main}, we assume that the entries of the noise tensor $Z$ are i.i.d.~sub-Gaussian. This is a commonly seen condition in the tensor literature  \citep[e.g.][]{zhang2018tensor,han2022optimal}. 

 The signal-to-noise ratio (SNR) condition   \eqref{eq:SNR} in \Cref{thm:main} is also commonly seen in literature. In fact, suppose the ground truth $X^*$ is exactly Tucker low-rank with Tucker rank $(r_1,r_2,r_3)$. Then $\sigma_{r_k+1}(\mathcal M_k(X^*)) =0 $, and   \eqref{eq:SNR} reduces to the SNR condition used in \cite{zhang2018tensor}  when $r_{\max}$ is a bounded constant. 
\end{remark}

We  present a relative Frobenius error bound as a   corollary of \Cref{thm:main}.

\begin{corollary}[Relative estimation error]\label{corollary:relative error}
Suppose the conditions of \Cref{thm:main} hold and $r_k \le \sqrt{p_{\min}}$ for $k=1,2,3$. Then, with probability at least $1-C_1p_1p_2p_3\exp(-C_2 p_{\min})$,
\begin{align}\label{eq:relative error}
\frac{\|\widetilde{X} - X^*\|_{\mathrm{F}}}{\|X^*\|_{\mathrm{F}}}
\;\le\; C_3\Bigg\{
\sqrt{\kappa^2 \sum_{k=1}^3 \frac{p_k}{\sigma^2_{r_k}\!\big(\mathcal M_k(X^*)\big)}}
\;+\;
\frac{\xi_{(r_1,r_2,r_3)}}{\|X^*\|_{\mathrm{F}}}
\Bigg\}.
\end{align}
\end{corollary}

We next show that the rate in \eqref{eq:tensor error bound} is minimax optimal by proving a matching lower bound.

\begin{theorem}[Minimax lower bound] \label{thm:lower}
Let $r_1, r_2,r_3, p_1, p_2, p_3 $ be arbitrary integers such that $r_i \le \min\{ p_i, \prod_{j\not = i} p_j\} $ for all $i\in \{1, 2,3\} $. Let $\mathcal{T}_{(r_1,r_2,r_3)}$ be the class of Tucker low-rank tensor as in \eqref{eq:tucker} and for $\xi>0$,  let 
$$ 
\mathcal{T}_{(r_1,r_2,r_3)}^{\xi }= \big \{B \in \mathbb{R}^{p_1\times p_2\times p_3 } : \inf_{A \in \mathcal{T}_{(r_1,r_2,r_3)} }\|B-A\|_\mathrm F \le \xi   \big\} .
$$
Consider the model $Y = X^\ast + Z \in \mathbb{R}^{p_1\times p_2\times p_3}$ where the entries of $Z$ are i.i.d.\ $\mathrm{subGaussian}(0,\kappa^2)$. For any ranks $(r_1,r_2,r_3)$ and any $0\le \xi\le \kappa^2 p_1p_2p_3$, there exist a universal constant  $c >0$ such that 
\begin{align}
\inf_{\widehat X    } \ \sup_{X^\ast\in \mathcal{T}_{(r_1,r_2,r_3)}^{\xi } }
\ \mathbb{E}\bigl\|\widehat X - X^\ast\bigr\|_{\mathrm F}^2
\ &\ge\ c \Bigl\{\kappa^2\big(\textstyle\sum_{k=1}^3 p_k r_k + r_1r_2r_3\big)   +\xi^2\Bigr\}  .\label{eq:minimax_expectation_lb}  
\end{align}
  
\end{theorem}
  
\medskip
\noindent\textbf{Tightness.}
The upper bound in \Cref{thm:main} and   the matching  lower bound in \Cref{thm:lower} together show  that the error bound  in \eqref{eq:tensor error bound} is minimax optimal up to universal constants  over the approximately Tucker low-rank class $\mathcal{T}^{\xi}_{(r_1,r_2,r_3)}$.

\medskip
\noindent
\textbf{Advantages of low-rank representations.}
Even when $X^*$ has full Tucker rank $(p_1,p_2,p_3)$, it is common to approximate $X^*$ by a Tucker rank-$(r_1,r_2,r_3)$ estimator. This is because (i) the approximation bias can be small when $X^*$ is well approximated by a low-rank tensor, so balancing bias and variance may reduce estimation error; and (ii) low-rank structure can yield substantial gains in computation and storage, as demonstrated in the following remark.

\begin{remark}[Computational and storage savings from tensor low-rank estimation]  \label{remark:Tensor computation cost}
  Suppose \(Y\in\mathbb{R}^{p_1 \times p_2 \times p_3}\) and let \[Y_{(r_1, r_2, r_3)} = S \times_1 U_1 \times_2 U_2 \times_3 U_3 \] be a Tucker rank-\((r_1, r_2, r_3)\) approximation of $Y$, where
\(U_k \in\mathbb{R}^{p_k\times r_k}\) satisfying \(U_k^\top U_k = I_{r_k}\) for $k = 1, 2, 3$, and \(S \in \mathbb{R}^{r_1 \times r_2 \times r_3}\) is the core tensor.

For dense multiplications with \(v_k \in\mathbb{R}^{p_k}\) along each mode $k$, computing $Y \times_1 v_1 \times_2 v_2 \times_3 v_3$ costs   
 $ O(p_1p_2p_3)
$ operations. 
In contrast, using the Tucker decomposition, we compute $Y_{(r_1, r_2, r_3)} \times_1 v_1 \times_2 v_2 \times_3 v_3$ successively as 
\[
t_k  = U_k^\top v_k \in \mathbb{R}^{r_k} \  (\text{cost } \approx 2\sum_{k = 1}^3p_kr_k),\quad
S^{\prime}  = S \times_1 t_1 \in \mathbb{R}^{r_2\times r_3} \  (\text{cost } \approx 2r_1r_2r_3),
\]
\[
s = t_2^{\top} S^{\prime} \in \mathbb{R}^{r_3} \ (\text{cost } \approx 2r_2r_3), \quad w = s^{\top} t_3 \in \mathbb{R}  \ (\text{cost } \approx r_3).
\]
Thus, vector multiplication with Tucker factors reduce the computational cost from $O(p_1p_2p_3)$ to  $O(\sum_{k = 1}^3p_kr_k + r_1r_2r_3)$.
The storage requirements of Tucker decomposition also drop significantly. Storing the full tensor $Y$ requires \(O(p_1p_2p_3)\) memory, while the Tucker representation requires only
\(\sum_{k = 1}^3p_kr_k + r_1r_2r_3\) scalar  storage. 

Therefore, both computation and storage of Tucker decomposition scale with $O(\sum_{k = 1}^3p_kr_k + r_1r_2r_3)$, which is substantially smaller than the dense case when $r_k \ll p_k$.
\end{remark}

\section{Numerical experiments}\label{sec:experiments}
We numerically demonstrate the bias-variance tradeoff developed in our theory. We begin by describing a practical approach of selecting the rank parameters used in both SVD and one-step HOSVD (\Cref{alg:approximate_low_rank_estimatiion}). 

\begin{remark}[Practical choice of target ranks] 
    There are several well-known approaches for selecting the target Tucker ranks. To select the rank parameters in a data-driven manner, we recommend following the discussion in \cite{ozdemir2017multienergy} and adopting an adaptive thresholding scheme. More precisely, one selects \(r_k\) such that \(\sigma_{r_k}^2(\mathcal{M}_k(Y)) \ge \tau\), where \(\tau > 0\) is prescribed threshold. A simple and effective choice is \(\tau = \alpha \|Y\|_{\mathrm F}^2\) with \(\alpha = 0.01\). Under this rule, all singular values of \(\mathcal{M}_k(Y)\) up to the largest index \(j\) satisfying \(\sigma_j^2(\mathcal{M}_k(Y)) \ge \tau\) are retained, and the remainder are discarded. 
\end{remark}
In many applied settings, however, selecting the single {best} rank $r$ is inherently difficult in finite samples, particularly with unknown latent structure. Our analysis instead provides guarantees for singular-value decompositions in both the matrix  and tensor  regimes across a broad range of $r$. Consequently, rather than relying on a specific rank-selection heuristic, we empirically demonstrate  that both truncated SVD and one-step HOSVD   are robust provided that the chosen ranks are neither too small (underfitting) nor too large (overfitting). This behavior is consistent with the bias–variance tradeoff formalized in \Cref{thm:matrix,thm:main}. Moreover, in practice the choice of $r$ is often guided by considerations beyond minimizing mean-squared error, such as visualization, interpretability, or performance on downstream tasks. This perspective aligns with the widespread and long standing use of low-rank methods across diverse application domains, biology and genomics \citep{price2006principal}, economics \citep{filmer2001estimating}, and computer vision \citep{turk1991eigenfaces}.

We evaluate the bias-variance tradeoff    on (i) a real 3D brain-MRI dataset with controlled additive noise \citep[T1-weighted volumes]{IXI}, and (ii) simulated data in both matrix and third–order tensor settings.  

\subsection{Real data: 3D brain MRI with controlled noise}\label{subsec:real}

We evaluate on {five} T1-weighted brain MRI volumes drawn from the IXI dataset \citep{IXI}. The selected {volumes are third-order tensors} of shape $256\times 256\times 150$. 

For each volume, we consider the original tensor and two wavelet-smoothed tensors obtained using orthonormal wavelet families Daubechies-6 (\texttt{db6}) and Symlet-8 (\texttt{sym8}). Wavelet smoothing is performed using separable 3D multilevel decompositions (\texttt{wavedecn}) followed by subband-wise soft-thresholding with BayesShrink \citep{DonohoJohnstoneBiometrika1994, ChangYuVetterliTIP2000}. This approach is well supported in the MRI literature as an effective smoothing strategy for MRI volumes \citep{NowakTIP1999, WoodMRI1999}.

\medskip
\noindent\textbf{Noise model and ranks.}
For each noise level {$\lambda \in \{0.2, 0.3, 0.4, 0.5, 0.6, 0.7\}$} and each volume $X^\ast$,  we generate  
\[
Y \;=\; X^* + Z,\qquad
Z_{ijk}\stackrel{\text{i.i.d.}}{\sim}\mathcal N(0,\kappa^2),\quad
\kappa \;=\; \lambda\,\frac{\|X^*\|_{\mathrm F}}{\sqrt{p_1 p_2 p_3}}\,,
\]
which keeps the per-entry SNR comparable across images and variants. We compute one-step HOSVD  reconstructions (\Cref{alg:approximate_low_rank_estimatiion}) $\widetilde X$ at Tucker rank $(r,r,r)$ for
{\(
r \in \{35,50,60,65,70,80,90,95\}.
\)}

\begin{figure}[h!]
\centering
\begin{minipage}[t]{0.48\textwidth}
  \centering
  \includegraphics[width=\linewidth]{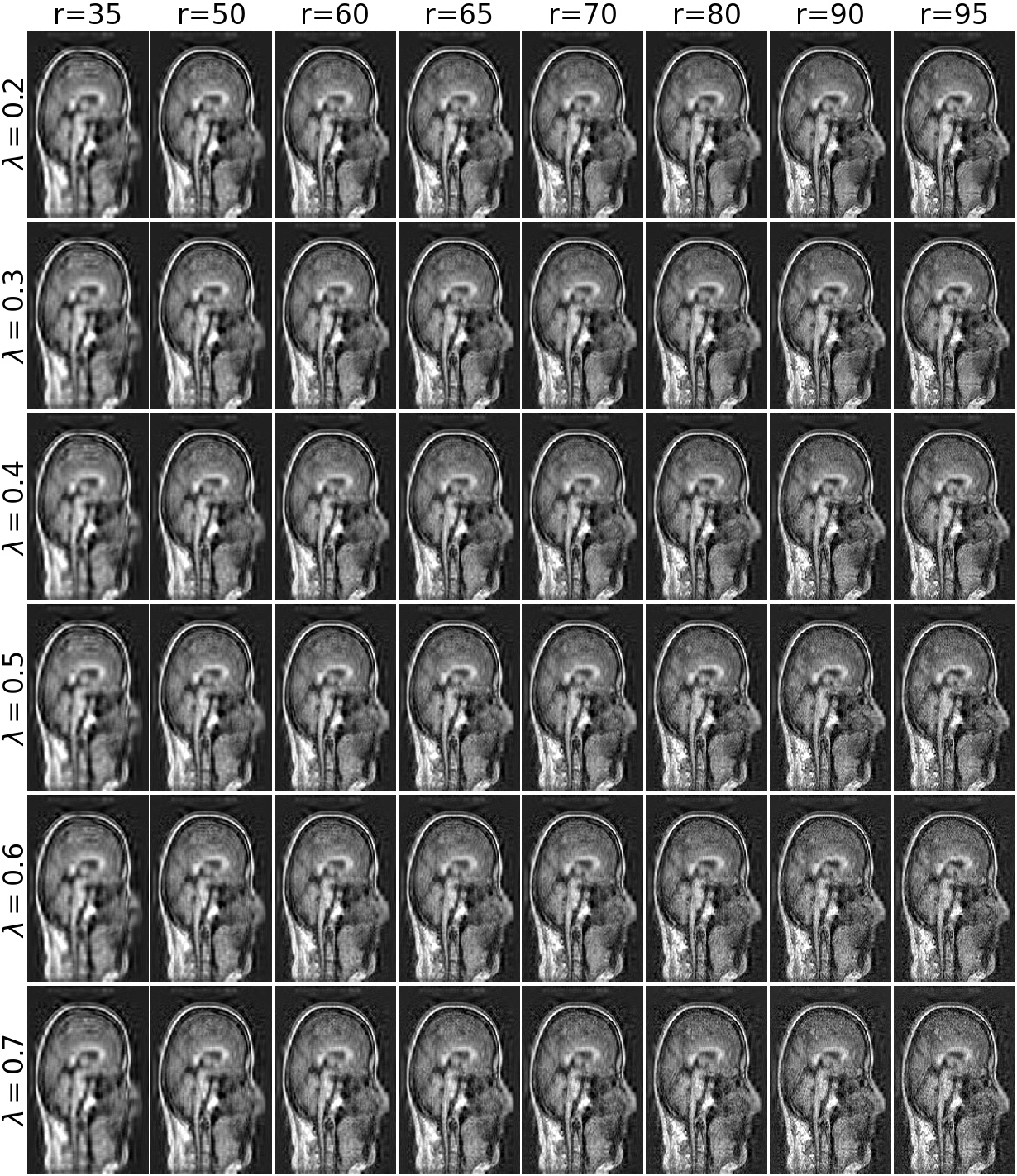}

\end{minipage}\hfill
\begin{minipage}[t]{0.48\textwidth}
  \centering
  \includegraphics[width=\linewidth]{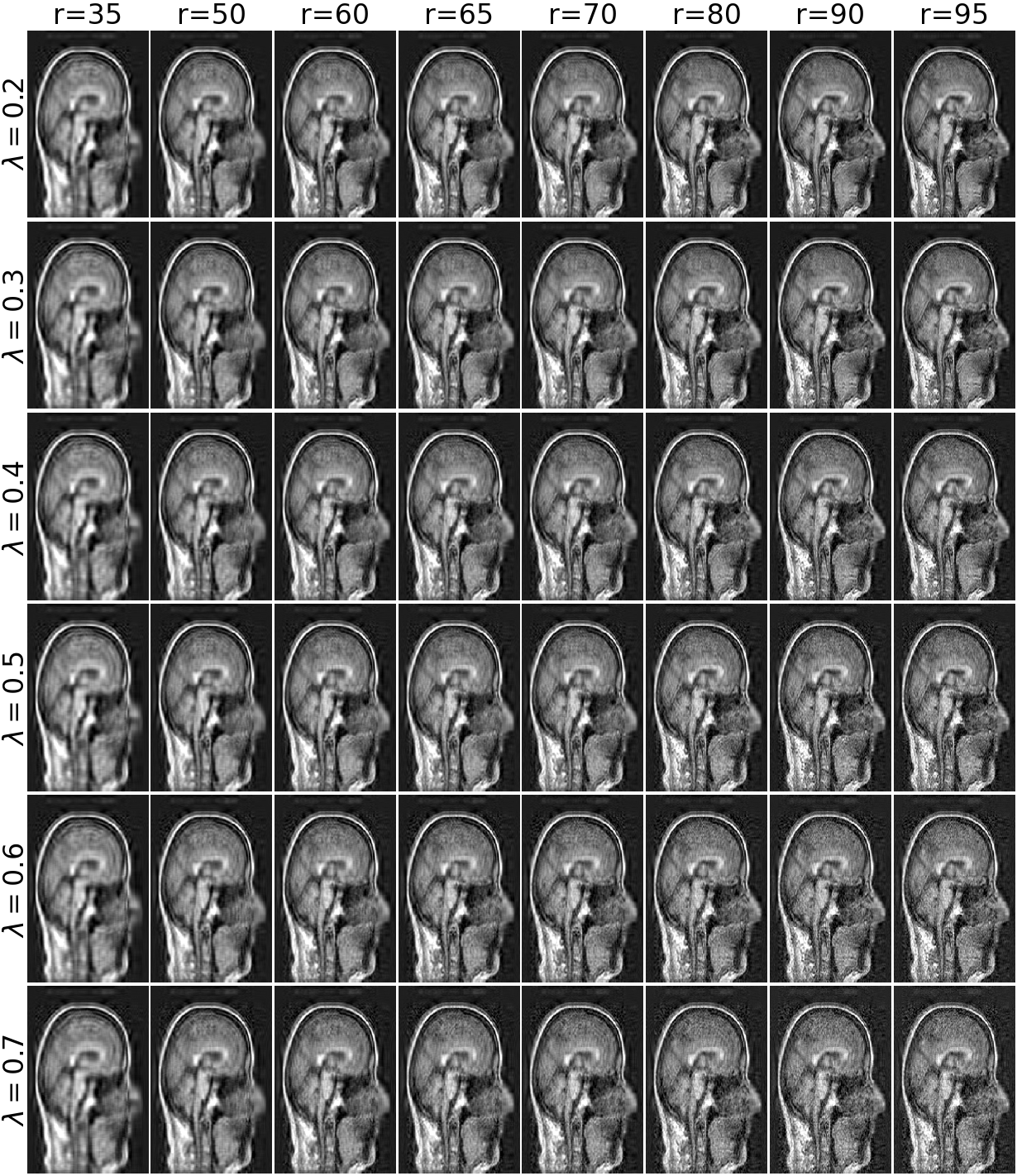}
\end{minipage}
\caption{Mid-sagittal MRI slices reconstructed with \Cref{alg:approximate_low_rank_estimatiion} at Tucker ranks \((r,r,r)\) across noise levels \(\lambda\) for a selected volume. Inputs smoothed with \texttt{db6} (left) and \texttt{sym8} (right) wavelets.}
\label{fig:mean_vs_brain}
\end{figure}

\medskip
\noindent\textbf{Findings.}
{Figure~\ref{fig:mean_vs_brain} displays mid-sagittal slices from reconstructions of a selected volume, comparing results obtained from inputs smoothed with \texttt{db6} and \texttt{sym8} wavelets. At intermediate ranks, the images sharpen and cortical detail appears while background fluctuations remain suppressed. We provide selected quantitative results in \Cref{tab:MRI}, which show this expected bias-variance tradeoff progression.}

\begin{table}[h!]
\centering
\caption{Mean and standard deviation (SD) of the relative reconstruction error with $5$ MRI volumnes. Results are shown across varying scaling parameters $\lambda$ and ranks $r$. 
\label{tab:MRI}}
\small
{
\begin{tabular}{ccccc}
\toprule
$\lambda$ & $r$ & original & db6 & sym8 \\
\midrule
\multirow{4}{*}{0.4}
 & 50 & 0.1753 (0.0330) & 0.1747 (0.0537) & 0.1681 (0.0466) \\
 & 65 & 0.1553 (0.0268) & 0.1561 (0.0435) & 0.1501 (0.0373) \\
 & 80 & 0.1490 (0.0183) & 0.1499 (0.0291) & 0.1451 (0.0245) \\
 & 95 & 0.1550 (0.0102) & 0.1558 (0.0159) & 0.1525 (0.0128) \\
\midrule
\multirow{4}{*}{0.5}
 & 50 & 0.1797 (0.0322) & 0.1793 (0.0521) & 0.1728 (0.0451) \\
 & 65 & 0.1655 (0.0248) & 0.1668 (0.0400) & 0.1610 (0.0341) \\
 & 80 & 0.1677 (0.0154) & 0.1691 (0.0244) & 0.1647 (0.0201) \\
 & 95 & 0.1840 (0.0073) & 0.1847 (0.0116) & 0.1819 (0.0089) \\
\midrule
\multirow{4}{*}{0.6}
 & 50 & 0.1851 (0.0311) & 0.1850 (0.0503) & 0.1787 (0.0434) \\
 & 65 & 0.1776 (0.0224) & 0.1793 (0.0362) & 0.1738 (0.0307) \\
 & 80 & 0.1890 (0.0124) & 0.1904 (0.0200) & 0.1866 (0.0162) \\
 & 95 & 0.2150 (0.0049) & 0.2154 (0.0085) & 0.2132 (0.0062) \\
\bottomrule
\end{tabular}
}
\end{table}

\subsection{Synthetic data: Tensors}\label{subsec:synthetic-tensors}
We generate third-order tensors $X^\ast\in\mathbb R^{p\times p\times p}$ in the Tucker form
{
\[
X^\ast = \mathcal G \times_1 U_1 \times_2 U_2 \times_3 U_3,\qquad
U_1,U_2,U_3\in\mathbb{R}^{p\times p},\ U_1^\top U_1=U_2^\top U_2=U_3^\top U_3 =I_p,
\]
where $U_1,U_2,U_3$ are drawn independently with orthonormal columns. The core tensor $\mathcal{G} \in \mathbb{R}^{p\times p\times p}$ is defined entrywise by
\[
\mathcal{G}_{ijk} = \gamma_{ijk}\beta^{i+j+k},
\qquad
\beta = 0.8,
\]
where the signs $\gamma_{ijk}$ are independent with $\mathbb{P}(\gamma_{ijk} = -1) = 1/3,
\mathbb{P}(\gamma_{ijk} = 1) = 2/3$.} We observe $Y = X^\ast + Z$ with i.i.d.\ Gaussian noise $Z_{ijk}\sim\mathcal N(0,1)$ entry-wise. The SNR is controlled by scaling the signal such that
\[
\|X^\ast\|_{\mathrm F} = \lambda\sqrt{p^3},\qquad
\lambda\in\{10,50\}.
\]
The estimator $\widetilde{X}$ is obtained via the one-step HOSVD (\Cref{alg:approximate_low_rank_estimatiion}) at the Tucker rank $(r,r,r)$.

\medskip
\noindent\textbf{Dimensions and evaluated ranks.}
We vary $p\in\{20,50,75,100\}$ and for each { $p$}, we evaluate two ranks $r$ as {outlined in \Cref{tab:matrix_tensor_side}.}

Each configuration {$(p,\lambda,r)$} is repeated $R=50$ times with freshly drawn $(U_1,U_2,U_3)$ and noise. Implementations use exact batched SVD on GPU via TensorLy (PyTorch backend) \citep{kossaifi2019tensorly}. We report the sample mean and standard deviation of the relative Frobenius error
\[
\mathrm{RelErr}(\widetilde X;X^\ast) = \frac{\|\widetilde X - X^\ast\|_{\mathrm F}}{\|X^\ast\|_{\mathrm F}}\,,
\]
summarized on the right panel of \Cref{tab:matrix_tensor_side}.  We observe that the error consistently decreases as the SNR parameter $\lambda$ increases.  Overall, one-step HOSVD is robust across the tested sizes and ranks, yielding accurate estimates on the synthetic tensors.

\medskip
\noindent\textbf{Bias--variance curve.}
To empirically quantify the contributions of bias and variance to the error $\| \widetilde{X} - X^\ast \|_{\mathrm{F}}$, we consider out synthetic tensor model with $p = 100$. For each candidate Tucker rank $r \in \{12, 15, 18, \ldots, 84, 87\}$, we run $50$ independent repetitions. The bias component is obtained as the error $\|\widetilde{X} - X^\ast \|_{\mathrm{F}}$, where $\widetilde{X}$ denotes the output of Algorithm~\ref{alg:approximate_low_rank_estimatiion} when applied directly to the noiseless tensor $X^\ast$. The variance component is obtained as
\begin{equation*}
    \resizebox{\linewidth}{!}{$
        \max\left\{\sqrt{r}\left\| \mathcal{M}_1(X^\ast)\cdot\left\{U_2^{(1)} \otimes U_3^{(1)}  \right\}\right\| ,\,\sqrt{r}\left\| \mathcal{M}_2(X^\ast)\cdot\left\{U_1^{(1)} \otimes U_3^{(1)}   \right\}\right\| ,\,\sqrt{r}\left\| \mathcal{M}_3(X^\ast)\cdot\left\{U_1^{(1)} \otimes U_2^{(1)} \right\}\right\| ,\, \left\| X^\ast \times_1 U_1^{(1)} \times_2 U_2^{(1)} \times_3 U_3^{(1)} \right\|_{\mathrm{F}}  \right\}.
    $}    
\end{equation*}

\begin{figure}[!thpb]
\centering
\begin{minipage}[t]{0.48\textwidth}
  \centering
  \includegraphics[width=\linewidth]{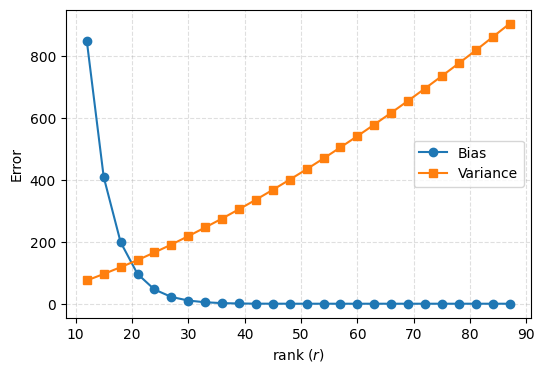}

\end{minipage}\hfill
\begin{minipage}[t]{0.48\textwidth}
  \centering
  \includegraphics[width=\linewidth]{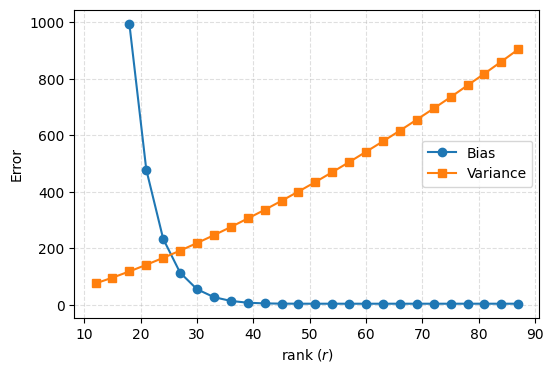}
\end{minipage}
\caption{Mean bias and variance components of the error as functions of the target rank in the synthetic experiment with $p = 100$. The left panel represent the curve with scaling parameter $\lambda = 10$ and the right panel $\lambda = 50$. }
\label{fig:bias_variance}
\end{figure}

Figure~\ref{fig:bias_variance} reports the resulting mean bias and variance curves as functions of the target rank: as the target rank increases, the bias term decreases while the variance term grows, leading to an intermediate rank that minimizes the overall error.

\subsection{Synthetic data: Matrices}\label{subsec:synthetic-matrices}
We examine the matrix analogue under a controlled SNR design that mirrors the tensor experiments. For each configuration, we draw an $m\times n$ latent signal
$$
X^\ast = U\,\Sigma\,V^\top,\qquad 
U\in\mathbb R^{m\times n},\ \ V\in\mathbb R^{n\times n},\ \ U^\top U=V^\top V=I_{n},
$$
where $\Sigma=\mathrm{diag}(\sigma_1,\ldots,\sigma_{n})$ has exponentially decaying singular values, i.e.~$\sigma_i=\beta^{\,i}$ with $\beta=0.8$. {We choose $m$ and $n$ as outlined in \Cref{tab:matrix_tensor_side}.}  We observe $Y = X^\ast +Z$, where $Z_{ij}\sim\mathcal N(0,1)$ is  i.i.d.\ Gaussian noise entry-wise. The SNR is controlled by scaling the signal such that
\[
\|X^\ast\|_{\mathrm F} = \lambda\sqrt{mn},\qquad
\lambda\in\{10,50\}.
\]
This design varies SNR via \(\lambda\) while keeping the per–entry noise scale identical across sizes. Each configuration $(m,n,\lambda,r)$ is repeated $R=50$ times with independent draws of $(U,V)$ and noise, and we report the mean and standard error of $\mathrm{RelErr}$. We use the truncated SVD estimator $Y_{(r)}$ as in Theorem~\ref{thm:matrix}. Results are shown in the left panel of Table~\ref{tab:matrix_tensor_side}. 
The truncated SVD is robust over a wide range of selected  ranks. 

\begin{table}[h!]
\centering
\caption{Mean and standard deviation (SD) of the relative error (RelErr) for matrices (left) and tensors (right). Results are shown across varying scaling parameters $\lambda$, dimensions ($m \times n$ for matrices; $p \times p \times p$ for tensors), and ranks $r$. For each configuration, results are averaged over $R = 50$ Monte Carlo simulations.}
\label{tab:matrix_tensor_side}
\begin{minipage}[t]{0.5\textwidth}
\centering
\textbf{Matrix}\par
\small
\begin{tabular}{@{}ccccl@{}}
\toprule
$\lambda$ & $m$ & $n$ & $r$ & Mean (SD) \\
\midrule
\multirow{8}{*}{10}
  & \multirow{2}{*}{100} & \multirow{2}{*}{15} & 10 & 0.1178 (0.00125) \\
  &                       &                     & 12 & 0.0906 (0.00164) \\
  & \multirow{2}{*}{250} & \multirow{2}{*}{25} & 10 & 0.1254 (0.00044) \\
  &                       &                     & 15 & 0.0868 (0.00077) \\
  & \multirow{2}{*}{375} & \multirow{2}{*}{20} & 10 & 0.1279 (0.00045) \\
  &                       &                     & 15 & 0.0927 (0.00076) \\
  & \multirow{2}{*}{500} & \multirow{2}{*}{80} & 30 & 0.0698 (0.00032) \\
  &                       &                     & 40 & 0.0795 (0.00035) \\
\midrule
\multirow{8}{*}{50}
  & \multirow{2}{*}{100} & \multirow{2}{*}{15} & 10 & 0.0844 (0.00007) \\
  &                       &                     & 12 & 0.0181 (0.00037) \\
  & \multirow{2}{*}{250} & \multirow{2}{*}{25} & 10 & 0.1079 (0.00002) \\
  &                       &                     & 15 & 0.0378 (0.00009) \\
  & \multirow{2}{*}{375} & \multirow{2}{*}{20} & 10 & 0.1068 (0.00002) \\
  &                       &                     & 15 & 0.0349 (0.00008) \\
  & \multirow{2}{*}{500} & \multirow{2}{*}{80} & 30 & 0.0134 (0.00008) \\
  &                       &                     & 40 & 0.0156 (0.00006) \\
\bottomrule
\end{tabular}
\end{minipage}\hfill
\begin{minipage}[t]{0.5\textwidth}
\centering
\textbf{Tensor}\par
\small
{
\begin{tabular}{@{}cccc@{}}
\toprule
$\lambda$ & $p$ & $r$ & Mean (SD) \\
\midrule
\multirow{8}{*}{10}
  & \multirow{2}{*}{20}  & 10 & 0.1420 (0.00350) \\
  &                       & 12 & 0.0977 (0.00208) \\
  & \multirow{2}{*}{50}  & 10 & 0.1370 (0.00347) \\
  &                       & 15 & 0.0459 (0.00102) \\
  & \multirow{2}{*}{75}  & 10 & 0.1367 (0.00348) \\
  &                       & 15 & 0.0427 (0.00113) \\
  & \multirow{2}{*}{100} & 30 & 0.0203 (0.00008) \\
  &                       & 40 & 0.0294 (0.00009) \\
\midrule
\multirow{8}{*}{50}
  & \multirow{2}{*}{20}  & 10 & 0.1361 (0.00381) \\
  &                       & 12 & 0.0840 (0.00207) \\
  & \multirow{2}{*}{50}  & 10 & 0.1362 (0.00359) \\
  &                       & 15 & 0.0410 (0.00109) \\
  & \multirow{2}{*}{75}  & 10 & 0.1368 (0.00374) \\
  &                       & 15 & 0.0412 (0.00098) \\
  & \multirow{2}{*}{100} & 30 & 0.0039 (0.00002) \\
  &                       & 40 & 0.0058 (0.00002) \\
\bottomrule
\end{tabular}
}
\end{minipage}
\end{table}

\section{Discussion}
This work provides a rank–adaptive analysis of HOSVD-based tensor denoising without assuming exact low-rank. Our main theorem (\Cref{thm:main}) yields an explicit bias–variance decomposition
\[
\|\widetilde X-X^*\|_{\mathrm F}
\;\lesssim\;
\kappa\,\sqrt{\textstyle\sum_{k=1}^3 p_k r_k \;+\; r_1 r_2 r_3}
\;+\;
\xi_{(r_1,r_2,r_3)},
\]
uniformly over all user–specified target Tucker ranks. The variance term scales with the effective degrees of freedom of the Tucker model, while the bias term is the best achievable approximation error at those ranks. Together with the matrix counterpart (\Cref{thm:matrix}), the results unify classical SVD intuition with the multilinear (tensor) setting and rigorously justify a practice common in applications: choose ranks large enough to suppress bias but not so large as to amplify noise. Our experiments on IXI brain MRI and controlled synthetic data corroborate this picture.

\medskip
\noindent\textbf{Future work.}
\emph{(i) Beyond i.i.d.~noise and full observations.} Extending the theory to heteroskedastic or correlated perturbations (e.g.~spatially correlated fields, Rician-like MRI noise) and to incomplete observations (tensor completion, masked entries) is natural. We expect the variance term to inherit problem-dependent effective dimensions (e.g.~leverage scores or sampling densities), while the bias term remains $\xi_{(r_1,r_2,r_3)}$. A key technical step is replacing isotropic concentration with mode-wise covariance-aware bounds for $\mathcal{M}_k(Z)$ and their projections. \emph{(ii) Higher orders and higher ranks.} The proof strategy appears to scale to $d$th-order tensors with the clean generalization
\[
\|\widetilde X - X^*\|_{\mathrm F}
\;\lesssim\;
\kappa\,\sqrt{\textstyle\sum_{k=1}^d p_k r_k \;+\; \prod_{k=1}^d r_k}
\;+\;
\xi_{(r_1,\ldots,r_d)},
\]
under appropriately formulated spectral gaps for each unfolding. This would offer a unified bias–variance law across orders and ranks, and would help explain the strong empirical performance of HOSVD-style estimators in truly high-order settings.

\bibliographystyle{apalike}
\bibliography{ref}

\newpage
\appendix
\begin{center}
    {\bf \Large Supplementary Materials for ``Bias--variance Tradeoff in Tensor Estimation''}
\end{center}
Throughout this appendix we work with centered sub-Gaussian random variables with parameter $\kappa^{2}$.
we assume, unless otherwise specified, that each such variable $X$ satisfies $\mathbb{E}X^{2}=\kappa^{2}$.

\section{Proofs in \Cref{sec:approx_low-rank_mat}}

\begin{proof}[Proof of \Cref{thm:matrix}]
By the triangle inequality,
\begin{equation}\label{eq: p11}
\|Y_{(r)}-X^*\|_{\mathrm{F}}
\le \|X_{(r)}^*-X^*\|_{\mathrm{F}} + \|Y_{(r)}-X_{(r)}^*\|_{\mathrm{F}}
= \xi_{(r)} + \|Y_{(r)}-X_{(r)}^*\|_{\mathrm{F}}.
\end{equation}
Since $\mathrm{rank}(Y_{(r)}-X_{(r)}^*)\le 2r$, we have
\begin{equation}\label{eq: p12}
\|Y_{(r)}-X_{(r)}^*\|_{\mathrm{F}}^2
= \sum_{k=1}^{2r}\sigma_k^2(Y_{(r)}-X_{(r)}^*).
\end{equation}
Write
\[
Y_{(r)} - X_{(r)}^* = (X^*-X_{(r)}^*) + Z + (Y_{(r)}-Y).
\]
It follows from  Ky Fan’s Theorem  (\Cref{thm: KyFan}) that,
\[
\sqrt{\sum_{i=1}^{2r}\sigma_i^2(Y_{(r)} -X _{(r)}^*)}
\le \sqrt{\sum_{i=1}^{2r}\sigma_i^2(X^*-X_{(r)} ^*)}
  + \sqrt{\sum_{i=1}^{2r}\sigma_i^2(Z)}
  + \sqrt{\sum_{i=1}^{2r}\sigma_i^2(Y_{(r)}-Y)}.
\]
Using the identity \( \sigma_i(X^* - X^*_{(r)}) = \sigma_{r + i}(X^*) \) and likewise for \( Y \), we obtain \begin{align} \|Y_{(r)} - X^*_{(r)}\|_{\mathrm{F}} &\le \sqrt{\sum_{i = 1}^{2r}\sigma_{r+i}^2(X^*)} + \sqrt{\sum_{i=1}^{2r} \sigma_i^2(Z)} + \sqrt{\sum_{i=1}^{2r}\sigma_{r+i}^2(Y)}. \label{eq: p14} \end{align} To further simplify the third term, we apply Weyl’s inequality (\Cref{thm:WeylSV}) to obtain \[ \sigma_{r+i}(Y) \le \sigma_{r+i}(X^*) + \sigma_{1}(Z), \] which gives \[ \sigma_{r+i}^2(Y) \le 2\left\{ \sigma_{r+i}^2(X^*) + \sigma_{1}^2(Z) \right\}. \] 
Substituting this into \eqref{eq: p14}, we get 
\begin{align*} \|Y_{(r)} - X^*_{(r)}\|_{\mathrm{F}} &\le \sqrt{\sum_{i = 1}^{2r}\sigma_{r+i}^2(X^*)} + \sqrt{\sum_{i=1}^{2r} \sigma_i^2(Z)} + \sqrt{2 \sum_{i = 1}^{2r}\sigma_{r+i}^2(X^*)} +2\sqrt{r}\sigma_1(Z)
\\ &\le (1 + \sqrt{2})\sqrt{\sum_{i = 1}^{2r}\sigma_{r+i}^2(X^*)} +( {2+\sqrt 2})\sqrt{r}\|Z\|\\ &= (1 + \sqrt{2})\sqrt{\sum_{i = r+1}^{3r}\sigma_{i}^2(X^*)} + ( {2+\sqrt 2})\sqrt{r}\|Z\|, \end{align*}
where the second inequality follows from the fact that for any $k\ge 1$,  we have $ \sigma_k (Z)\le \sigma_1 (Z) = \|Z\|. $  Substituting the bound of $\|Y_{(r)} - X^*_{(r)}\|_{\mathrm{F}} $ into \eqref{eq: p11}, we obtain \[ \|Y_{(r)} - X^*\|_{\mathrm{F}} \le (2 + \sqrt{2})\sqrt{\sum_{i = r+1}^{\min\{m,n\}}\sigma^2_i(X^*)} + (2 + \sqrt{2})\sqrt{r}\|Z\|. \] 
\end{proof}

\begin{proof}[Proof of \Cref{coro:subGaussian_mat}] By \Cref{lemma:subGaussian matrix}, $\|Z\| \le C_1 \sqrt{m+n}$ with probability at least $1 - \exp(-(m+n))$. The desired result follows immediately from \Cref{thm:matrix}.
\end{proof}

\begin{proof}[Proof of \Cref{thm:lower bound  matrix}]
Without loss of generality, suppose that 
$ m\le n$ and $r \le m/2 $. Therefore it suffices to show that \begin{align} \label{eq:lower bound 2 matrix}
\inf_{\widehat X     }   \sup_{X^*\in \mathcal{M}_{r}^{\xi } }
\ \mathbb{E}\bigl\|\widehat X - X^*\bigr\|_{\mathrm F}^2
\ &\ge\ c_1 \Bigl\{\kappa^2  n r    + \xi^2\Bigr\}  .
\end{align}
Let $ r'=  \lfloor \xi^2 / (n\kappa^2)    \rfloor $. Note that since  $\xi^2  \le  (m-r) n \kappa^2 $, it follows that 
$$ r' +r \le m . $$
{\bf Step 1.}
Consider the set  of matrices 
$$ \mathcal B =\{B \in \mathbb R^{m\times n} : \| B\|_{\infty } \le \kappa^2, B_{i,j}= 0  \text{ if } i>r+r'\}.$$
In this step, we show that 
\begin{align}
   \mathcal B \subset \mathcal M^{\xi}_r.
\end{align} 
To see this, let $B\in \mathcal B$, and denote $B'$ to be the matrix such that 
the first $r$ rows of $B'$ equals the first $r$ rows of $B$, and the rest of $(m-r)$ rows of $B'$ are all $0$. Then since 
$B'$ has at most $r$ rows that are nonzero, 
$\rank(B') \le r$ and 
$$ \| B-B'\| _{\mathrm F}^2 \le r' n\| B\|_\infty  = r'n \kappa^2 \le \xi ^2. $$ 
 Therefore 
 $  \mathcal B \subset \mathcal M^{\xi}_r$. So 
 \begin{align}
\inf_{\widehat X     }   \sup_{X^*\in \mathcal{M}_{r}^{\xi } }
\ \mathbb{E}\bigl\|\widehat X - X^*\bigr\|_{\mathrm F}^2
\  \ge\inf_{\widehat X     }   \sup_{X^*\in  \mathcal B }
\ \mathbb{E}\bigl\|\widehat X - X^*\bigr\|_{\mathrm F}^2  .
\end{align}
\
\\
{\bf Step 2.} Note that matrices in $\mathcal B$ are having at most $(r+r')n$ nonzero entries,   the set of matrices $\mathcal B$ can be viewed as vectors in $\mathbb  R^d$, with $d= (r+r')n$.   Denote 
$$\mathcal V= \{ V\in \mathbb R^d, \|V\|_{\infty} \le  \kappa \}.$$
Then 
$$\inf_{\widehat X     }   \sup_{X^*\in  \mathcal B }
\ \mathbb{E}\bigl\|\widehat X - X^*\bigr\|_{\mathrm F}^2  =\inf_{\widehat V} \sup _{  V\in \mathcal V } \mathbb E \| \widehat V -V \| ^2      . $$
Since by  \Cref{lemma:lower bound vector}, $ \inf_{\widehat V} \sup _{  V\in \mathcal V } \mathbb E \| \widehat V -V \| ^2   \ge c_2 \kappa^2 d, $  it follows that 
$$\inf_{\widehat X     }   \sup_{X^*\in   \mathcal{M}_{r}^{\xi } }\mathbb{E}\bigl\|\widehat X - X^*\bigr\|_{\mathrm F}^2  \ge \inf_{\widehat X     }   \sup_{X^*\in  \mathcal B }
\ \mathbb{E}\bigl\|\widehat X - X^*\bigr\|_{\mathrm F}^2 =\inf_{\widehat V} \sup _{  V\in \mathcal V } \mathbb E \| \widehat V -V \| ^2    \ge c_2\kappa^2 d = c_2 \kappa^2(r+r')n  . $$ 
{\bf Step 3.} 
There are two cases. {\bf (a)} Suppose  $\xi^2  \ge  n\kappa^2 $, then $r' =\lfloor \xi^2 / (n\kappa^2)    \rfloor\ge 1$  and therefore 
$$ \inf_{\widehat X     }   \sup_{X^*\in   \mathcal{M}_{r}^{\xi } }\mathbb{E}\bigl\|\widehat X - X^*\bigr\|_{\mathrm F}^2 \ge 
   c_2 \kappa^2(r+r')n   \ge c_2 rn + c_3 \xi^2. $$
 {\bf (b)} Suppose  $\xi^2  <  n\kappa^2 $. Then  $r'=0$ 
 \begin{align} \label{eq:lower bound 3 matrix}
\inf_{\widehat X     }   \sup_{X^*\in \mathcal{M}_{r}^{\xi } }
\ \mathbb{E}\bigl\|\widehat X - X^*\bigr\|_{\mathrm F}^2
\ &\ge\ c_2   \kappa^2 r  n    \ge \frac{c_2}{2}    (\kappa^2 r  n   + \xi^2),
\end{align}
where the first inequality follows from 
  {\bf Step 1} and {\bf Step 2},  the second inequality follows from $\xi^2  <  n\kappa^2 $.
\end{proof}

\begin{proof}[Proof of \Cref{coro:cov}]
Let 
\(
\widetilde Z = Y - X^*.
\)
By Remark 4.7.3 of \cite{vershynin2018high}, with probability at least $1-\exp(-c n)$ one has
\(
\|\widetilde Z\| \;\le\; C_1 \kappa^2 \Bigl[ \sqrt{\frac{n}{N}}+\frac{n}{N}\Bigr].
\)
Applying Theorem~\ref{thm:matrix} with $Z$ replaced by $\widetilde Z$ yields the desired bound.
\end{proof} 
\section{Proofs in  \Cref{sec:tensor}}

\begin{proof}[Proof of \Cref{thm:main}]
For any   orthogonal matrix    $U  \in \mathbb{O}^{p \times r}$, we   denote $ U _{\perp}  \in \mathbb {O}^{p \times (p-r)}$ to be the orthogonal complement of $U$, and   $$\mathcal{P}_{  U} =  {U} {U}^{\top} \quad \text{and} \quad  \mathcal{P}_{  U_{\perp}} =  {U}_{\perp} {U}_{\perp}^{\top} = I_p - \mathcal{P}_{  U}.$$
For $k\in \{ 1,2,3\}$, 
let $U_k^* \in \mathbb{O}^{p_k \times r_k}$ be the   matrix whose columns corresponds to the    top $r_k$ singular vectors of $\mathcal{M}_k(X^*)$.
\\
Throughout this proof, we assume the following good events hold:
    \begin{align}\label{eq:Z_F_norm}
    & 
    \sup_{\substack{A \in \mathbb{R}^{p_1 \times p_2 
    \times p_3}, \\ \|A\|_{\mathrm{F}} \leq 1,  A \in \mathcal T_{ (r_1, r_2, r_3)}} }\langle Z, A \rangle \leq C\kappa  \sqrt{ r_1r_2r_3 +  p_1r_1+p_2r_2+p_3r_3 } ;
    \\
    \label{eq:Z_op_norm}
    &\left\| \mathcal{M}_1(Z) \cdot W_{2} \otimes W_{3} \right\|
     \leq C\kappa\left(\sqrt{p_1 +s_{2}s_{3}   } \right)   \text{ for    non-random   } W_2 \in \mathbb O^{p_2\times s_2}, \, W_3 \in \mathbb O^{p_3\times s_3};  
    \\
    \label{eq:initial_sintheta}
    & \left\| \sin\Theta(U_{k}^{(0)}, U_k^*)\right\|= \| U_{k} ^{*\top}  U_{k }^{(0)} \| 
     \leq \frac{1}{2 \sqrt{ r_{\max}}  } \text{ for } k  \in \{ 1,2,3\}.
    \end{align}
     Indeed in  \Cref{lemma:concentration_Z},  \Cref{lemma:projection operator} and \Cref{corollary:initial sintheta}, we show that \eqref{eq:Z_F_norm},   \eqref{eq:Z_op_norm}, and \eqref{eq:initial_sintheta} hold with probability 
 at least $1 - C\exp(-c p_{\min})$.
\\
\\
 Note that 
\begin{align*}
    &\left\| \widetilde{X} - X^* \right\|_{\mathrm{F}} = \left\|  Y \times_1 \mathcal{P}_{U_1^{(1)}} \times_2 \mathcal{P}_{U_2^{(1)}} \times_3 \mathcal{P}_{U_3^{(1)}} - X^*\right\|_{\mathrm{F}}\\
    & \leq \left\| X^* \times_1 \mathcal{P}_{U_1^{(1)}} \times_2 \mathcal{P}_{U_2^{(1)}} \times_3 \mathcal{P}_{U_3^{(1)}} - X^*\right\|_{\mathrm{F}} + \left\| Z \times_1 \mathcal{P}_{U_1^{(1)}} \times_2 \mathcal{P}_{U_2^{(1)}} \times_3 \mathcal{P}_{U_3^{(1)}}\right\|_{\mathrm{F}}\\
    & = \mathrm{I}_1 + \mathrm{I}_2.
\end{align*}
\
\\
\textbf{Step 1}. For the term $\mathrm{I}_2$, observe that
\begin{align*}
    \mathrm{I}_2 &= \left\| Z \times_1 \mathcal{P}_{U_1^{(1)}} \times_2 \mathcal{P}_{U_2^{(1)}} \times_3 \mathcal{P}_{U_3^{(1)}}\right\|_{\mathrm{F}}\\
    &= \sup_{\substack{W \in \mathbb{R}^{p_1 \times p_2 
    \times p_3}, \|W\|_{\mathrm{F}} \leq 1}}\left\langle Z\times_1 \mathcal{P}_{U_1^{(1)}} \times_2 \mathcal{P}_{U_2^{(1)}} \times_3 \mathcal{P}_{U_3^{(1)}},\, W \right\rangle\nonumber\\
    &= \sup_{\substack{W \in \mathbb{R}^{p_1 \times p_2 
    \times p_3}, \|W\|_{\mathrm{F}} \leq 1}}\left\langle Z,\, \mathcal{W}\times_1 \mathcal{P}_{U_1^{(1)}} \times_2 \mathcal{P}_{U_2^{(1)}} \times_3 \mathcal{P}_{U_3^{(1)}} \right\rangle\nonumber\\
    & \leq 
    \sup_{\substack{A \in \mathbb{R}^{p_1 \times p_2 
    \times p_3}, \\ \|A\|_{\mathrm{F}} \leq 1,  A \in \mathcal T_{ (r_1, r_2, r_3)}} }\langle Z, A \rangle \leq C\kappa\left(r_1r_2r_3 + \sum_{k = 1}^3p_kr_k\right)^{1/2},
\end{align*}
where the second equality follows from the duality of the Frobenius norm, and the last inequality follows from \eqref{eq:Z_F_norm}.
\\
\\
\textbf{Step 2}.
For the term $\mathrm{I}_1$, we have that
\begin{align*}
    \mathrm{I}_1 &= \left\| X^* \times_1 \mathcal{P}_{U_1^{(1)}} \times_2 \mathcal{P}_{U_2^{(1)}} \times_3 \mathcal{P}_{U_3^{(1)}} - X^*\right\|_{\mathrm{F}}\\
    & \leq  \left\| X^* \times_1 (I_{p_1} - \mathcal{P}_{U_1^{(1)}})\right\|_{\mathrm{F}} + \left\| X^* \times_1 \mathcal{P}_{U_1^{(1)}} \times_2 (I_{p_2} - \mathcal{P}_{U_2^{(1)}})\right\|_{\mathrm{F}} \\
    & \quad + \left\| X^* \times_1 \mathcal{P}_{U_1^{(1)}} \times_2 \mathcal{P}_{U_2^{(1)}} \times_3 (I_{p_3} - \mathcal{P}_{U_3^{(1)}})\right\|_{\mathrm{F}}\\
    & \leq \sum_{k = 1}^3\left\| X^* \times_k (I_{p_k} - \mathcal{P}_{U_k^{(1)}})\right\|_{\mathrm{F}},
\end{align*}
where the last inequality follows from the observation that 
$$ \left\| X^* \times_1 \mathcal{P}_{U_1^{(1)}} \times_2 (I_{p_2} - \mathcal{P}_{U_2^{(1)}})\right\|_{\mathrm{F}} \le \left\| X^*   \times_2 (I_{p_2} - \mathcal{P}_{U_2^{(1)}})\right\|_{\mathrm{F}} \| \mathcal{P}_{U_1^{(1)}}\| \le \left\| X^*   \times_2 (I_{p_2} - \mathcal{P}_{U_2^{(1)}})\right\|_{\mathrm{F}} . $$
We only consider the case when $k = 1$, since the same arguments apply for $k = 2, 3$. We have that
\begin{align*}
    & \left\| X^* \times_1 (I_{p_1} - \mathcal{P}_{U_1^{(1)}})\right\|_{\mathrm{F}}\\
    &\leq \left\| X^* \times_1 (I_{p_1} - \mathcal{P}_{U_1^{(1)}}) \times_{2} \mathcal{P}_{U_{2}^*}\right\|_{\mathrm{F}} + \left\| X^* \times_1 (I_{p_1} - \mathcal{P}_{U_1^{(1)}}) \times_{2} (I_{p_2} - \mathcal{P}_{U_{2}^*})\right\|_{\mathrm{F}}\\
    &\leq \left\| X^* \times_1 (I_{p_1} - \mathcal{P}_{U_1^{(1)}}) \times_{2} \mathcal{P}_{U_{2}^*}\right\|_{\mathrm{F}} + \left\| X^* \times_{2} (I_{p_2} - \mathcal{P}_{U_{2}^*})\right\|_{\mathrm{F}}\\
    &\leq \left\| X^* \times_1 (I_{p_1} - \mathcal{P}_{U_1^{(1)}}) \times_{2} \mathcal{P}_{U_{2}^*} \times_{3} \mathcal{P}_{U_{3}^*} \right\|_{\mathrm{F}} + 
    \left\|   X^* \times_1 (I_{p_1} - \mathcal{P}_{U_1^{(1)}}) \times_{2}   \mathcal{P}_{U_{2}^*} \times_{3} (I_{p_3} - \mathcal{P}_{U_{3}^*})\right\|_{\mathrm{F}}
    \\ 
    &+ 
    \left\| X^* \times_{2} (I_{p_2} - \mathcal{P}_{U_{2}^*})\right\|_{\mathrm{F}}   
    \\
    &
    \le   \left\| X^* \times_1 (I_{p_1} - \mathcal{P}_{U_1^{(1)}}) \times_{2} \mathcal{P}_{U_{2}^*} \times_{3} \mathcal{P}_{U_{3}^*} \right\|_{\mathrm{F}} + 
    \left\|   X^*  \times_{3} (I_{p_3} - \mathcal{P}_{U_{3}^*})\right\|_{\mathrm{F}}+ 
    \left\| X^* \times_{2} (I_{p_2} - \mathcal{P}_{U_{2}^*})\right\|_{\mathrm{F}}   \\ 
    &= \left\| X^* \times_1 (I_{p_1} - \mathcal{P}_{U_1^{(1)}}) \times_{2} \mathcal{P}_{U_{2}^*} \times_{3} \mathcal{P}_{U_{3}^*} \right\|_{\mathrm{F}} +  2\xi_{(r_1,r_2,r_3)},
\end{align*}
where the   second follows from that $\|I_{p_1} - \mathcal{P}_{U_1^{(1)}}\| \leq 1$,   and the equality follows from \Cref{lemma:project_forbuious}. 
\\
\\
\textbf{Step 3}.
We bound $\| X^* \times_1 (I_{p_1} - \mathcal{P}_{U_1^{(1)}}) \times_{2} \mathcal{P}_{U_{2}^*} \times_{3} \mathcal{P}_{U_{3}^*} \|_{\mathrm{F}}$ in this step. Consider the different but relevant  quantity   
\begin{align*}
    &\left\| X^* \times_1 (I_{p_1} - \mathcal{P}_{U_1^{(1)}}) \times_{2} \mathcal{P}_{U_{2}^{(0)}} \times_{3} \mathcal{P}_{U_{3}^{(0)}} \right\|_{\mathrm{F}}\\
    &=  \left\|(I_{p_1} - \mathcal{P}_{U_1^{(1)}}) \cdot \mathcal{M}_1(X^*) \cdot (\mathcal{P}_{U_{2}^{(0)}} \otimes \mathcal{P}_{U_{3}^{(0)}}) \right\|_{\mathrm{F}}\\
    &=  \left\|(I_{p_1} - \mathcal{P}_{U_1^{(1)}}) \cdot \mathcal{M}_1(X^*) \cdot (U_{2}^{(0)} \otimes U_{3}^{(0)})(U_{2}^{(0)} \otimes U_{3}^{(0)})^{\top} \right\|_{\mathrm{F}}\\
    &=  \left\|(I_{p_1} - \mathcal{P}_{U_1^{(1)}}) \cdot \mathcal{M}_1(X^*) \cdot (U_{2}^{(0)} \otimes U_{3}^{(0)}) \right\|_{\mathrm{F}}\\
    &\geq  \left\|(I_{p_1} - \mathcal{P}_{U_1^{(1)}}) \cdot \mathcal{M}_1(X^*) \cdot \mathcal{P}_{U_2^* \otimes U_3^*} \cdot (U_{2}^{(0)} \otimes U_{3}^{(0)}) \right\|_{\mathrm{F}}\\
    & \quad - \left\|(I_{p_1} - \mathcal{P}_{U_1^{(1)}}) \cdot \mathcal{M}_1(X^*) \cdot (I_{p_2p_3} - \mathcal{P}_{U_2^* \otimes U_3^*}) \cdot (U_{2}^{(0)} \otimes U_{3}^{(0)}) \right\|_{\mathrm{F}}\\
    &= \mathrm{II}_1 - \mathrm{II}_2,
\end{align*}
where the first inequality follows from that $(U_{2}^{(0)} \otimes U_{3}^{(0)}) \in \mathbb{O}^{(p_2p_3) \times (r_2r_3)}$, and \Cref{lemma:Fobenius norm preserve}.
\\
Before analyzing the terms $\mathrm{II}_1$ and $\mathrm{II}_2$, we firstly note that
\begin{equation}\label{eq:facts}
\begin{aligned}
     (U_2^* \otimes U_3^*)^{\top} \cdot (U_{2}^{(0)} \otimes U_{3}^{(0)}) &= (U_2^{*\,\top} U_{2}^{(0)}) \otimes (U_3^{*\,\top} U_{3}^{(0)}),\\
     \sigma_{\min}\left((U_2^{*\,\top} U_{2}^{(0)}) \otimes (U_3^{*\,\top} U_{3}^{(0)})\right) &= \sigma_{\min}\left(U_2^{*\,\top} U_{2}^{(0)}\right)  \sigma_{\min}\left(U_3^{*\,\top} U_{3}^{(0)}\right),\\
    \sigma_{\min}^2\left(U_k^{*\,\top} U_{k}^{(0)}\right) &= 1 - \left\|U_{k\,\perp}^{*\,\top} U_{k}^{(0)}\right\|^2 = 1 - \left\|\sin\Theta(U_{k}^{*}, U_{k}^{(0)})\right\|^2,
\end{aligned}
\end{equation}
which hold due to the properties of the Kronecker product and \Cref{lemma:sintheta}.
\\
For the term $\mathrm{II}_1$, we have that
\begin{align*}
    \mathrm{II}_1 &=  \left\|(I_{p_1} - \mathcal{P}_{U_1^{(1)}}) \cdot \mathcal{M}_1(X^*) \cdot (U_{2}^{*} \otimes U_{3}^{*}) \cdot (U_{2}^{*} \otimes U_{3}^{*})^{\top} \cdot (U_{2}^{(0)} \otimes U_{3}^{(0)}) \right\|_{\mathrm{F}}\\
    &\geq \left\|(I_{p_1} - \mathcal{P}_{U_1^{(1)}}) \cdot \mathcal{M}_1(X^*) \cdot (U_{2}^{*} \otimes U_{3}^{*})\right\|_{\mathrm{F}} \sigma_{\min}(U_{2}^{*\,\top} U_{2}^{(0)}) \sigma_{\min}(U_{3}^{*\,\top} U_{3}^{(0)})\\
    &= \left\|X^* \times_1 (I_{p_1} - \mathcal{P}_{U_1^{(1)}}) \times_2 \mathcal{P}_{U_{2}^{*}} \times_3 \mathcal{P}_{U_{3}^{*}}\right\|_{\mathrm{F}} \sqrt{\left(1 - \left\|\sin\Theta(U_{2}^{*}, U_{2}^{(0)})\right\|^2\right)\left(1 - \left\|\sin\Theta(U_{3}^{*}, U_{3}^{(0)})\right\|^2\right)}
    \\
    &\ge 
    \frac{3}{4}\left\|X^* \times_1 (I_{p_1} - \mathcal{P}_{U_1^{(1)}}) \times_2 \mathcal{P}_{U_{2}^{*}} \times_3 \mathcal{P}_{U_{3}^{*}}\right\|_{\mathrm{F}} 
\end{align*}
where the first inequality follows from \Cref{lemma:lb_forbuious}, and the second equality follows from \eqref{eq:facts}, and the last inequality follows from \eqref{eq:initial_sintheta}.
\\
\\
For the term $\mathrm{II}_2$, we have that
\begin{align*}
    \mathrm{II}_2 &= \left\|(I_{p_1} - \mathcal{P}_{U_1^{(1)}}) \cdot \mathcal{M}_1(X^*) \cdot (I_{p_2p_3} - \mathcal{P}_{U_2^* \otimes U_3^*}) \cdot (U_{2}^{(0)} \otimes U_{3}^{(0)}) \right\|_{\mathrm{F}}\\
    &= \left\|(I_{p_1} - \mathcal{P}_{U_1^{(1)}}) \cdot \mathcal{M}_1(X^*) \cdot (U_2^* \otimes U_3^*)_{\perp} \cdot (U_2^* \otimes U_3^*)_{\perp}^{\top} \cdot (U_{2}^{(0)} \otimes U_{3}^{(0)}) \right\|_{\mathrm{F}}\\
    &\leq \left\|I_{p_1} - \mathcal{P}_{U_1^{(1)}}\right\| \left\|\mathcal{M}_1(X^*) \cdot (U_2^* \otimes U_3^*)_{\perp} \right\|_{\mathrm{F}} \left\| (U_2^* \otimes U_3^*)_{\perp}^{\top} \cdot (U_{2}^{(0)} \otimes U_{3}^{(0)}) \right\|\\
    &\le   \left\|\mathcal{M}_1(X^*) \cdot (U_2^* \otimes U_3^*)_{\perp} \right\|_{\mathrm{F}}  .
\end{align*}
Note that 
\begin{align*}
    &\left\| \mathcal{M}_1(X^*) \cdot (U_2^* \otimes U_3^*)_{\perp}\right\|_{\mathrm{F}} \\
    &= \left\| \mathcal{M}_1(X^*) \cdot  [ U_{2 \,\perp} ^* \otimes U_3^* \quad U_{2} ^* \otimes U_{3 \,\perp}^* \quad U_{2 \,\perp} ^* \otimes U_{3 \,\perp}^*] \right\|_{\mathrm{F}} \\
    &= \sqrt{\left\| \mathcal{M}_1(X^*) \cdot  (U_{2 \,\perp} ^* \otimes U_3^*) \right\|_{\mathrm{F}}^2 + \left\| \mathcal{M}_1(X^*) \cdot  (U_{2} ^* \otimes U_{3 \,\perp}^*) \right\|_{\mathrm{F}}^2 + \left\| \mathcal{M}_1(X^*) \cdot  (U_{2 \,\perp} ^* \otimes U_{3 \,\perp}^*) \right\|_{\mathrm{F}}^2}\\
    &\leq \left\| \mathcal{M}_1(X^*) \cdot  (U_{2 \,\perp} ^* \otimes U_3^*) \right\|_{\mathrm{F}} + \left\| \mathcal{M}_1(X^*) \cdot  (U_{2} ^* \otimes U_{3 \,\perp}^*) \right\|_{\mathrm{F}} + \left\| \mathcal{M}_1(X^*) \cdot  (U_{2 \,\perp} ^* \otimes U_{3 \,\perp}^*) \right\|_{\mathrm{F}}\\
    &= \left\| X^* \times_2  U_{2 \,\perp}^* \times_3 U_3^* \right\|_{\mathrm{F}} + \left\| X^* \times_2 U_{2}^* \times_3 U_{3 \,\perp}^* \right\|_{\mathrm{F}} + \left\| X^* \times_2  U_{2 \,\perp}^* \times_3 U_{3 \,\perp}^* \right\|_{\mathrm{F}}\\
    &\le  \left\| X^* \times_2  U_{2 \,\perp}^* \right\|_{\mathrm{F}} + \left\| X^*   \times_3 U_{3 \,\perp}^* \right\|_{\mathrm{F}} + \left\| X^* \times_2  U_{2 \,\perp}^*   \right\|_{\mathrm{F}}
    \\
    &\le  3\xi_{(r_1,r_2,r_3)}
\end{align*}
where the last inequality follows from 
\Cref{lemma:project_forbuious}. 
Therefore,
\begin{align*}
    \mathrm{II}_2 \leq 3\xi_{(r_1,r_2,r_3)}.
\end{align*}
Combining $\mathrm{II}_1$ and $\mathrm{II}_2$, we have
\begin{align*}
      \left\| X^* \times_1 (I_{p_1} - \mathcal{P}_{U_1^{(1)}}) \times_{2} \mathcal{P}_{U_{2}^*} \times_{3} \mathcal{P}_{U_{3}^*} \right\|_{\mathrm{F}} 
    \le  \frac{4}{3}\left\|X^* \times_1 (I_{p_1} - \mathcal{P}_{U_1^{(1)}}) \times_2 \mathcal{P}_{U_{2}^{(0)}} \times_3 \mathcal{P}_{U_{3}^{(0)}}\right\|_{\mathrm{F}}  + 4 \xi_{(r_1,r_2,r_3)}.
\end{align*}
{\bf Step 4.}
We bound $\left\|X^* \times_1 (I_{p_1} - \mathcal{P}_{U_1^{(1)}}) \times_2 \mathcal{P}_{U_{2}^{(0)}} \times_3 \mathcal{P}_{U_{3}^{(0)}}\right\|_{\mathrm{F}}$ in this step.
Note that 
\begin{align*}
  &  \| X^* \times_1 (I_{p_1} - \mathcal{P}_{U_1^{(1)}}) \times_{2} \mathcal{P}_{U_{2}^{(0)}} \times_{3} \mathcal{P}_{U_{3}^{(0)}} \|_{\mathrm{F}}
    \\ 
    = &\| \mathcal{P}_{U_{1\perp} ^{(1)}}  \cdot \mathcal M_1 (X^*)  \cdot  ( \mathcal{P}_{U_{2}^{(0)}}  \otimes  \mathcal{P}_{U_{3}^{(0)}}  ) \| _{\mathrm{F}}
    = \| \mathcal{P}_{U_{1\perp} ^{(1)}}  \cdot \mathcal M_1 (X^*)  \cdot  (  U_{2}^{(0)}   \otimes   U_{3}^{(0)})  \| _{\mathrm{F}}
\end{align*}  
It suffices to apply \Cref{lemma:perterbation_approx}  to  bound  
$$ \| \mathcal{P}_{U_{1\perp} ^{(1)}}  \cdot \mathcal M_1 (X^*)  \cdot  (  U_{2}^{(0)}   \otimes   U_{3}^{(0)})  \| _{\mathrm{F}}.$$
 Since $U_{1\perp} ^{(1)}$   corresponds to the SVD of $\mathcal M_1 (Y) \cdot  (  U_{2}^{(0)}   \otimes   U_{3}^{(0)}) $, let 
 $$ A= \mathcal M_1 (Y) \cdot  (  U_{2}^{(0)}   \otimes   U_{3}^{(0)}) \quad \text{and}\quad  B =\mathcal M_1 (X^*) \cdot  (  U_{2}^{(0)}   \otimes   U_{3}^{(0)})  .$$ It follows from \Cref{lemma:perterbation_approx}  that
\begin{align}\nonumber
     \left\| X^* \times_1 (I_{p_1} - \mathcal{P}_{U_1^{(1)}}) \times_{2} \mathcal{P}_{U_{2}^{(0)}} \times_{3} \mathcal{P}_{U_{3}^{(0)}}  \right\|_{\mathrm{F}} 
    = & \| \mathcal{P}_{U_{1\perp} ^{(1)}}  \cdot \mathcal M_1 (X^*)  \cdot  (  U_{2}^{(0)}   \otimes   U_{3}^{(0)})  \| _{\mathrm{F}} 
    \\ \label{eq:step 3 bound} \le&  C_1    \|B-B_{r_1}\|_F +  C_2\sqrt{r_1} \| A-B\|.
\end{align}
Here 
\begin{align}\nonumber 
     \left\|A-B \right\| = &\left\|\mathcal{M}_1(Z) \cdot  U_{2}^{(0)} \otimes U_{3}^{(0)} \right\|\\
    \nonumber  = & \left\|\mathcal{M}_1(Z) \cdot (   U^*_2U^{*\top} _2   + U^{ *}   _{2\perp} U^{\top *}   _{2\perp} ) U_{2}^{(0)} \otimes U_{3}^{(0)} \right\|
    \\\nonumber 
    \le & \left\|\mathcal{M}_1(Z) \cdot   (  U^*_2U^{*\top} _2     U_{2}^{(0)}  ) \otimes U_{3}^{(0)} \right\| + \left\|\mathcal{M}_1(Z) \cdot (    U^{ *}   _{2\perp} U^{\top *}   _{2\perp} U_{2}^{(0)}  )\otimes U_{3}^{(0)} \right\|
    \\ \label{eq:A-B term 1}
    \le & \left\|\mathcal{M}_1(Z) \cdot      U^*_2   \otimes U_{3}^{(0)} \right\| \| U^{*\top} _2     U_{2}^{(0)} \| + 
    \left\|\mathcal{M}_1(Z) \cdot      U^{ *}   _{2\perp}   \otimes U_{3}^{(0)} \right\| \|U^{\top *}   _{2\perp} U_{2}^{(0)} \| .
\end{align}
Note that 
\begin{align}\nonumber 
&\left\|\mathcal{M}_1(Z) \cdot      U^*_2   \otimes U_{3}^{(0)} \right\| \| U^{*\top} _2     U_{2}^{(0)} \| 
=  \left\|\mathcal{M}_1(Z) \cdot      U^*_2   \otimes (   U^*_3U^{*\top} _3   + U^{ *}   _{3 \perp} U^{\top *}   _{3\perp} )  U_{3}^{(0)} \right\|  
\\\nonumber 
\le & \left\|\mathcal{M}_1(Z) \cdot      U^*_2   \otimes (   U^*_3U^{*\top} _3 U_{3}^{(0)} ) \right\|  +\left\|\mathcal{M}_1(Z) \cdot      U^*_2   \otimes (     U^{ *}   _{3 \perp} U^{\top *}   _{3\perp} )  U_{3}^{(0)} \right\|  
\\  \nonumber 
\le & \left\|\mathcal{M}_1(Z) \cdot      U^*_2   \otimes     U^*_3  \right\| \| U^{*\top} _3 U_{3}^{(0)}\|     + \left\|\mathcal{M}_1(Z) \cdot      U^*_2   \otimes     U^{ *}   _{3 \perp} \right\|  \|U^{\top *}   _{3\perp}   U_{3}^{(0)} \| 
\\ \label{eq:A-B term 2}
\le & C_4\kappa ( \sqrt{p_1 +r_2r_3}  )+ C_5\kappa (\sqrt{ p_1 +  p_3 r_2 }) r^{-1/2}_{\max} ,
\end{align}
where the last inequality follows from 
\eqref{eq:Z_op_norm} and the fact that $\| U^{*\top} _2     U_{2}^{(0)} \| \le 1$.
In addition
\begin{align} \nonumber 
     & \left\|\mathcal{M}_1(Z) \cdot      U^{ *}   _{2\perp}   \otimes U_{3}^{(0)} \right\| \|U^{\top *}   _{2\perp} U_{2}^{(0)} \|  
\le   \left\|\mathcal{M}_1(Z) \cdot      U^{ *}   _{2\perp}  \otimes(U^{ *}   _{3  } U^{\top *}   _{3 }   + U^{ *}   _{3 \perp} U^{\top *}   _{3\perp} ) U_{3}^{(0)} \right\| \|U^{\top *}   _{2\perp} U_{2}^{(0)} \| 
\\ \nonumber  
\le &\left\|\mathcal{M}_1(Z) \cdot      U^*_{2\perp}   \otimes (   U^*_3U^{*\top} _3 U_{3}^{(0)} ) \right\|  \|U^{\top *}   _{2\perp} U_{2}^{(0)} \| +\left\|\mathcal{M}_1(Z) \cdot      U^*_2   \otimes (     U^{ *}   _{3 \perp} U^{\top *}   _{3\perp}    U_{3}^{(0)} )\right\|   \|U^{\top *}   _{2\perp} U_{2}^{(0)} \| 
\\ \nonumber  
\le &\left\|\mathcal{M}_1(Z) \cdot      U^*_{2\perp}   \otimes     U^*_3   \right\| \|U^{*\top} _3 U_{3}^{(0)}\|  \|U^{\top *}   _{2\perp} U_{2}^{(0)} \| +\left\|\mathcal{M}_1(Z) \cdot      U^*_2   \otimes       U^{ *}   _{3 \perp}  \right\| \|U^{\top *}   _{3\perp}   U_{3}^{(0)}\|  \|U^{\top *}   _{2\perp} U_{2}^{(0)} \| 
\\\label{eq:A-B term 3}
      \le &C\kappa ( \sqrt{ p_1 + p_2r_3}   ) r_{\max}^{-1/2}+
      C\kappa ( \sqrt{ p_1 + p_3r_2}   ) r_{\max}^{-  1}.
\end{align}
Therefore \eqref{eq:A-B term 1}, \eqref{eq:A-B term 2} and \eqref{eq:A-B term 3} leads to
\begin{align}
    \label{eq:A-B operator norm boubd term 1}\| A-B\| \le  C_4 \kappa ( \sqrt {p_1} + \sqrt{r_2r_3} + \sqrt{p_2r_3} r_{\max}^{-1/2}  +\sqrt{p_3r_2} r_{\max}^{-1/2}  ).
\end{align} 
In addition,
\begin{align} \nonumber 
    \|B-B_{r_1}\|_{\mathrm F} = &\|  \mathcal M_1 (X^*) \cdot     U_{2}^{(0)}   \otimes   U_{3}^{(0)} -   \{   \mathcal M_1 (X^*) \cdot     U_{2}^{(0)}   \otimes   U_{3}^{(0)} \} _{r_1}\|_{\mathrm{F}}
    \\ \nonumber 
    \le & \|  \mathcal M_1 (X^*) \cdot     U_{2}^{(0)}   \otimes   U_{3}^{(0)} -    \{   \mathcal M_1 (X^*)\}_{r_1} \cdot     U_{2}^{(0)}   \otimes   U_{3}^{(0)} \|_{\mathrm{F}}
    \\ \nonumber 
    \le &\| ( \mathcal M_1 (X^*)  -  \{   \mathcal M_1 (X^*)\}_{r_1} )   U_{2}^{(0)}   \otimes   U_{3}^{(0)} \|_{\mathrm{F}}
    \\ \label{eq:A-B operator norm boubd term 2}
    \le & \|  \mathcal M_1 (X^*)  -  \{   \mathcal M_1 (X^*)\}_{r_1} \|_{\mathrm{F}} = \sqrt{  \sum_{j=r_1+1}^{\rank (\mathcal M_1 (X^*)) }  \sigma_j ^2(\mathcal M_1 (X^*)) }\le \xi_{(r_1,r_2,r_3)} .
\end{align}
Here  $\{   \mathcal M_1 (X^*) \cdot     U_{2}^{(0)}   \otimes   U_{3}^{(0)} \} _{r_1}$ indicate the best rank $r_1$ estimate of the matrix $\mathcal M_1 (X^*) \cdot     U_{2}^{(0)}   \otimes   U_{3}^{(0)}$ in the first equality, and so for any rank $r_1$ matrix $\Phi$, 
$$ \|  \mathcal M_1 (X^*) \cdot     U_{2}^{(0)}   \otimes   U_{3}^{(0)} -   \{   \mathcal M_1 (X^*) \cdot     U_{2}^{(0)}   \otimes   U_{3}^{(0)} \} _{r_1}\|_{\mathrm{F}} \le \|  \mathcal M_1 (X^*) \cdot     U_{2}^{(0)}   \otimes   U_{3}^{(0)} -   \Phi\|_{\mathrm{F}} ;$$
the second inequality holds because  $\{   \mathcal M_1 (X^*)\}_{r_1} \cdot     U_{2}^{(0)}   \otimes   U_{3}^{(0)}$
is at most rank $r_1$, and the last inequality follows from \Cref{lemma:project_forbuious}.
\\
\\
It follows from \eqref{eq:step 3 bound}, \eqref{eq:A-B operator norm boubd term 1} and \eqref{eq:A-B operator norm boubd term 2} that 
$$ \left\| X^* \times_1 (I_{p_1} - \mathcal{P}_{U_1^{(1)}}) \times_{2} \mathcal{P}_{U_{2}^{(0)}} \times_{3} \mathcal{P}_{U_{3}^{(0)}}  \right\|_{\mathrm{F}}  \le C_5 \kappa ( \sqrt {p_1r_1} + \sqrt{r_1r_2r_3}  +\sqrt{p_2r_2} + \sqrt{p_3r_3}    ) + C_5\xi_{(r_1,r_2,r_3)}.  $$
 \
 \\
\textbf{Step 5}. The conclusions of 
{\bf Step 3} and {\bf Step 4}   lead to 
$$\left\| X^* \times_1 (I_{p_1} - \mathcal{P}_{U_1^{(1)}}) \times_{2} \mathcal{P}_{U_{2}^*} \times_{3} \mathcal{P}_{U_{3}^*} \right\|_{\mathrm{F}} \le C_6 \kappa ( \sqrt {p_1r_1} + \sqrt{r_1r_2r_3}  +\sqrt{p_2r_2} + \sqrt{p_3r_3}    ) + C_6\xi_{(r_1,r_2,r_3)}. $$
This bound together with {\bf Step 1} and {\bf Step 2} leads to 
\begin{align*}
    &\left\| \widetilde{X} - X^* \right\|_{\mathrm{F}} \leq C_7 \kappa ( \sqrt {p_1r_1}  +\sqrt{p_2r_2} + \sqrt{p_3r_3} + \sqrt{r_1r_2r_3}    ) + C_7\xi_{(r_1,r_2,r_3)}.
\end{align*}
 
\end{proof}

\begin{proof}[Proof of \Cref{corollary:relative error}]
    It suffices to observe that for $k=1,2,3$
    $$\|X^* \|_{\mathrm F} ^2 \ge  r_k\sigma_{r_k} ^2 (\mathcal M_k(X^*)) . $$
    The desired follows from the assumption that $r_k \le \sqrt {p_{\min}}$ and \eqref{eq:tensor error bound}.
\end{proof}

\begin{proof}[Proof  of \Cref{thm:lower}] 
\ 
\\
{\bf Step 1.} We follow a similar strategy as in the proof of \Cref{thm:lower bound  matrix} to show 
\begin{align} \label{eq:lower bound 3 tensor}
\inf_{\widehat X     }   \sup_{X^\ast\in \mathcal{T}_{(r_1,r_2,r_3)}^{\xi }  }
\ \mathbb{E}\bigl\|\widehat X - X^\ast\bigr\|_{\mathrm F}^2
\ &\ge\ c   ( \kappa^2 r_1r_2r_3   +\xi^2) .
\end{align}
Let 
$$r_1' = \lfloor \xi^2 /(r_2r_3\kappa^2)\rfloor. $$
Consider the set  of tensors 
$$ \mathcal B =\{B \in \mathbb R^{p_1\times p_2 \times p_3} : \| B\|_{\infty } \le \kappa^2, B_{i,j,k}= 0  \text{ if } i> r_1 +r_1'\text{ or } j> r_2 \text{ or } k> r_3 \}.$$
In this step, we show that 
\begin{align}
   \mathcal B \subset  \mathcal{T}_{(r_1,r_2,r_3)}^{\xi }.
\end{align} 
To see this, let $B\in \mathcal B$, and denote 
$B'$ to be the matrix such that 
$$ B'_{i,j,k} =\begin{cases}
    B_{i,j,k} &\text{if }  i\le r_1 \text{ and } j\le r_2 \text{ and } k\le r_3;\\
    0  & \text{otherwise}.
\end{cases}$$
Then $B' \in \mathcal T_{(r_1,r_2,r_3)}$ and 
$$ \| B-B'\| _{\mathrm F}^2 \le r'_1r_2r_3 \| B\|_\infty  = r'_1 r_2r_3  \kappa^2 \le \xi ^2. $$ 
 Therefore 
 $  \mathcal B \subset \mathcal{T}_{(r_1,r_2,r_3)}^{\xi } $. So 
 \begin{align}
\inf_{\widehat X     }   \sup_{X^*\in \mathcal{T}_{(r_1,r_2,r_3)}^{\xi }  }
\ \mathbb{E}\bigl\|\widehat X - X^*\bigr\|_{\mathrm F}^2
\  \ge\inf_{\widehat X     }   \sup_{X^*\in  \mathcal B }
\ \mathbb{E}\bigl\|\widehat X - X^*\bigr\|_{\mathrm F}^2  .
\end{align}
\
\\
{\bf Step 1.1.} Note that the tensors in $\mathcal B$ are having at most $(r_1+r_1')r_2r_3$ nonzero entries,   the set of matrices $\mathcal B$ can be viewed as vectors in $\mathbb  R^d$, with $d= (r_1+r_1')r_2r_3$.   Denote 
$$\mathcal V= \{ V\in \mathbb R^d, \|V\|_{\infty} \le  \kappa \}.$$
Then 
$$\inf_{\widehat X     }   \sup_{X^*\in  \mathcal B }
\ \mathbb{E}\bigl\|\widehat X - X^*\bigr\|_{\mathrm F}^2  =\inf_{\widehat V} \sup _{  V\in \mathcal V } \mathbb E \| \widehat V -V \| ^2      . $$
Since by  \Cref{lemma:lower bound vector}, $ \inf_{\widehat V} \sup _{  V\in \mathcal V } \mathbb E \| \widehat V -V \| ^2   \ge c_2 \kappa^2 d, $  it follows that 
$$\inf_{\widehat X     }   \sup_{X^*\in  \mathcal B }
\ \mathbb{E}\bigl\|\widehat X - X^*\bigr\|_{\mathrm F}^2 =\inf_{\widehat V} \sup _{  V\in \mathcal V } \mathbb E \| \widehat V -V \| ^2    \ge c_2\kappa^2 d = c_2 \kappa^2(r_1+r_1')r_2r_3 . $$ 
{\bf Step 1.2.} To show \eqref{eq:lower bound 3 tensor}, there are two cases. {\bf (a)} Suppose  $\xi^2  \ge  r_2r_3\kappa^2 $, then $r_1' = \lfloor  \xi^2 / (r_2r_3\kappa^2)    \rfloor \ge 1$  and therefore 
$$ \inf_{\widehat X     }   \sup_{X^*\in  \mathcal{T}_{(r_1,r_2,r_3)}^{\xi }  }\mathbb{E}\bigl\|\widehat X - X^*\bigr\|_{\mathrm F}^2 \ge 
    c_2 \kappa^2(r_1+r_1')r_2r_3    \ge c_2 r_1r_2r_3 + c_3 \xi^2. $$
 {\bf (b)} Suppose  $\xi^2  < r_2r_3\kappa^2  $. Then  $r'_1=0$ 
 \begin{align*}  
\inf_{\widehat X     }   \sup_{X^*\in \mathcal{T}_{(r_1,r_2,r_3)}^{\xi } }
\ \mathbb{E}\bigl\|\widehat X - X^*\bigr\|_{\mathrm F}^2
\ &\ge\ c_2   \kappa^2 r_1r_2r_3  \ge \frac{c_2}{2}    (\kappa^2    r_1r_2r_3  + \xi^2),
\end{align*}
where the first inequality follows from 
  {\bf Step 1.1},  the second inequality follows from $\xi^2  < r_2r_3\kappa^2  $. 
\
\\
{\bf Step 2.} Note that  
\begin{align}
    \label{eq:lower bound 3}\inf_{\widehat X   } \ \sup_{X^*\in \mathcal{T}_{(r_1,r_2,r_3)}^\xi   }
\ \mathbb{E}\bigl\|\widehat X - X^*\bigr\|_{\mathrm F}^2 \ge \inf_{\widehat X   } \ \sup_{X^*\in \mathcal{T}_{(r_1,r_2,r_3)}  }
\ \mathbb{E}\bigl\|\widehat X - X^*\bigr\|_{\mathrm F}^2
\  \ge\ c_3   \kappa^2 \sum_{k=1}^3 p_k r_k  ,
\end{align}  
where the second inequality follows from \citet{zhang2018tensor}.
The desired result directly follows from \eqref{eq:lower bound 3 tensor} and \eqref{eq:lower bound 3}.
\end{proof}

\section{SVD for Unbalanced Matrices}

\begin{lemma}\label{lemma:svd}  Suppose 
    \[
    Y = X + Z \in \mathbb{R}^{n \times m},
    \] where $X$ is a non-random  matrix of arbitrary rank, and $Z$ is a random matrix whose entries are i.i.d.~sub-Gaussian random variables with mean zero and the sub-Gaussian norm $\|Z_{ij}\|_{\psi_2} = \kappa < \infty$. For any $r \leq \min\{n,m\}$, write the full SVD of $X$ as
\[
X = U \Sigma V^\top = \begin{bmatrix} U_{r} \ U_{\perp} \end{bmatrix} \cdot \begin{bmatrix} \Sigma_{r} & \\ & \Sigma_{\perp} \end{bmatrix} \cdot \begin{bmatrix} V_{r}^\top \\ V_{\perp}^\top \end{bmatrix} = X_{r} + X_{\perp}.
\]
Here $U_{r} \in \mathbb{O}_{m, r}$, $V_{r} \in \mathbb{O}_{m, r}$ correspond to the leading $r$ left and right singular vectors of $X$. 
Suppose that 
\[\{\sigma_r(X) - \sigma_{r+1}(X)\}^2 \geq  C_{\mathrm{gap}}\kappa^2\{\sqrt{mn} + m\}\]   where  $C_{\mathrm{gap}} > 0$ is a sufficient large constant. Then with probability at least $1 - C_1\exp(-C_2n)$, it holds that
\begin{align*}
    \left\|\sin\Theta\left(\widehat{V}_{r}, V_{r}\right) \right\| ^2 \leq C_3 \bigg\{  \frac{m\kappa^2 }{(\sigma_r (X) - \sigma_{r+1} (X)  ) ^2 } +  \frac{\kappa^4 nm }{(\sigma_r (X) - \sigma_{r+1} (X)  ) ^4 } \bigg\} ,
\end{align*}
where $C_1, C_2, C_3 > 0$ are  absolute constants only depending on $C_{\mathrm{gap}}$.
\end{lemma} 

\begin{proof}
Note that by assumption, we have 
\begin{align*}
       \mathbb{E}[Z^{\top}Z]  = n\kappa^2I_m,
       \quad 
       \mathbb{E}[Y^{\top}Y]  = V_{r}\Sigma_{r}^2V_{r}^{\top} + V_{\perp}\Sigma_{\perp}^2V_{\perp}^{\top} + n\kappa^2I_m,\quad 
       \mathbb{E}[V_{r}^{\top}Y^{\top}YV_{r}]  = \Sigma_{r}^2 + n\kappa^2I_r.
\end{align*}
Define the    diagonal weighting matrix
\[
M = \diag\left((\sigma_1^2 + n\kappa^2)^{-1/2}, \dots, (\sigma_r^2 + n\kappa^2)^{-1/2}\right) \in \mathbb R^{r\times r}.
\]
Then it holds that
\begin{align*}
        YV_{r}M &= \left(X_{r} + X_{\perp} + Z \right)V_{r}M = \left(X_{r} + Z \right)V_{r}M,
        \\
        M^{\top}\,\mathbb{E}\left[ V_{r}^{\top}Y^{\top}YV_{r} \right]\,M &= M^{\top}\, \mathbb{E}\left[V_{r}^{\top} \left(X_{r} + Z\right)^{\top} \left(X_{r} + Z\right) V_{r} \right]\,M = I_r.
\end{align*}
  \textbf{Step 1}. 
   For $\sigma_r(YV_{r})$, observe that
    \begin{align} 
        \sigma_r^2(YV_{r}) = &\sigma_r^2\left(\{X_{r} + Z\} V_{r}\right) = \sigma_r^2\left(\{X_{r} + Z\} V_{r} M M^{-1}\right) \ge \sigma_r^2 (\{X_{r} + Z\} V_{r} M) \sigma_{\min} ^2(M^{-1} ) 
        \nonumber\\
        = &  \sigma_r^2\left(\{X_{r} + Z\} V_{r} M\right)
        \{\sigma^2_r(X) + n\kappa^2 \}
        \nonumber\\
        = &  \sigma_r\big(M^{\top} V_{r}^{\top} \{X_{r} + Z\}^{\top} \{X_{r} + Z\} V_{r} M\big)
        \{ \sigma^2_r(X) + n\kappa^2 \}  \label{eq:singular value term 1}
    \end{align}
    where the  inequality follows from \Cref{lemma:singular values of matrix product lower bound}.    
    \\
    \\
    Consider the term $ \sigma_r\big(M^{\top} V_{r}^{\top} \{X_{r} + Z\}^{\top} \{X_{r} + Z\} V_{r} M\big)$.   Note that 
    \begin{align}\nonumber
        &M^{\top} V_{r}^{\top} \left(X_{r} + Z \right)^{\top} \left(X_{r} + Z \right) V_{r} M - I_r
        \\\nonumber
        =\,&M^{\top} V_{r}^{\top} \left(X_{r} + Z \right)^{\top} \left(X_{r} + Z \right) V_{r} M - \mathbb{E}\left[M^{\top} V_{r}^{\top} \left(X_{r} + Z \right)^{\top} \left(X_{r} + Z \right) V_{r} M \right]
        \\\nonumber
        =\,& \underbrace{M^\top V_{r}^{\top} X_{r}^\top X_{r} V_{r} M - \mathbb{E}\left[ M^\top V_{r}^{\top} X_{r}^\top X_{r} V_{r} M \right] }_{{=0}} \, + \, M^{\top} V_{r}^{\top} X_{r}^{\top} Z V_{r} M
        \\\nonumber
        & \, + \, M^{\top} V_{r}^{\top} Z^{\top} X_{r} V_{r} M \, + \, M^\top V_{r}^{\top} Z^\top Z V_{r} M - \mathbb{E}\left[ M^\top V_{r}^{\top} Z^\top Z V_{r} M  \right]\\\nonumber
        & \, + \, \underbrace{\mathbb{E}\left[M^{\top} V_{r}^{\top} X_{r}^{\top} Z V_{r} M \right]}_{=0} \, + \, \underbrace{\mathbb{E}\left[M^{\top} V_{r}^{\top} Z^{\top} X_{r} V_{r} M \right]}_{=0}
        \\\label{eq:step1_op}
        = \, & M^{\top} V_{r}^{\top} X_{r}^{\top} Z V_{r} M  \, + \, M^{\top} V_{r}^{\top} Z^{\top} X_{r} V_{r} M \, + \, M^{\top} V_{r}^{\top} \left(Z^{\top}Z -  n\kappa^2I_m\right) V_{r} M.
    \end{align}
      Since, 
    $$
    \left\| X_{r} V_{r} M \right\|^2 = \max_{k=1,\ldots, r} \frac{\sigma_k^2(X)}{\sigma_k^2(X) + n \kappa^2} \le 1, \qquad \text{and} \qquad \left\| V_{r} M \right\|^2 = \left\| M \right\|^2 = \frac{1}{\sigma_r^2(X) + n \kappa^2}, 
    $$
    it follows by Lemma \ref{lemma:SG matrix operator}  that
    \begin{equation}\label{eq: step1e1}
        \mathbb{P}\Big( \left\| M^{\top} V_{r}^{\top} X_{r}^{\top} Z V_{r} M\right\| \ge x \Big) \le 2\, \exp \left( Cr - c x^2\, \frac{\sigma_r^2(X) + n \kappa^2}{\kappa^2} \right).
    \end{equation}
    Similarly,  \Cref{lemma:covariance operator bound} implies that 
    \begin{equation}\label{eq: step1e2}
        \begin{split}
            &\mathbb{P}\Big( \left\| M^{\top} V_{r}^{\top} \left(Z^{\top}Z -  n\kappa^2I_m\right) V_{r} M \right\| \ge x \Big) 
            \\
            &\le 2\, \exp \left( Cr - c \min\left\{ x^2\, \frac{\left(\sigma_r^2(X) + n \kappa^2 \right)^2}{n\kappa^4},\, x\,\frac{\sigma_r^2(X) + n \kappa^2}{\kappa^2} \right\} \right),
            \\
            & \le 2\, \exp \left( Cr - c\, \frac{\sigma_r^2(X) + n \kappa^2}{\kappa^2} \min\left\{ x^2,\, x \right\} \right),
        \end{split}
    \end{equation}
    where the last inequality  follows from the fact that $\frac{ \{\sigma_r^2(X) + n{\kappa}^2\}^{2}}{\kappa^4n} \geq \frac{  \sigma_r^2(X) + n{\kappa}^2 }{\kappa^2}$. Thus, combining \eqref{eq:step1_op}, \eqref{eq: step1e1}, and \eqref{eq: step1e2}, we have
\begin{align}
\mathbb{P}\!\left( \big\| M^{\top} V_{r}^{\top} (X_{r}+Z)^{\top} (X_{r}+Z) V_{r} M - I_r \big\| \ge x \right)
\le 6 \exp\!\left( C r - c \frac{\sigma_r^2(X)+n\kappa^2}{\kappa^2}\,\min\{x^2,x\} \right).
\label{eq:Step1_op_3_1}
\end{align}
Here we apply a union bound to the three terms in \eqref{eq:step1_op} with thresholds $x/3$ each. 
The same tail bound as in \eqref{eq: step1e1} holds for $M^{\top} V_{r}^{\top} Z^{\top} X_{r} V_{r} M$ by symmetry, 
and \eqref{eq: step1e2} controls the centered quadratic term $M^{\top} V_{r}^{\top} (Z^{\top}Z - n\kappa^2 I_m) V_{r} M$.
All absolute constants are absorbed into $C,c$.
 This implies that 
        \begin{align}\nonumber 
         \mathbb P \bigg(  \sigma_r\big(M^{\top} V_{r}^{\top} \{X_{r} + Z\}^{\top} \{X_{r} + Z\} V_{r} M\big)  \ge 1-x  \bigg) 
         \\ \label{eq:Step1_op_3_2}
         \ge 1- 6  \exp \left( Cr - c\, \frac{\sigma_r^2(X) + n \kappa^2}{\kappa^2} \min\left\{ x^2,\, x \right\} \right),
    \end{align}
       \eqref{eq:Step1_op_3_2} and \eqref{eq:singular value term 1} together imply that 
    \begin{equation}\nonumber
        \mathbb{P}\Big(\sigma_{r}^2(YV_{r}) \ge    \{ \sigma_{r}^2(X) + n\kappa^2 \} (1 - x)\Big) \ge 1- 6\, \exp \left( Cr - c\, \frac{\sigma_r^2(X) + n \kappa^2}{\kappa^2} \min\left\{ x^2,\, x \right\} \right).
    \end{equation}
\
\\
    Setting $x = \tfrac{1}{6}\,\tfrac{\sigma_r^2(X) - \sigma_{r+1}^2(X)}{\sigma_r^2(X)+n\kappa^2}$, we have
    \begin{align}\label{eq:tail_prob1}
        &\mathbb{P}\left(\sigma_{r}^2(YV_{r}) \ge \sigma_{r}^2(X) + n\kappa^2 - \frac{\sigma_r^2(X) - \sigma_{r+1}^2(X)}{6}\right)\nonumber
        \\
        &\ge 1- 6\exp\left(Cr - c\min\left\{ \frac{1}{36 \kappa^2}\, \frac{\left(\sigma_r^2(X) - \sigma_{r+1}^2(X)\right)^2}{ \sigma^2_r(X) + n\kappa^2 },\, \frac{\sigma_r^2(X) - \sigma_{r+1}^2(X)}{6\kappa^2}\right\}\right).
    \end{align}
\\
\textbf{Step 2}. We upper bound the term $\sigma_{r+1}^2(Y)$. Note that
\begin{align*}
    \sigma_{r +1}(Y) = \min_{\rank(B) \leq r}\|Y - B\| \leq \left\|Y - Y\cdot V_{r}V_{r}^{\top}\right\| = \sigma_{\max}(YV_{\perp}).
\end{align*}
Moreover,
\begin{align}\nonumber
    \sigma_{\max}^2(YV_{\perp}) &= \left\|V_{\perp}^{\top}Y^{\top}YV_{\perp}\right\| = \left\|V_{\perp}^{\top}(X_{r} + X_{\perp} + Z)^{\top}(X_{r} + X_{\perp} + Z)V_{\perp}\right\|
    \\\nonumber
    &= \left\|V_{\perp}^{\top}(X_{\perp} + Z)^{\top}(X_{\perp} + Z)V_{\perp}\right\|
    \\\nonumber
    &\leq  \left\|V_{\perp}^{\top}Z^{\top}ZV_{\perp}\right\| + \left\|V_{\perp}^{\top}Z^{\top}X_{\perp}V_{\perp}\right\| + \left\|V_{\perp}^{\top}X_{\perp}^{\top}ZV_{\perp}\right\| + \left\|V_{\perp}^{\top}X_{\perp}^{\top}X_{\perp}V_{\perp}\right\|
    \\\label{eq:step2e0}
    & =  \underbrace{\left\|V_{\perp}^{\top}Z^{\top}ZV_{\perp}\right\|}_{\mathrm{I}_1}  + 2 \underbrace{\left\|V_{\perp}^{\top}Z^{\top}X_{\perp}V_{\perp}\right\|}_{\mathrm{I}_2} + \underbrace{\left\|V_{\perp}^{\top}X_{\perp}^{\top}X_{\perp}V_{\perp}\right\|}_{\mathrm{I}_3}.
\end{align}
For the term $\mathrm{I}_3$, we have
\begin{align}\label{eq:term_I4}
    \mathrm{I}_3 = \sigma^2_{1}(X_{\perp}V_{\perp}) = \sigma^2_{1}(U_{\perp}\Sigma_{\perp}V_{\perp}^{\top}V_{\perp}) = \sigma^2_{1}(\Sigma_{\perp}) = \sigma^2_{r+1}(X).
\end{align}
For $\mathrm{I}_2$, note that 
    $$
    \left\| X_{\perp}V_{\perp} \right\| ^2  = \left\| U_\perp \Sigma_\perp \right\| ^2 = \sigma_{r+1}^2(X), \qquad \text{and} \qquad \left\| V_\perp \right\|^2 = 1. 
    $$
    it follows  from \Cref{lemma:SG matrix operator} that 
    \begin{equation}\label{eq: step2e1}
        \mathbb{P}\Big( \mathrm{I}_2 \ge x \Big) \le 2\, \exp \left( C_1m -  \frac{c_1 x^2}{\kappa^2\, \sigma_{r+1}^2(X)} \right).
    \end{equation}
   For $\mathrm{I}_1$,  note that by  \Cref{lemma:covariance operator bound} and the fac that $V_\perp^\top V_\perp =I_{m-r} $, we have 
    \begin{equation}\label{eq: step2e2}
        \begin{split}
            &\mathbb{P}\Big( \left\|V_{\perp}^{\top}Z^{\top}ZV_{\perp} - n \kappa^2 I_{m-r} \right\|  \ge t \Big) \le 2\, \exp \left( C_2m - c_2 \min\left\{  \frac{t^2}{n\,\kappa^4 },\, \frac{t}{\kappa^2} \right\} \right),
            \\
            \implies & \mathbb{P}\Big( \mathrm{I}_1 \ge n\kappa^2(1+t) \Big) \le 2\, \exp \left( C_2m - c_2n \min\left\{  t^2,\, t \right\} \right).
        \end{split}
    \end{equation}
    Combining the calculations in this step,    with $x = \kappa\sigma_{r+1}(X)\sqrt{\tfrac{2C_1}{c_1}m }$ in \eqref{eq: step2e1}, and with $t = \tfrac{\sigma_r^2(X) - \sigma_{r+1}^2(X)}{6n\kappa^2}$ in \eqref{eq: step2e2}, it follows that 
    \begin{equation}\label{eq:tail_prob2}
        \begin{split}
          &\mathbb{P}\left( \sigma^2_{r+1}(Y) \ge n\kappa^2 + \frac{\sigma_r^2(X) - \sigma_{r+1}^2(X)}{6} + \kappa\sigma_{r+1}(X)\sqrt{\frac{2C_1}{c_1}m\,} + \sigma_{r+1}^2(X)\right) 
          \\
          & \quad \le 2\,\exp\left(C_2m - c_2\,\min\!\left\{\,\frac{\left(\sigma_{r}^{2}(X) -\sigma_{r+1}^{2}(X)\right)^{2}}{36\,\kappa^{4}\,n},\frac{\sigma_{r}^{2}(X)-\sigma_{r+1}^{2}(X)}{6\,\kappa^{2}}\right\}\right) + 2\,\exp\bigl(-C_3\,m\bigr).
        \end{split}
    \end{equation}
\
\textbf{Step~3}. Recall we define  
\[
M = \diag\left((\sigma_1^2 + n\kappa^2)^{-1/2}, \dots, (\sigma_r^2 + n\kappa^2)^{-1/2}\right) \in \mathbb R^{r\times r}.
\]
We have
\begin{align}\label{eq:term_project}
\left\|\mathcal{P}_{YV_{r}}YV_{\perp}\right\| &= \left\|\mathcal{P}_{YV_{r}M}YV_{\perp}\right\|\nonumber\\
&= \left\|(YV_{r}M)\left((YV_{r}M)^{\top}(YV_{r}M)\right)^{-1}(YV_{r}M)^{\top}YV_{\perp}\right\|\nonumber\\
&\leq \left\|(YV_{r}M)\left((YV_{r}M)^{\top}(YV_{r}M)\right)^{-1}\right\|\left\|M^{\top}V_{r}^{\top}Y^{\top}YV_{\perp}\right\|\nonumber\\
&\leq \sigma_{\min}^{-1}(YV_{r}M)\left\|M^{\top}V_{r}^{\top}Y^{\top}YV_{\perp}\right\| = \sigma_{r}^{-1}(YV_{r}M)\left\|M^{\top}V_{r}^{\top}Y^{\top}YV_{\perp}\right\|,
\end{align}
where the first equality follows from the fact that $YV_{r}$ and $YV_{r}M$ have the same column spaces (since $M$ is invertible), the last inequality follows from Lemma \ref{eq:regression operator}, 
and for the last equality we use that the singular values of $YV_{r}M$ are in nonincreasing order so that its smallest singular value equals $\sigma_{r}(YV_{r}M)$.
\\
\\
By \eqref{eq:Step1_op_3_1}, we have for every \(x>0\)
\[
\mathbb{P}\Bigl(
\bigl\|M^{\top}V_{r}^{\top}(X_{r}+Z)^{\top}(X_{r}+Z)V_{r}M-I_{r}\bigr\|\ge x
\Bigr)
\le 6\exp\!\left(Cr -c\,\frac{\sigma_r^2(X)+n\kappa^2}{\kappa^2}\min\{x^2,x\}\right).
\]
Taking \(x=\frac{1}{2}\) gives
\begin{align}\label{eq:term_sigular_min}
\mathbb{P}\Bigl(\bigl\|M^{\top}V_{r}^{\top}(X_{r}+Z)^{\top}(X_{r}+Z)V_{r}M-I_{r}\bigr\|<\tfrac{1}{2}\Bigr)
\ge 1-6\exp\!\left(Cr -c\,\frac{\sigma_r^2(X)+n\kappa^2}{4\kappa^2}\right).
\end{align}
In particular, on this event all eigenvalues of 
\(M^{\top}V_{r}^{\top}(X_{r}+Z)^{\top}(X_{r}+Z)V_{r}M\) are at least \(1/2\),
so \(\sigma_{r}^2(YV_{r}M)\ge 1/2\) with the same probability bound.
Consider$\left\|M^{\top}V_{r}^{\top}Y^{\top}YV_{\perp}\right\|$.   Since $ V_r^\top X_\perp^\top =0$, $ X_r V_\perp = 0  $ and $X_\perp^\top X_r =0 $, it follows that 
\begin{align*}
    M^{\top}V_{r}^{\top}Y^{\top}YV_{\perp} &= M^{\top}V_{r}^{\top} \left(X_{r} + X_{\perp} + Z \right)^{\top} \left(X_{r} + X_{\perp} + Z \right) V_{\perp}
    \\
    &= M^{\top}V_{r}^{\top}X_{r}^{\top}Z V_{\perp} + M^{\top}V_{r}^{\top} Z^\top X_\perp V_{\perp} + M^{\top}V_{r}^{\top} Z^\top Z V_{\perp} 
      \\
    &= M^{\top}V_{r}^{\top}X_{r}^{\top}Z V_{\perp} + M^{\top}V_{r}^{\top} Z^\top X_\perp V_{\perp} + M^{\top}V_{r}^{\top} Z^\top Z V_{\perp} - \underbrace{M^{\top}V_{r}^{\top} \left(n \kappa^2 I_m \right) V_{\perp}}_{=0},
\end{align*}
    Since, 
    $$
    \left\| X_{r} V_{r} M \right\| ^2 \le 1,\, \left\| V_{r} M \right\| ^2=  \frac{1}{\sigma_r^2(X) + n \kappa^2}  \text{ and } \left\| X_\perp V_\perp\right\|  ^2 = \left\| X_\perp\right\| ^2 = \sigma^2_{r+1}(X),
    $$
    it follows  from \Cref{lemma:SG matrix operator} that
    \begin{equation}\label{eq: step3e1}
        \begin{split}
            \mathbb{P}\Big( \left\| M^{\top} V_{r}^{\top} X_{r}^{\top} Z V_\perp \right\|  \ge x \Big) &\le 2\, \exp \left( Cm - \frac{c x^2}{\kappa^2} \right),
            \\
            \mathbb{P}\Big( \left\|  M^{\top}V_{r}^{\top} Z^\top X_\perp V_{\perp} \right\|  \ge x \Big) &\le 2\, \exp \left( Cm - \frac{c x^2}{\kappa^2}\, \frac{\sigma_r^2(X) + n \kappa^2}{\sigma_{r+1}^2(X)} \right) \le  2\, \exp \left( Cm - \frac{c x^2}{\kappa^2} \right).
        \end{split}
    \end{equation}
    Similarly,  \Cref{lemma:covariance operator bound} implies that 
    \begin{equation}\label{eq: step3e2}
        \begin{split}
            &\mathbb{P}\Big( \left\| M^{\top} V_{r}^{\top} \left(Z^{\top}Z -  n\kappa^2I_m\right) V_\perp \right\|  \ge x \Big) 
            \\
            &\le 2\, \exp \left( Cm  - c \min\left\{ x^2\, \frac{\sigma_r^2(X) + n \kappa^2}{n\kappa^4},\, x\,\frac{\sqrt{\sigma_ r^2(X) + n \kappa^2}}{\kappa^2} \right\} \right),
            \\
            & \le 2\, \exp \left( Cm - c \min\left\{ \frac{x^2}{\kappa^2},\, x\,\frac{\sqrt{\sigma_r^2(X) + n \kappa^2}}{\kappa^2} \right\} \right),
        \end{split}
    \end{equation}
    where the last inequality uses $\frac{ \sigma_R^2(X) + n{\kappa}^2}{ n\,\kappa^2} \geq 1$. Thus, combining \eqref{eq:term_project}, \eqref{eq:term_sigular_min}, \eqref{eq: step3e1} and \eqref{eq: step3e2} with $x = \kappa\sqrt{\frac{2C}{c} m\,}$, we have   \begin{align}\label{eq:tail_prob3}
\mathbb{P}\!\left(\left\|\mathcal{P}_{YV_{r}}YV_{\perp}\right\|^2 
\ge \frac{36C}{c}\, m \kappa^2 \right)
&\le 6\exp\!\left(Cr -c \frac{\sigma_r^2(X) + n\kappa^2}{4\kappa^2}\right)  + 4\exp(-Cm)\\
&\quad + 2\exp\!\left(Cm - c\min\!\left\{ \tfrac{2C}{c}\,m,\,
\sqrt{\tfrac{2C}{c}\,m\;\tfrac{\sigma_r^2(X) + n \kappa^2}{\kappa^2}}\right\}\right).\nonumber
\end{align}
\\
\textbf{Step~4}. Define the event
\begin{align*}
    \mathcal{E} = \Bigg\{&\sigma_{r}^2(YV_{r}) \geq \sigma_{r}^2(X) + n\kappa^2 - \frac{\sigma_r^2(X) - \sigma_{r+1}^2(X)}{6}\,;
    \\
    &\sigma_{r+1}^2(Y) \leq n\kappa^2 + \frac{\sigma_r^2(X) - \sigma_{r+1}^2(X)}{6} + \kappa\sigma_{r+1}(X)\sqrt{\frac{2C}{c}m\,} + \sigma_{r+1}^2(X) \,;
    \\
    &\left\|\mathcal{P}_{YV_{r}}YV_{\perp}\right\|^2 \leq \frac{8C}{c}\, m \kappa^2 \Bigg\}.
    \end{align*} It follows from \eqref{eq:tail_prob1}, \eqref{eq:tail_prob2} and \eqref{eq:tail_prob3} that by the union bound, 
\begin{align}\label{eq:union bound term 1}
    \mathbb{P}(\mathcal{E}^c) &\leq 6\exp\left(Cr - c\min\left\{ \frac{1}{36 \kappa^2}\, \frac{\left(\sigma_r^2(X) - \sigma_{r+1}^2(X)\right)^2}{ \sigma^2_r(X) + n\kappa^2 },\, \frac{\sigma_r^2(X) - \sigma_{r+1}^2(X)}{6\kappa^2}\right\}\right)
    \\
    &\quad+  2\,\exp\left(Cm \;-\;c\,\min\!\left\{\,\frac{\left(\sigma_{r}^{2}(X) -\sigma_{r+1}^{2}(X)\right)^{2}}{36\,\kappa^{4}\,n},\frac{\sigma_{r}^{2}(X) -\sigma_{r+1}^{2}(X)}{6\,\kappa^{2}}\right\}\right) + 2\,\exp\bigl(-C\,m\bigr)
    \\
    &\quad+ 6\exp\left(Cr -c \frac{\sigma_r^2(X) + n\kappa^2}{4\kappa^2}\right)  + 4\exp(-Cm)
    \\
    &\quad + 2\exp\left(Cm - c\min\left\{\frac{2C}{c}\,m,\, \sqrt{\frac{2C}{c}\,m \frac{\sigma_r^2(X) + n \kappa^2}{\kappa^2}}\right\}\right).
\end{align}
In what follows, we show that under the SNR assumption 
$$
\Big(\sigma_r (X) - \sigma_{r+1} (X)\Big)^2   \geq  C_{\mathrm{gap}}\kappa^2\Big(\sqrt{nm} + m\Big)
$$
with sufficient large absolute constant $C_{\mathrm{gap}} > 0$, we have
\begin{align*}
\mathbb{P}(\mathcal{E}^c) &\leq C\exp\left(-Cm\right),
\end{align*} 
where \(C > 0\) is an absolute constant, appropriately scaled to absorb the other constants.

We illustrate how to bound \eqref{eq:union bound term 1}, as the rest of the terms can be bounded in a similar and simpler way.
Note that 
\begin{align*}
   \frac{1}{36 \kappa^2}\, \frac{\left(\sigma_r^2(X) - \sigma_{r+1}^2(X)\right)^2}{ \sigma^2_r(X) + n\kappa^2 } &\ge \frac{\left(\sigma_r^2(X) - \sigma_{r+1}^2(X)\right)^2}{72 \kappa^2} \min\left\{\frac{1}{\sigma_r^2(X)},\, \frac{1}{n\kappa^2} \right\}
    \\
   & \ge  \min\left\{\frac{\left(\sigma_r^2(X) - \sigma_{r+1}^2(X)\right)^2}{72\,\kappa^2 \sigma_r^2(X)},\, \frac{\left(\sigma_r^2(X) - \sigma_{r+1}^2(X)\right)^2}{72 n\kappa^4} \right\}.
\end{align*}
We have 
\begin{align*}
    \frac{\big(\sigma_r^2(X) - \sigma_{r+1}^2(X)\big)^2}{72\,\kappa^2 \sigma_r^2(X)} = \frac{\big(\sigma_r(X) + \sigma_{r+1}(X)\big)^2\, \big(\sigma_r(X) - \sigma_{r+1}(X)\big)^2}{72\,\kappa^2 \sigma_r^2(X)} \ge \frac{\big(\sigma_r(X) - \sigma_{r+1}(X)\big)^2}{72 \, \kappa^2} \ge \frac{2C}{c} m ,
\\
    \frac{\left(\sigma_{r}^{2}(X) -\sigma_{r+1}^{2}(X)\right)^{2}}{72\,\kappa^{4}\,n}
     \geq \frac{\left(\sigma_{r}(X) -\sigma_{r+1}(X)\right)^{4}}{72\,\kappa^{4}\,n} \geq \frac{2C}{c}m.
\end{align*}
So 
$$ \frac{1}{36 \kappa^2}\, \frac{\left(\sigma_r^2(X) - \sigma_{r+1}^2(X)\right)^2}{ \sigma^2_r(X) + n\kappa^2 }  \ge \frac{2C}{c}m .$$
In addition,
\begin{align*}
    \frac{\sigma_R^2(X) - \sigma_{R+1}^2(X)}{6 \kappa^2} \ge \frac{\big(\sigma_R^2(X) - \sigma_{R+1}^2(X) \big)^2}{72\,\kappa^2 \sigma_R^2(X)} \ge \frac{2C}{c} m.
\end{align*}
So
$  
\eqref{eq:union bound term 1}  \leq C\exp\left(-C m\right),
$  
for some absolute constant $C>0$.
\\
\\
{\bf Step 5.} Under the event $\mathcal{E}$, by \Cref{prop:sintheta}, we have that
\begin{align*}
    \left\|\sin\Theta\left(\widehat{V}_{r}, V_{r}\right) \right\|^2 &\leq \frac{\sigma_r^2(YV_{r})\left\|\mathcal{P}_{YV_{r}}YV_{\perp}\right\|^2}{\left(\sigma_r^2(YV_{r}) - \sigma_{r+1}^2(Y)\right)^2}\leq C_7\frac{\sigma_r^2(YV_{r}) m\kappa^2 }{\left(\sigma_r^2(YV_{r}) - \sigma_{r+1}^2(Y)\right)^2} \nonumber\\
    &\leq C_8\frac{\left(\sigma_{r}^2(X) + n\kappa^2 - \frac{\sigma_r^2(X) - \sigma_{r+1}^2(X)}{6}\right) m\kappa^2 }{\left(\sigma_{r}^2(X) + n\kappa^2 - \frac{\sigma_r^2(X) - \sigma_{r+1}^2(X)}{6} - \sigma_{r+1}^2(Y)\right)^2}\nonumber\\
    &\leq C_8\frac{\left(\sigma_{r}^2(X) + n\kappa^2 - \frac{\sigma_r^2(X) - \sigma_{r+1}^2(X)}{6} \right)m\kappa^2 }{\left( (1- \frac{1}{3})\left(\sigma_r^2(X) - \sigma_{r+1}^2(X) \right) - \kappa\sigma_{r+1}(X)\sqrt{\frac{2C}{c}m\,}\right)^2}\nonumber\\
    &\leq C_9\frac{\left(\sigma_{r}^2(X) + n\kappa^2 - \frac{\sigma_r^2(X) - \sigma_{r+1}^2(X)}{6}\right)m\kappa^2}{ (1- \frac{1}{2})^2\big( \sigma_r^2(X) - \sigma_{r+1}^2(X)\big)^2}\nonumber\\
    &\leq C_{10}\frac{\left(\sigma_{r}^2(X) + n\kappa^2 - \frac{\sigma_r^2(X) - \sigma_{r+1}^2(X)}{6}\right)m\kappa^2}{\big(\sigma_r^2(X) - \sigma_{r+1}^2(X)\big)^2},
\end{align*}
Here, the third inequality follows from the fact that $x^2/(x^2 - y^2)^2$ is a decreasing function of $x$ and an increasing function of $y$ when $x > y \geq 0$, together with the fact that the event $\mathcal{E}$ holds. The fifth inequality follows from the fact that, under the assumption $\Big(\sigma_r(X) - \sigma_{r+1}(X)\Big)^2 \geq C_{\mathrm{gap}}\kappa^2(\sqrt{nm} + m)$ with $C_{\mathrm{gap}} > 0$ being large enough,
\begin{align*}
    \frac{\Big( \sigma_r^2(X) - \sigma_{r+1}^2(X)\Big)^
2}{36} =&
\Big( \sigma_r (X) + \sigma_{r+1} (X)\Big)^
2\frac{\Big( \sigma_r (X) - \sigma_{r+1} (X)\Big)^
2}{36}
\\
\ge &
\sigma_{r }^2(X) \frac{ C _{\mathrm{gap} } m \kappa^2}{36}  \geq Cm\kappa^2\sigma_{r}^2(X) \geq Cm\kappa^2\sigma_{r+1}^2(X).
\end{align*}

Therefore, with probability at least $1 - C\exp(-Cm)$,
\begin{align*}
    \left\|\sin\Theta\left(\widehat{V}_{r}, V_{r}\right) \right\| ^2 
    &\leq C_3\frac{ \left(\sigma_{r}^2(X) + n\kappa^2 - \frac{\sigma_r^2(X) - \sigma_{r+1}^2(X)}{6}\right)m\kappa^2 }{( \sigma_r^2(X) - \sigma_{r+1}^2(X))^2}\\
   & \leq  C_3\frac{ \left(\sigma_{r}^2(X) + n\kappa^2  \right)m\kappa^2 }{ ( \sigma_r^2(X) - \sigma_{r+1}^2(X))^2} = C_3\frac{  \sigma_{r}^2(X) m\kappa^2    }{ ( \sigma_r^2(X) - \sigma_{r+1}^2(X))^2} + C_3\frac{      nm\kappa^4    }{ ( \sigma_r^2(X) - \sigma_{r+1}^2(X))^2}  
   \\
    &\leq C_4 \bigg\{  \frac{m\kappa^2 }{(\sigma_r (X) - \sigma_{r+1} (X)  ) ^2 } +  \frac{\kappa^4 nm }{(\sigma_r (X) - \sigma_{r+1} (X)  ) ^4 } \bigg\} ,
\end{align*}
where the third inequality follows from the observation that 
$$ \frac{\sigma_r^2(X) }{ (\sigma_r^2(X) - \sigma_{r+1}^2(X)  ) ^2 } = \frac{\sigma_r^2(X) }{  (\sigma_r (X) + \sigma_{r+1} (X)  )^2  (\sigma_r (X) - \sigma_{r+1} (X)  ) ^2  }  \le  \frac{1}{(\sigma_r (X) - \sigma_{r+1} (X)  ) ^2 } ,$$
and last display follows from
\[
\sigma_r^2(X) - \sigma_{r+1}^2(X) = ( \sigma_r (X) + \sigma_{r+1}(X))( \sigma_r (X) - \sigma_{r+1} (X) ) \geq \Big(\sigma_r(X) - \sigma_{r+1}(X)\Big)^2.
\] 
\end{proof}

\begin{corollary}\label{corollary:initial sintheta} 
Suppose the conditions of \Cref{thm:main} hold, in particular the 
condition in Equation (\ref{eq:SNR}) for each mode-$k$ unfolding $\mathcal M_k(X^*)$ with target rank $r_k$.
Let $U_k^{(0)}$ be the matrix of top $r_k$ left singular vectors of the mode-$k$ unfolding $Y^{(k)} =\mathcal M_k(Y)$.
Then for each $k\in\{1,2,3\}$, with probability at least $1 - C_1\exp(-C_2 p_k)$,
\[
\bigl\|\sin\Theta\!\left(U_k^{(0)},\,U_k^*\right)\bigr\| \;\le\; \frac{1}{2\sqrt{r_{\max}}}\,.
\]
\end{corollary}

\begin{proof}
Apply \Cref{lemma:svd} to 
\(Y^{(k)}=\mathcal M_k(Y)\in\mathbb R^{p_k\times p_{\neq k}}\)
with signal \(X^{(k)}=\mathcal M_k(X^*)\) and noise \(Z^{(k)}=\mathcal M_k(Z)\), where \(p_{\neq k}=\prod_{j\neq k}p_j\).
Use rank \(r=r_k\) and identify \(n =p_k\) (rows) and \(m=p_{\neq k}\) (columns).
The lemma (applied to left singular vectors, or equivalently to the transpose) gives
\[
\bigl\|\sin\Theta(U_k^{(0)},U_k^*)\bigr\|^2
\;\le\;
C\left\{\frac{p_k\,\kappa^2}{\Delta_k^2}
+\frac{\kappa^4\,p_k\,p_{\neq k}}{\Delta_k^4}\right\},
\]
where 
\(\Delta_k =\sigma_{r_k}\!\bigl(\mathcal M_k(X^*)\bigr)
-\sigma_{r_k+1}\!\bigl(\mathcal M_k(X^*)\bigr)\)
is the mode-$k$ spectral gap.

By the assumption \eqref{eq:SNR} in \Cref{thm:main} (applied to mode $k$),
\[
\Delta_k^2 \;\ge\; 
C_{\mathrm{gap}}\kappa^2\!\left(\sqrt{p_k\,p_{\neq k}\,r_{\max}} + r_{\max}\sum_{j=1}^3 p_j\right).
\]
Choosing \(C_{\mathrm{gap}}\) sufficiently large makes the right–hand side above at most \(1/(4r_{\max})\), hence 
\[
\bigl\|\sin\Theta(U_k^{(0)},U_k^*)\bigr\| \;\le\; \frac{1}{2\sqrt{r_{\max}}}\,.
\]
The probability bound \(1 - C_1\exp(-C_2p_k)\) matches the row dimension in the matrix lemma.
\end{proof}

\section{Auxiliary results}

\subsection{Deviations bounds}
\begin{lemma} [Theorem~4.4.3 in \cite{vershynin2018high}]\label{lemma:concentration_Z_mat}
    Assume all the entries of $Z \in \mathbb{R}^{m \times n}$ are independent mean-zero sub-Gaussian random variables, i.e.
    \[
    \|Z_{ij}\|_{\psi_2} = \sup_{q \geq 1}\mathbb{E}(|Z_{ij}|^q)^{1/q}/q^{1/2} \leq \kappa.
    \]
    Then there exist some universal constant $C > 0$, such that
    \[
    \|Z\| \leq C\kappa\left(\sqrt{m} + \sqrt{n} \right)
    \]
    with probability at least $1 - \exp(-(m+n)))$.
\end{lemma}

\begin{lemma} \label{lemma:concentration_Z}
    Suppose  all the entries of $Z \in \mathbb{R}^{p_1 \times p_2 
    \times p_3}$ are independent mean-zero sub-Gaussian random variables, i.e.
    \[
    \|Z_{ijk}\|_{\psi_2} = \sup_{q \geq 1}\mathbb{E}(|Z_{ijk}|^q)^{1/q}/q^{1/2} \leq \kappa.
    \]
    Then there exist some universal constants $C,c > 0$, such that
    \[
    \sup_{\substack{A \in \mathbb{R}^{p_1 \times p_2 
    \times p_3}, \\ \|A\|_{\mathrm{F}} \leq 1,  A \in \mathcal T_{ (r_1, r_2, r_3)}} }\langle Z, A \rangle \leq C\kappa\left(r_1r_2r_3 + \sum_{k = 1}^3p_kr_k\right)^{1/2}
    \]
    with probability at least $1 - \exp(-c\sum_{k = 1}^3p_kr_k)$.
\end{lemma}
\begin{proof}
   This directly follows from   Lemma~E.5 in \cite{han2022optimal}.
\end{proof}

\begin{lemma} \label{lemma:projection operator}
 Suppose  all the entries of $Z \in \mathbb{R}^{p_1 \times p_2 
    \times p_3}$ are independent mean-zero sub-Gaussian random variables, i.e.
    \[
    \|Z_{ijk}\|_{\psi_2} = \sup_{q \geq 1}\mathbb{E}(|Z_{ijk}|^q)^{1/q}/q^{1/2} \leq \kappa.
    \] 
    Let $W_2\in \mathbb O^{p_2\times s_2}$ and $W_3\in \mathbb O^{p_3\times s_3}$ be non-random. 
    Then   there exists absolute positive constants $C_1,C_2$ and $c$ such that 
\begin{align*}
       \mathbb P \bigg(  
    \big\| \mathcal{M}_1(Z)\,(W_{2} \otimes W_{3}) \big\| 
     \ge  
    C_1\kappa  (\sqrt{p_1+  s_{2}s_{3}   }  ) \bigg)
    \le C_2\exp \left(-c  p_ 1     \right) ,
\end{align*}    
\end{lemma}
  \begin{proof}
      It suffices to observe that 
      $ W_{2} \otimes W_{3} \in \mathbb O^{p_2p_3\times r_2r_3}$. The desired result is a direct consequence of \Cref{lemma:SG matrix operator}.
  \end{proof}
 
\begin{lemma} \label{lemma:concentration_Z_2}
 Suppose  all the entries of $Z \in \mathbb{R}^{p_1 \times p_2 
    \times p_3}$ are independent mean-zero sub-Gaussian random variables, i.e.
    \[
    \|Z_{ijk}\|_{\psi_2} = \sup_{q \geq 1}\mathbb{E}(|Z_{ijk}|^q)^{1/q}/q^{1/2} \leq \kappa.
    \]  Then   there exists absolute positive constants $C_1,C_2$ and $c$ such that 
\begin{align*}
      &\mathbb P \bigg( \sup_{\substack{V_{2} \in \mathbb{R}^{p_{2} \times r_{2}},\,\|V_{2}\|  \le 1\\
    V_{3} \in \mathbb{R}^{p_{3} \times r_{3}},\, \|V_{3}\|  \leq 1}}
    \big\| \mathcal{M}_1(Z)\,(V_{2} \otimes V_{3}) \big\| 
     \ge  
    C_1\kappa\Big(\sqrt{p_1+  r_{2}r_{3} +p_{ 2}r_{2}+ p_{3}r_{3} } \Bigg) 
      \\
      &\le C_2\exp \left(-c (p_ 1 +p_2+p_3  )\right) ,
\end{align*}    
  
\end{lemma}

\begin{proof} It follows from the assumption that  $ 
 \left(\mathcal{M}_1(Z)\right)_{i,j} \overset{i.i.d}{\sim} \mathrm{subGaussian}(0,\kappa^2)$. 
\
\\
\textbf{Step 1}. For fixed  $V_{2} \in \mathbb{R}^{p_{2} \times r_{2}} $ with $\|V_{2}\|  \le 1 $ and $
    V_{3} \in \mathbb{R}^{p_{3} \times r_{3}}$ with $ \|V_{3}\|  \leq 1    $,  it follows that  $$    \left\|V_2 \otimes V_3 \right\|  = \left\|V_2 \right\|  \left\|V_3 \right\|  \le 1,  
$$  We upper bound $\|\mathcal{M}_1(Z) \cdot (V_2 \otimes V_3)\|$ for any fixed $V_2$ and $V_3$. Since,
$$
\mathcal{M}(Z) \in \mathbb{R}^{p_1 \times p_2p_3} ,\left(\mathcal{M}(Z)\right)_{i,j} \overset{i.i.d}{\sim} \mathrm{subGaussian}(0,\kappa^2), \quad \text{and} \quad  \left\|V_2 \otimes V_3 \right\| = \left\|V_2 \right\|\left\|V_3 \right\| = 1.  
$$
It follows from \Cref{lemma:SG matrix operator} that  
\begin{align*}
    &\mathbb{P}\Big(\left\|\mathcal{M}_1(Z) \cdot (V_2\otimes V_3)\right\| > x \Big) \leq 2\exp\left(C(p_1 + r_2r_3) - c x^2\kappa^{-2}\right).
\end{align*}
\
\\
\textbf{Step 2}.
Let $\mathcal N_{p_2,r_2} ( \epsilon)$ denote an $\epsilon$ net of the set 
$$ \{A\in \mathbb R^{p_2\times r_2}: \|A\|\le 1 \}$$
with respect to the operator norm $\| \cdot \|$.   Similarly, let 
Let $\mathcal N_{p_3,r_3} ( \epsilon)$ denote an $\epsilon$ net of the set 
$$ \{B\in \mathbb R^{p_3\times r_3}: \|B\|\le 1 \}$$
with respect to the operator norm $\| \cdot \|$.
Denote the random quantity 
$$
\psi= \sup_{\substack{V_{2} \in \mathbb{R}^{p_{2} \times r_{2}},\,\|V_{2}\|  \le 1\\
    V_{3} \in \mathbb{R}^{p_{3} \times r_{3}},\, \|V_{3}\|  \leq 1}}
    \big\| \mathcal{M}_1(Z)\,(V_{2} \otimes V_{3}) \big\|.
$$
For any given $V_2\in \mathbb R^{p_2\times r_2}$ and $V_3\in \mathbb R^q$ with $\|V_2\|\le1 $ and $\|V_3\|\le 1$, let $\widetilde V_2 \in \mathcal N_{p_2,r_2}(1/4)  $ and $\widetilde V_3 \in \mathcal N_{p_3,r_3}(1/4)  $ be such that 
        $$\| V_2- \widetilde V_2 \|  \le 1/4 \quad \text{and} \quad \| V_3- \widetilde V_3 \| \le 1/4. $$
        Then
\begin{align*}
    \| \mathcal M_1(Z)  V_2\otimes V_3 \| \le \| \mathcal M_1(Z) ( V_2- \widetilde V_2)  \otimes V_3\| + 
    \|   \mathcal M_1(Z) \widetilde V_2\otimes ( V_3- \widetilde V_3)     \| + \| \mathcal M_1(Z) \widetilde V_2\otimes   \widetilde V_3 \|. 
\end{align*}
        Note that $\| V_2- \widetilde V_2 \|  \le 1/4   $. So
        $$  \| \mathcal M_1(Z) ( V_2- \widetilde V_2)  \otimes V_3\|   =\frac{1}{4}  \| \mathcal M_1(Z) \{  4( V_2- \widetilde V_2) \} \otimes V_3\|   \le \frac{\psi}{4}. $$
        Similarly
        $$ \| \mathcal M_1(Z) V_2\otimes ( V_3- \widetilde V_3)   \|    \le \frac{\psi}{4}  .$$
        In addition,
        $$ \| \mathcal M_1(Z) \widetilde V_2\otimes   \widetilde V_3 \|   \le \sup_{ V_2 \in \mathcal  N_{p_2,r_2} (1/4) , b\in \mathcal  N_{p_3,r_3}(1/4)}  \| \mathcal M_1(Z)   V_2\otimes    V_3\|  .$$
      So for any $V_2$ and  $V_3$,
\begin{align*}
    \| \mathcal M_1(Z)  V_2\otimes V_3 \|   \le  \frac{1}{2}\psi + \sup_{ V_2 \in \mathcal  N_{p_2,r_2} (1/4) , b\in \mathcal  N_{p_3,r_3}(1/4)}  \| \mathcal M_1(Z)   V_2\otimes    V_3\| . 
\end{align*}
      Taking sup over all   $V_2\in\{A\in \mathbb R^{p_2\times r_2}: \|A\|\le 1 \} $ and $V_3\in\{B\in \mathbb R^{p_3\times r_3}: \|B\|\le 1 \}$, it follows that 
      $$ \psi \le \frac{1}{2}\psi +\sup_{ V_2 \in \mathcal  N_{p_2,r_2} (1/4) , b\in \mathcal  N_{p_3,r_3}(1/4)}  \| \mathcal M_1(Z)   V_2\otimes    V_3\| ,$$
        or simply
        $$ \psi  \le 2\sup_{ V_2 \in \mathcal  N_{p_2,r_2} (1/4) , b\in \mathcal  N_{p_3,r_3}(1/4)}  \| \mathcal M_1(Z)   V_2\otimes    V_3\|  . $$
        {\bf Step 3.}  By Proposition 8 in \cite{pajor1998metric},  the cardinality of $\mathcal N_{p_2,r_2} ( \epsilon)$ is bounded $(\frac{C}{\epsilon})^{p_2r_2}$, and $\mathcal N_{p_3,r_3} ( \epsilon)$ is bounded $(\frac{C}{\epsilon})^{p_3r_3}$. Therefore 
        \begin{align*}
           \mathbb P  ( \psi \ge 2t)  \le  & \mathbb P\bigg(\sup_{ V_2 \in \mathcal  N_{p_2,r_2} (1/4) , b\in \mathcal  N_{p_3,r_3}(1/4)}  \| \mathcal M_1(Z)   V_2\otimes    V_3\|  \ge t  \bigg) 
           \\
           \le &
        C_2 ^{p_2r_2 }C_3 ^{p_3r_3 } \sup_{ V_2 \in \mathcal  N_{p_2,r_2} (1/4) , b\in \mathcal  N_{p_3,r_3}(1/4)}    \mathbb P\bigg( \| \mathcal M_1(Z)   V_2\otimes    V_3\|  \ge t  \bigg)  \\
        \le &   2\exp\left(C(p_1 + r_2r_3+p_2r_2+p_3r_3) - c t^2\kappa^{-2}\right).
        \end{align*} Here  $C$ and $C_3$ are positive constants.
        The desired result follows  by noting
        $$\psi= \sup_{\substack{V_{2} \in \mathbb{R}^{p_{2} \times r_{2}},\,\|V_{2}\|  \le 1\\
    V_{3} \in \mathbb{R}^{p_{3} \times r_{3}},\, \|V_{3}\|  \leq 1}}
    \big\| \mathcal{M}_1(Z)\,(V_{2} \otimes V_{3}) \big\| .$$

\end{proof}

\begin{lemma} \label{lemma:SG matrix operator}
    Suppose $Z \in \mathbb{R}^{n \times m}$, with $Z_{ij} \overset{i.i.d}{\sim} \mathrm{subGaussian}(0,\kappa^2)$. Let   $A \in \mathbb{R}^{p\times n} $ and $B \in \mathbb{R}^{m \times q}$  be non-random   matrices. Then for any $t >0$
    \begin{align}
        \label{eqref: SG b}\mathbb{P}\left( \left\|A  \, Z \, B \right\|   >   t    \right) \le C_1 \, \exp\left( C  (p+q)   -\frac{c\,t^2}{\kappa^2\, \left\|A \right\|^2 \, \left\|B \right\|^2} \right).
    \end{align}
\end{lemma}
\begin{proof} 

{\bf Step 1.}
Let $u\in \mathbb R^{n}$ and $v \in \mathbb R^{m}$ be non-random. Then
     that $u^\top \, Z \, v = \sum_{i=1}^n \sum_{j=1}^n u_i Z_{ij} v_j$.  Since $Z_{ij}$ are i.i.d. sub-Gaussian with parameter $\kappa^2$, it follows that 
     $ u^\top \, Z \, v  $ is  sub-Gaussian with parameter $\kappa^2 \|u\|_2^2 \|v\|_2^2$. Consequently  by Hoeffding’s inequality,
     $$\mathbb{P}\left( \left|u^\top \, Z \, v \right| \, > t \,  \right) \le 2\, \exp\left(-\frac{c\,t^2}{\kappa^2 \, \left\|u \right\|_2^2 \, \left\|v \right\|_2^2}\right) . $$
   {\bf Step  2.} 
  Let $a \in \mathbb{R}^p$ and $b \in \mathbb{R}^q$ be non-random vectors  such that $\|a\|_2, \|b\|_2 \le 1$. By {\bf Step 1}, it follows that 
        $$
            \mathbb{P}\left( \left|a^\top A  Z \, B b \right| \, > t \,  \right) \le 2\, \exp\left(-\frac{ct^2}{\kappa^2 \, \left\|A a \right\|_2^2 \, \left\|B b \right\|_2^2}\right) \le 2   \exp\left(-\frac{ct^2}{\kappa^2 \, \left\|A \right\|^2 \, \left\|B \right\|^2 }\right). 
        $$
{\bf Step 3.}
        Let $\mathcal N_{p}(\epsilon)$ be an $\epsilon$-net of the unit ball in $\mathbb R^{p}$. It follows that 
        for any $a \in \mathbb R^{p}$ with $\|a\|_2=1$, 
        there exists $\widetilde a \in \mathcal N_{p}(\epsilon)  $ such that  $$  \| a- \widetilde a \|_2 \le \epsilon .$$
        Similarly let $\mathcal N_{q}(\epsilon)$ be an $\epsilon$-net of the unit ball in $\mathbb R^{q}$.  
        \\
        \\
        Denote the random quantity $$\psi= \| AZB\| =  \sup_{a\in \mathbb R^p , b\in \mathbb R^q \|a\|_2 =\|b\|_2=1}|  a^\top AZBb |.$$
        For any given $a\in \mathbb R^p$ and $b\in \mathbb R^q$, let $\widetilde a \in \mathcal N_p(1/4)  $ and $\widetilde b\in \mathcal N_q(1/4) $ be such that 
        $$\| a- \widetilde a \|_2 \le 1/4 \quad \text{and} \quad \| b- \widetilde b \|_2 \le 1/4. $$
        Then
\begin{align*}
    | a^\top AZBb | \le | (a- \widetilde a) ^\top   AZBb| + 
    |   \widetilde a  ^\top   AZB (\widetilde b- b) | + | \widetilde a^\top AZB \widetilde b |. 
\end{align*}
        Note that $\| a- \widetilde a \| \le \frac{1}{4}$. So
        $$ | (a- \widetilde a) ^\top   AZBb| =\frac{1}{4} | \{ 4 (a- \widetilde a) ^\top \} AZBb | \le \frac{\psi}{4}. $$
        Similarly
        $$|   \widetilde a  ^\top   AZB (\widetilde b -b)|   \le \frac{\psi}{4}  .$$
        In addition,
        $$ | \widetilde a^\top AZB \widetilde b | \le \sup_{a \in \mathcal  N_p(1/4) , b\in \mathcal  N_q(1/4)}  |a^\top AZBb| .$$
      So for any $a$ and  $b$,
\begin{align*}
    | a^\top AZBb | \le  \frac{1}{2}\psi + \sup_{a \in N_p(1/4) , b\in \mathcal N_q(1/4)}  |a^\top AZBb|. 
\end{align*}
      Taking sup over all unit vectors $a\in \mathbb R^p$ and $b\in \mathbb R^q$, it follows that 
      $$ \psi \le \frac{1}{2}\psi + \sup_{a \in  \mathcal N_p(1/4) , b\in \mathcal N_q(1/4)}  |a^\top AZBb|,$$
        or simply
        $$ \psi  \le 2\sup_{a \in \mathcal N_p(1/4) , b\in \mathcal N_q(1/4)}  |a^\top AZBb| . $$
        {\bf Step 4.}  By \cite{vershynin2018high},
        the cardinality of  $\mathcal N_p(\epsilon)  $ is bounded by $(\frac{C}{\epsilon})^p $, and the cardinality of  $\mathcal N_q(\epsilon)  $ is bounded by $(\frac{C}{\epsilon})^q $. Therefore 
        \begin{align*}
           \mathbb P  ( \psi \ge 2t) =  \mathbb P\bigg(\sup_{a \in \mathcal N_p(1/4) , b\in \mathcal N_q(1/4)}  |a^\top AZBb|  \ge t  \bigg)  \le &
        {C_2}^p{C_2}^q \sup_{a \in \mathcal N_p(1/4) , b\in \mathcal N_q(1/4)} \mathbb P( |a^\top AZBb|  \ge t ) \\
        \le & 2  {C_2}^p{C_2}^q \exp\left(-\frac{ct^2}{\kappa^2 \, \left\|A \right\|^2 \, \left\|B \right\|^2 }\right),
        \end{align*}
        where $c,C$ are positive constants. The desired result follows  by noting
        $\psi= \| AZB\| $.
\end{proof}

\begin{lemma} \label{lemma:covariance operator bound}
    Suppose $Z \in \mathbb{R}^{n \times m}$, with $Z_{ij} \overset{i.i.d}{\sim} \mathrm{subGaussian}(0,\kappa^2)$. Let   $A \in \mathbb{R}^{m\times p} $ and $B \in \mathbb{R}^{m \times q}$  be non-random   matrices. Then for any $t >0$
    \begin{align}
        \label{eqref: SG bb}\mathbb{P}\left( \left\|A^\top     Z^\top Z B  -  n \kappa^2 A^\top B    \right\|   >   t    \right) \le C_1 \, \exp\left( C  (p+q)   -  \min \left( \frac{t^2}{n\,\kappa^4\, \left\|B \right\|^2 \, \left\|A \right\|^2  },\, \frac{t}{\kappa^2 \left\|B \right\| \, \left\|A \right\|}  \right)   \right),
    \end{align}
    where $C$ and $C_1$ are positive constants.
\end{lemma} 

\begin{proof}
 For any non-random $u,v\in \mathbb R^{m}$ , 
it follows that 
\begin{align*}
    u ^\top Z^\top Z v-  n \kappa^2 u^\top v   = \sum_{j=1}^n  (u^\top Z_j )(v^\top Z_j ) -\mathbb E \{  (u^\top Z_j )(v^\top Z_j )\},
\end{align*}
where $Z_j$ is the $j$-th row of $Z$. Note that $(u^\top Z_j )$ is sub-Gaussian with parameter $\kappa^2\|u\|_2^2 $, and  $(v^\top Z_j )$ is sub-Gaussian with parameter $\kappa^2\|v\|_2^2 $. Since $Z$ have i.i.d. entries, it follows that  $\{ (u^\top Z_j )(v^\top Z_j )\}_{j=1}^n$ are i.i.d. sub-exponential with parameter $ \kappa^4 \|u\|_2^2 \|v\|_2^2$. So
$$\mathbb P \bigg( |u ^\top Z^\top Z v-  n \kappa^2 u^\top v   | \ge t \bigg)\le 2 \exp\bigg (-c\min \bigg\{ \frac{t^2}{ n \kappa^4 \|u\|_2^2 \|v\|_2^2 } , \frac{t}{ \kappa^2 \|u\|_2 \|v\|_2}\bigg\}  \bigg) . $$
\
\\
{\bf Step 1.}
        Let $\mathcal N_{p}(\epsilon)$ be the $\epsilon$-net of the unit ball in $\mathbb R^{p}$. It follows that 
        for any $a \in \mathbb R^{p}$ with $\|a\|_2=1$, 
        there exists $\widetilde a \in \mathcal N_{p}(\epsilon)  $ such that  $$  \| a- \widetilde a \|_2 \le \epsilon .$$
        Similarly let $\mathcal N_{q}(\epsilon)$ be the $\epsilon$-net of the unit ball in $\mathbb R^{q}$.  
        \\
        \\
        Denote the random quantity $$\psi= \| A^\top     Z^\top Z B  -  n \kappa^2 A^\top B   \| =  \sup_{a\in \mathbb R^p , b\in \mathbb R^q \|a\|_2 =\|b\|_2=1}\bigl| a^\top (A^\top Z^\top Z B - n\kappa^2 A^\top B ) b \bigr|.$$
      \
      \\
        For any given $a\in \mathbb R^p$ and $b\in \mathbb R^q$, let $\widetilde a \in \mathcal N_p(1/4)  $ and $\widetilde b\in \mathcal N_q(1/4) $ be such that 
        $$\| a- \widetilde a \|_2 \le 1/4 \quad \text{and} \quad \| b- \widetilde b \|_2 \le 1/4. $$
        Then
\begin{align*}
    | a^\top  (A^\top     Z^\top Z B  -  n \kappa^2 A^\top B ) b | \le | (a- \widetilde a) ^\top  (A^\top     Z^\top Z B  -  n \kappa^2 A^\top B )b|  
    \\+
    |   \widetilde a  ^\top  (A^\top     Z^\top Z B  -  n \kappa^2 A^\top B )  (\widetilde b- b) | + | \widetilde a^\top (A^\top     Z^\top Z B  -  n\kappa^2  A^\top B ) \widetilde b |. 
\end{align*}
        Note that $\| a- \widetilde a \|_2 \le \frac{1}{4}$. 
        So
        $$ | (a- \widetilde a) ^\top  (   A^\top     Z^\top Z B  -  n \kappa^2 A^\top B )  b| =\frac{1}{4} | \{ 4 (a- \widetilde a) ^\top \} (   A^\top     Z^\top Z B  -  n \kappa^2 A^\top B )  b | \le \frac{\psi}{4}. $$
        Similarly
        $$|   \widetilde a  ^\top   (   A^\top     Z^\top Z B  -  n \kappa^2 A^\top B )   (\widetilde b -b)|   \le \frac{\psi}{4}  .$$
        In addition,
        $$ | \widetilde a^\top (   A^\top     Z^\top Z B  -  n \kappa^2 A^\top B )   \widetilde b | \le \sup_{a \in \mathcal  N_p(1/4) , b\in \mathcal  N_q(1/4)}  |a^\top (   A^\top     Z^\top Z B  -  n\kappa^2  A^\top B )  b| .$$
      So for any $a$ and  $b$,
\begin{align*}
    | a^\top (   A^\top     Z^\top Z B  -  n \kappa^2 A^\top B )  b | \le  \frac{1}{2}\psi + \sup_{a \in N_p(1/4) , b\in \mathcal N_q(1/4)}  |a^\top  (   A^\top     Z^\top Z B  -  n \kappa^2 A^\top B )  b|. 
\end{align*}
      Taking sup over all unit vectors $a\in \mathbb R^p$ and $b\in \mathbb R^q$, it follows that 
      $$ \psi \le \frac{1}{2}\psi + \sup_{a \in  \mathcal N_p(1/4) , b\in \mathcal N_q(1/4)}  |a^\top (   A^\top     Z^\top Z B  -  n \kappa^2 A^\top B )  b|,$$
        or simply
        $$ \psi  \le 2\sup_{a \in \mathcal N_p(1/4) , b\in \mathcal N_q(1/4)}  |a^\top (   A^\top     Z^\top Z B  -  n \kappa^2 A^\top B ) b| . $$    
        {\bf Step 2.}  By \cite{vershynin2018high},
        the cardinality of  $\mathcal N_p(\epsilon)  $ is bounded by $(\frac{C}{\epsilon})^p $, and the cardinality of  $\mathcal N_q(\epsilon)  $ is bounded by $(\frac{C}{\epsilon})^q $, for a positive constant $C$. Therefore 
        \begin{align*}
           \mathbb P  ( \psi \ge 2t) =  &\mathbb P\bigg(\sup_{a \in \mathcal N_p(1/4) , b\in \mathcal N_q(1/4)}  |a^\top(   A^\top     Z^\top Z B  -  n \kappa^2 A^\top B )   b|  \ge t  \bigg) 
           \\
           \le &
        {C_2}^p{C_2}^q \sup_{a \in \mathcal N_p(1/4) , b\in \mathcal N_q(1/4)} \mathbb P( |a^\top (   A^\top     Z^\top Z B  -  n \kappa^2 A^\top B ) b|  \ge t )  
        \\
         = &  {C_2}^p{C_2}^q \sup_{a \in \mathcal N_p(1/4) , b\in \mathcal N_q(1/4)} \mathbb P( |(Aa)^\top     Z^\top Z  (B   b)   -  n \kappa^2  (Aa)^\top  (B   b)|  \ge t )  
         \\
         \le & 2 {C_2}^p{C_2}^q  \exp\bigg (-c\min \bigg\{ \frac{t^2}{ n \kappa^4 \|Aa\|_2^2 \|Bb\|_2^2 } , \frac{t}{ \kappa^2 \|Aa\|_2 \|Bb\|_2}\bigg\}  \bigg),
        \end{align*}
        where $C_2$ is a positive constant.
        The desired result follows from the observation that 
        $  \| Aa\|_2 \le \|A\|\|a\|_2 \le \|A\| $, and 
       $  \| Bb\|_2 \le \|B\|\|b\|_2 \le \|B\|$.
\end{proof}

\begin{lemma}\label{lemma:subGaussian matrix}
Suppose $Z \in \mathbb R^{m\times n}$ is a sub-Gaussian random matrix in the sense that for any $u \in \mathbb R^{m},v \in \mathbb R^{n}$, it holds that 
    $$  \| u^\top Z v \|_{\psi_2} \leq \kappa \|u\|_2\|v\|_2 .$$   
    Then with probability at least $ 1-\exp(-c(m+n))$, it holds that 
   \begin{align*}
         \mathbb{P}\left( \left\|  Z  \right\|   >   t    \right) \le C_1 \, \exp\left( C  (m+n)   -\frac{c\,t^2}{\kappa^2\,  } \right),
    \end{align*}
  where  $c, C$ and $C_1$ are positive constants.
\end{lemma}
 \begin{proof}By assumption,  $$\mathbb{P}\left( \left|u^\top \, Z \, v \right| \, > t \,  \right) \le 2\, \exp\left(-\frac{c\,t^2}{\kappa^2 \, \left\|u \right\|_2^2 \, \left\|v \right\|_2^2}\right) . $$     
The rest of the proof is similar and simpler than \Cref{lemma:SG matrix operator} and is omitted. 
  
 \end{proof}

\subsection{Matrix perturbation bounds}

\begin{lemma}\label{eq:regression operator}
   Suppose that  $A \in \mathbb{R}^{n \times r}$. Then
   $$\| A(A^{\top}A)^{-1}  \| \le \sigma_{r}^{-1} (A). $$

\end{lemma}
\begin{proof}
    If $\sigma_{r}  (A) =0$, then the desired result trivially follows. So suppose $\rank(A)=r$. Therefore $ A^\top A $ is invertible. 
  Let   the SVD of $A$ satisfies  $A = U_A\Sigma_AV_A^{\top}$, then
\[
\left\| A(A^{\top}A)^{-1} \right\| = \left\| U_A\Sigma_AV_A^{\top}(V_A\Sigma_A^2V_A^{\top})^{-1} \right\| = \left\| U_A\Sigma_A^{-1}V_A^{\top} \right\| = \sigma_{\min}^{-1}(A) = \sigma_{r}^{-1}(A) .
\]
\end{proof}

\begin{lemma} \label{lemma:singular values of matrix product}
    Suppose $A \in \mathbb R^{m\times n}$ and $B\in \mathbb R^{ n\times k}$ are any two matrices. Then 
    $$ \sigma_j(AB) \le \sigma_j(A)\sigma_{\max} (B ).$$
    
\end{lemma}
\begin{proof}
Let $\lambda_j(M)$ denote the $j-th$ eigenvalues of $M$ in the absolute value order. Then 
    \begin{align}\label{eq:singular values of matrix product term 1} \sigma_j(AB) = \sqrt { \lambda_j (A  B B ^ \top  A^ \top)}  \quad \text{and} \quad \sigma_j(A) = \sqrt { \lambda_j (A    A^ \top)} .
    \end{align}
     Since    $ BB^\top \preceq \sigma^2_{\max} (B ) I_n   $, it follows that 
    $$ A BB^\top A^\top  \preceq  A  (  \sigma^2_{\max} (B )  I_n) A^\top = \sigma^2_{\max } (B) AA^\top .$$
    By the monotonicity of eigenvalues under the  positive definite matrices, it follows that 
    \begin{align}\label{eq:singular values of matrix product term 2}  \lambda_j ( A BB^\top A^\top  ) \le \sigma_{\max}^2(B) \lambda_j ( A A^\top  ). \end{align}
    The desired result follows from \eqref{eq:singular values of matrix product term 1}.
\end{proof}
\begin{lemma} \label{lemma:singular values of matrix product lower bound}
    Suppose $A \in \mathbb R^{m\times n}$ and $B\in \mathbb R^{ n\times n}$ are any two matrices. Then 
    $$ \sigma_j(AB) \ge \sigma_j(A)\sigma_{\min} (B ).$$
    
\end{lemma}
\begin{proof}
    Suppose $\sigma_{\min} (B ) =0$. Then the desired result immediately follows. Therefore it suffices to assume 
$\sigma_{\min} (B ) >0$ and $B $ is invertible.
It suffices to observe that 
\begin{align*}
    \sigma_j(A ) = \sigma_j(ABB^{-1}) \le  \sigma_j(AB)\sigma_{\max }(B^{-1}) = \sigma_j(AB)\sigma^{-1}_{\min }(B),
\end{align*}
where the inequality follows from \Cref{lemma:singular values of matrix product}.

\end{proof} 
\begin{lemma}\label{lemma:lb_forbuious}
    For any   real matrices $A \in \mathbb R^{n\times m }$ and $B\in \mathbb R^{m \times m } $, it holds that 
    \[
    \|AB\|_{\mathrm{F}} \geq \sigma_{\min}(B) \|A\|_{\mathrm{F}} .
    \]
\end{lemma}
\begin{proof}[Proof of \Cref{lemma:lb_forbuious}]
Since $B $ is a square matrix, it follows that 
$$ \lambda_{\min }(BB^\top ) = \sigma_{\min}^2(B),$$
where $\lambda_{\min}(\cdot)$ denotes the minimum eigenvalue.
Note that 
$$ BB^\top  \succeq \lambda _{\min}(BB^\top ) I_m .$$
Therefore 
$$ABB^{\top}A^{\top} \succeq  A \{ \lambda _{\min}(BB^\top ) I_m\}  A ^{\top}, $$
and so 
$$\tr(ABB^{\top}A^{\top})     \geq \tr(A  \{ \lambda _{\min}(BB^\top ) I_m\}  A^{\top}) .$$
Then 
    \begin{align*}
        \|AB\|_{\mathrm{F}}^2 = \tr(ABB^{\top}A^{\top})     \geq \tr(A \{ \lambda _{\min}(BB^\top ) I_m\}  A^{\top})  = \lambda _{\min}(BB^\top ) \tr(A   A^{\top})    =   \sigma_{\min}^2(B) \|A\|^2_{\mathrm{F}}  .
    \end{align*}

\end{proof}

\begin{lemma} \label{lemma:Fobenius norm preserve}
    Let $A\in \mathbb R^{p\times q}$ and $U \in \mathbb O^{q\times r}$. Then 
    $$ \|AUU^\top  \|_{\mathrm{F}} = \|AU\|_{\mathrm{F}}. $$
\end{lemma}
\begin{proof}Observe that 
    \begin{align*}
        \|AUU^\top  \|_{\mathrm{F}} ^2 = \tr ( AUU^\top UU^\top A^\top ) = \tr ( AU U^\top A^\top ) = \|AU\|_{\mathrm{F}}^2.
    \end{align*}
\end{proof}

\begin{lemma}\label{lemma:perterbation_approx}
    Suppose $B, Z \in \mathbb{R}^{n \times m}$. For all $1 \leq R \leq \min\{n,m\}$, write the full SVD of $A$ as
\[
A = B + Z = \widehat{U} \widehat{\Sigma} \widehat{V}^\top = \begin{bmatrix} \widehat{U}_{(R)} \ \widehat{U}_{\perp} \end{bmatrix} \cdot \begin{bmatrix} \widehat{\Sigma}_{(R)} & \\ & \widehat{\Sigma}_{\perp} \end{bmatrix} \cdot \begin{bmatrix} \widehat{V}_{(R)}^\top \\ \widehat{V}_{\perp}^\top \end{bmatrix},
\]
where $\widehat{U}_{(R)} \in \mathbb{O}_{n, R}$, $\widehat{V}_{(R)} \in \mathbb{O}_{m, R}$ correspond to the leading $R$ left and right singular vectors; and $\widehat{U}_{\perp} \in \mathbb{O}_{n, n - R}$, $\widehat{V}_{\perp} \in \mathbb{O}_{m, m - R}$ correspond to their orthonormal complement. We have
    \begin{align*}
    \left\|\mathcal P_{\widehat{U}_{\perp}}B\right\|_{\mathrm{F}} \leq& 3\sqrt{\sum_{j = R+1}^{\min\{n,m\}}\sigma_j^2(B)} + 2\min\left\{\sqrt{R}\|Z\|, \|Z\|_{\mathrm{F}}\right\}\\
    =& 3\left\|B_{(R)} - B\right\|_{\mathrm{F}} + 2\min\left\{\sqrt{R}\|Z\|, \|Z\|_{\mathrm{F}}\right\},
    \end{align*}
    where $B_{(R)}$ denote the rank-$R$ truncated SVD of $B$, this is, if $B=U\Sigma V^\top$ then 
$B_{(R)} =U_{(R)}\Sigma_{(R)}V_{(R)}^\top$. 
\end{lemma}
\begin{proof}
    Without loss of generality, assume $n \leq m$.
    For $A \in \mathbb{R}^{n\times m}$, let $\Sigma(A) \in \mathbb{R}^{n\times m}$ denote the non-negative diagonal matrices whose diagonal entries are the non-increasingly ordered singular values of $A$. For all $1 \leq R \leq n$, let $B_{(R)}$ denote the truncated SVD of $B$ with rank $R$, and we have by the Eckart–Young–Mirsky theorem
    \[
    \left\|B_{(R)} - B\right\|_{\mathrm{F}} = \sqrt{\sum_{j = R+1}^{n}\sigma_{j}^2(B)}.
    \]
    For a matrix $A \in \mathbb{R}^{m \times n}$, let $\Sigma(A) \in \mathbb{R}^{m \times n}$ be a non-negative (rectangular) diagonal matrix whose diagonal entries are the non-increasingly ordered singular values of $A$.
    
    We have that
        \begin{align*}
        &\left\|\mathcal P_{\widehat{U}_{\perp}}B\right\|_{\mathrm{F}} \leq \left\|\mathcal P_{\widehat{U}_{\perp}}B_{(R)}\right\|_{\mathrm{F}} + \left\|\mathcal P_{\widehat{U}_{\perp}}(B-B_{(R)})\right\|_{\mathrm{F}} = \sqrt{\sum_{j = 1}^R\sigma_j^2(\mathcal P_{\widehat{U}_{\perp}}B_{(R)})} + \left\|\mathcal P_{\widehat{U}_{\perp}}(B-B_{(R)})\right\|_{\mathrm{F}}\\
        \leq& \sqrt{\sum_{j = 1}^R\sigma_j^2(\mathcal P_{\widehat{U}_{\perp}}B_{(R)})} + \left\|B-B_{(R)}\right\|_{\mathrm{F}} = \sqrt{\sum_{j = 1}^R\sigma_j^2(\mathcal P_{\widehat{U}_{\perp}}B_{(R)})} + \sqrt{\sum_{j = R+1}^{n}\sigma_{j}^2(B)}\\
        \leq& \left\|(\sigma_{1}(\mathcal P_{\widehat{U}_{\perp}}B_{(R)}) - \sigma_{1}(\mathcal P_{\widehat{U}_{\perp}}B), \dots, \sigma_{R}(\mathcal P_{\widehat{U}_{\perp}}B_{(R)}) - \sigma_{R}(\mathcal P_{\widehat{U}_{\perp}}B))^{\top}\right\|_2 + \left\|(\sigma_{1}(\mathcal P_{\widehat{U}_{\perp}}B), \dots, \sigma_{R}(\mathcal P_{\widehat{U}_{\perp}}B))^{\top}\right\|_2\\
        &+ \sqrt{\sum_{j = R+1}^{n}\sigma_{j}^2(B)}\\
        \leq& \left\|\Sigma(\mathcal P_{\widehat{U}_{\perp}}B_{(R)}) - \Sigma(\mathcal P_{\widehat{U}_{\perp}}B)\right\|_{\mathrm{F}} + \left\|(\sigma_{1}(\mathcal P_{\widehat{U}_{\perp}}B), \dots, \sigma_{R}(\mathcal P_{\widehat{U}_{\perp}}B))^{\top}\right\|_2 + \sqrt{\sum_{j = R+1}^{n}\sigma_{j}^2(B)}\\
        \leq& \left\|\mathcal P_{\widehat{U}_{\perp}}(B_{(R)} - B)\right\|_{\mathrm{F}} + \sqrt{\sum_{j = 1}^R\sigma_{j}^2(\mathcal P_{\widehat{U}_{\perp}}B)} + \sqrt{\sum_{j = R+1}^{n}\sigma_{j}^2(B)}\\
        \leq& \sqrt{\sum_{j = 1}^R\sigma_{j}^2(\mathcal P_{\widehat{U}_{\perp}}B)} + 2\sqrt{\sum_{j = R+1}^{n}\sigma_{j}^2(B)},
    \end{align*}
    where the first equality follows from $\rank(B_{(R)}) = R$, and the fifth inequality follows from \Cref{thm:Mirsky}.
    To upper bound $\sqrt{\sum_{j = 1}^{R}\sigma_j^2(\mathcal P_{\widehat{U}_{\perp}}B)}$, we first consider $\sqrt{\sum_{j = 1}^{R}\sigma_j^2(\mathcal P_{\widehat{U}_{\perp}}A)}$.
    Note that
    \[
        \mathcal P_{\widehat{U}_{\perp}}A = \sum_{j = R+1}^{n}\sigma_j(A)\widehat u_j \widehat v_j^{\top},
    \]
    where $\widehat u_j$ and $\widehat v_j$ are the left and right singular vectors associated with the $j$th largest singular value $\sigma_j(A)$. Note that $\sigma_j(A) = \sigma_j(B) = 0$ for $j > n$.
    It follows that
    \begin{align}\label{eq:proof_singularvalue_perp}
        &\sqrt{\sum_{j = 1}^{R}\sigma_j^2(\mathcal P_{\widehat{U}_{\perp}}A)} = \sqrt{\sum_{j = R+1}^{2R}\sigma_j^2(A)} = \left\|(\sigma_{R+1}(A), \dots, \sigma_{2R}(A))^{\top}\right\|\nonumber\\
        \leq& \left\|(\sigma_{R+1}(A) - \sigma_{R+1}(B), \dots, \sigma_{2R}(A) - \sigma_{2R}(B))^{\top}\right\| + \left\|(\sigma_{R+1}(B), \dots, \sigma_{2R}(B))^{\top}\right\|\nonumber\\
        \leq& \min\left\{\sqrt{R}\|Z\|, \|Z\|_{\mathrm{F}}\right\} + \sqrt{\sum_{j = R+1}^{n}\sigma_j^2(B)},
    \end{align}
    where the first inequality follows from the triangle inequality, and second inequality follows from Weyl's inequality \citep{weyl1912asymptotische}, i.e.~$|\sigma_j(A) - \sigma_j(B)| \leq \|A - B\|$ for all $1 \leq j \leq n$, as well as the fact that
    \[
    \left\|(\sigma_{R+1}(A) - \sigma_{R+1}(B), \dots, \sigma_{2R}(A) - \sigma_{2R}(B))^{\top}\right\| \leq \left\|\Sigma(A) - \Sigma(B)\right\|_{\mathrm{F}} \leq \left\|Z\right\|_{\mathrm{F}},
    \]
    where the last inequality follows from \Cref{thm:Mirsky}.
    It then follows from \eqref{eq:proof_singularvalue_perp},
    \begin{align*}
        &\sqrt{\sum_{j = 1}^{R}\sigma_j^2(\mathcal P_{\widehat{U}_{\perp}}B)} = \left\|(\sigma_{1}(\mathcal P_{\widehat{U}_{\perp}}(A-Z)), \dots, \sigma_{R}(\mathcal P_{\widehat{U}_{\perp}}(A-Z))^{\top}\right\|\\
        \leq& \left\|(\sigma_{1}(\mathcal P_{\widehat{U}_\perp}(A-Z)) - \sigma_{1}(\mathcal P_{\widehat{U}_\perp}A), \dots, \sigma_{R}(\mathcal P_{\widehat{U}_{\perp}}(A-Z)) - \sigma_{R}(\mathcal P_{\widehat{U}_{\perp}}A))^{\top}\right\|\\& + \left\|(\sigma_{1}(\mathcal P_{\widehat{U}_{\perp}}A), \dots, \sigma_{R}(\mathcal P_{\widehat{U}_\perp}A))^{\top}\right\|\\ \leq& \min\left\{\sqrt{R}\|\mathcal P_{\widehat{U}_\perp}Z\|, \|\mathcal P_{\widehat{U}_\perp}Z\|_{\mathrm{F}} \right\} + \sqrt{\sum_{j = 1}^{R}\sigma_j^2(\mathcal P_{\widehat{U}_\perp}A)}\\
         \leq& \min\left\{\sqrt{R}\|Z\|, \|Z\|_{\mathrm{F}}\right\} + \sqrt{\sum_{j = 1}^{R}\sigma_j^2(P_{\widehat{U}_\perp}A)}\\
          \leq& 2\min\left\{\sqrt{R}\|Z\|, \|Z\|_{\mathrm{F}}\right\} + \sqrt{\sum_{j = R+1}^{n}\sigma_j^2(B)},
    \end{align*}
    where the first two inequalities follow from the same arguments as in \eqref{eq:proof_singularvalue_perp}.
    Consequently,
    \[
    \left\|\mathcal P_{\widehat{U}_\perp}B\right\|_{\mathrm{F}} \leq 3\sqrt{\sum_{j = R+1}^{n}\sigma_j^2(B)} + 2\min\left\{\sqrt{R}\|Z\|, \|Z\|_{\mathrm{F}}\right\}.
    \]
\end{proof}

\begin{lemma}[Proposition 1 of \cite{cai2018rate}]\label{prop:sintheta}
    Suppose $Y \in \mathbb{R}^{m \times n}$, $\widehat V = [\widehat V_{r} \  \widehat V_{\perp}] \in \mathbb{O}_{n}$  where  $\widehat V_{r} \in \mathbb{O}_{n, r}$, $\widehat V_{\perp} \in \mathbb{O}_{n, n - r}$ correspond to the first $r$ and last $(n - r)$ right singular vectors of $Y$ respectively.  Let $ V  = [ V_{r} \  V_{\perp}] \in \mathbb{O}_{n,n}$ be any orthogonal matrix with $ V _{r} \in \mathbb{O}_{n, r}$, $ V_{\perp} \in \mathbb{O}_{n, n - r}$. Given that $\sigma_R(Y V_{r}) > \sigma_{r+1}(Y)$, we have
\begin{align}\label{eq:unblanced svd lemma}
\| \sin \Theta(V_{r}, \widehat{V}_{r}) \| \leq \frac{\sigma_r(Y V _{r}) \| \mathcal P_{Y V_{r}} Y V_{\perp} \|}{\sigma_r^2(Y V _{r}) - \sigma_{r+1}^2(Y)} \wedge 1,
\end{align}
where $\mathcal P_A$ is the projection operator onto the column space of $A$.
\end{lemma}

\begin{lemma}[Properties of the sin$\Theta$ distances ]\label{lemma:sintheta}
\
\\
The following properties hold for the $\sin\Theta$ distances.
\begin{enumerate}
    \item (Equivalent Expressions) Suppose $V, \widehat{V} \in \mathbb{O}_{p, R}$. If $V_{\perp}$ is an orthogonal extension of $V$, namely $\begin{bmatrix} V & V_{\perp}\end{bmatrix} \in \mathbb{O}_{p}$, we have the following equivalent forms for $\| \sin \Theta(\widehat{V}, V) \|$ and $\| \sin \Theta(\widehat{V}, V) \|_{\mathrm{F}}$,
    \[
    \| \sin \Theta(\widehat{V}, V) \| = \sqrt{1 - \sigma_{\min}^2(\widehat{V}^T V)} = \| \widehat{V}^T V_{\perp} \|,
    \]
    \[
    \| \sin \Theta(\widehat{V}, V) \|_{\mathrm{F}} = \sqrt{r - \| V^T \widehat{V} \|_{\mathrm{F}}^2} = \| \widehat{V}^T V_{\perp} \|_{\mathrm{F}}.
    \]
    \item (Triangle Inequality) For all $V_1, V_2, V_3 \in \mathbb{O}_{p, R}$,
    \[
    \| \sin \Theta(V_2, V_3) \| \leq \| \sin \Theta(V_1, V_2) \| + \| \sin \Theta(V_1, V_3) \|,
    \]
    \[
    \| \sin \Theta(V_2, V_3) \|_{\mathrm{F}} \leq \| \sin \Theta(V_1, V_2) \|_{\mathrm{F}} + \| \sin \Theta(V_1, V_3) \|_{\mathrm{F}}.
    \]
    \item (Equivalence with Other Metrics)
\[
\| \sin \Theta(\widehat{V}, V) \| \leq \sqrt{2} \| \sin \Theta(\widehat{V}, V) \|,
\]
\[
\| \sin \Theta(\widehat{V}, V) \|_{\mathrm{F}} \leq \sqrt{2} \| \sin \Theta(\widehat{V}, V) \|_{\mathrm{F}},
\]
\[
\| \sin \Theta(\widehat{V}, V) \| \leq \| \widehat{V} \widehat{V}^\top - V V^\top \| \leq 2 \| \sin \Theta(\widehat{V}, V) \|,
\]
\[
\| \widehat{V} \widehat{V}^\top - V V^\top \|_{\mathrm{F}} = \sqrt{2} \| \sin \Theta(\widehat{V}, V) \|_{\mathrm{F}}.
\]

\end{enumerate}
    
\end{lemma}

\begin{theorem}[Mirsky's singular value inequality in \cite{mirsky1960symmetric}]\label{thm:Mirsky}
For any matrices $A, B \in \mathbb{R}^{m\times n}$, let $A = V_1\Sigma(A)W_1^\top$ and $B = V_2\Sigma(B)W_2^\top$ be the full SVDs of $A$ and $B$, respectively. Note that $\Sigma(A),\Sigma(B) \in \mathbb{R}^{m \times n}$ are non-negative (rectangular) diagonal matrices whose diagonal entries are the non-increasingly ordered singular values of $A$ and $B$, respectively. Then
\begin{equation}
    \|\Sigma(A) - \Sigma(B)\| \leq \|A - B\|
\end{equation}
for any unitarily invariant norm $\|\cdot\|$ on $\mathbb{R}^{m\times n}$. 
\end{theorem}

\begin{theorem}[Weyl's inequality for singular values]
\label{thm:WeylSV}
Let $A,B\in\mathbb{R}^{m\times n}$ and denote their singular values (in nonincreasing order) by $\{\sigma_i(A)\}$ and $\{\sigma_i(B)\}$  respectively. In addition denote the singular values of $A+B$ as $\{\sigma_i(A+B)\}$. Then for all indices $i,j$ satisfying $i+j-1\le \min\{m,n\}$,
\[
\sigma_{i+j-1}(A+B) \le \sigma_i(A) + \sigma_j(B).
\]
\end{theorem}

Let $\mathcal T _{(r_1,r_2,r_3)}$  denote the class of tensor in $\mathbb R^{p_1\times p_2\times p_3}$ with tucker ranks at most $(r_1, r_2,r_3)$. More precisely
$$ \mathcal T _{(r_1,r_2,r_3)} = \{A \in \mathbb R^{p_1\times p_2\times p_3} : \rank ( \mathcal M_k (A)) \le r_k , k =1,2,3 \}.$$
\begin{lemma}   
 \label{lemma:project_forbuious}
Let $X^* \in \mathbb{R}^{p_1 \times p_2 \times p_3}$. For  $k  \in  \{ 1, 2, 3\}$,  suppose the $k$-th matricization of $X^*$ satisfies 
\[
\mathcal{M}_k(X^*) = \begin{bmatrix}
    U_k^* & U_{k\,\perp}^*
\end{bmatrix}
\begin{bmatrix}
    \Sigma_k^* & 0\\
    0 & \Sigma_{k\,\perp}^*
\end{bmatrix}
\begin{bmatrix}
    V_k^* & V_{k\,\perp}^*
\end{bmatrix}^{\top}
\]
 where $U_k^* \in \mathbb{O}_{p_k,r_k}$ corresponds to the    the top $r_k$ singular vectors of $\mathcal{M}_k(X^*)$. Then for  $k  \in  \{ 1, 2, 3\}$, it holds that  
    \begin{align*}
    \left\|X^* \times_k     U_{k\perp}^*  \right\|_{\mathrm{F}} =  \left\|X^* \times_k (I_{p_k} - \mathcal{P}_{U_k^*})\right\|_{\mathrm{F}} =\sqrt{\sum_{j = r_k+1}^{  \rank( \mathcal M_k(X^*) ) }\sigma^2_j(\mathcal{M}_k(X^*))} \le \xi_{(r_1,r_2,r_3)},
    \end{align*}
    where  \[
\xi_{(r_1, r_2, r_3)} = \inf_{ A \in   \mathcal T _{(r_1,r_2,r_3)}}\|A - X^*\|_{\mathrm{F}}.
\]
\end{lemma}
\begin{proof} By symmetry, it suffices to consider $k=1$.
Note that 
$$\left\|X^* \times_k (I_{p_k} - \mathcal{P}_{U_k^*})\right\|_{\mathrm{F}}  =\left\|X^* \times_k   \mathcal{P}_{U_{k\perp^*} }\right\|_{\mathrm{F}}=\left\|X^* \times_k     U_{k\perp}^*  \right\|_{\mathrm{F}}. $$
In addition
\begin{align*}
    \left\|X^* \times_1 (I_{p_1} - \mathcal{P}_{U_1^*})\right\|_{\mathrm{F}} = &\left\|(I_{p_1} - \mathcal{P}_{U_1^*}) \cdot \mathcal{M}_1(X^*) \right\|_{\mathrm{F}} 
     = \left\|U_{1\,\perp}^*U_{1\,\perp}^{*\,\top} \cdot (U_1^*\Sigma^*_1V_1^{*\,\top} + U_{1\,\perp}^*\Sigma^*_{1\,\perp}V_{1\,\perp}^{*\,\top})\right\|_{\mathrm{F}}\\
    =& \left\|U_{1\,\perp}^*U_{1\,\perp}^{*\,\top} \cdot (U_{1\,\perp}^*\Sigma^*_{1\,\perp}V_{1\,\perp}^{*\,\top})\right\|_{\mathrm{F}}
    =\sqrt{\sum_{j = r_1+1}^{  \rank( \mathcal M_1(X^*) ) }\sigma^2_j(\mathcal{M}_1(X^*))}.
    \end{align*}
    Note that by the properties of SVD, for any $W \in \mathbb R^{p_1\times p_2p_3}$ such that $\rank(W)\le r_1$, it holds that 
    \begin{align*}
        \sqrt{\sum_{j = r_1+1}^{  \rank( \mathcal M_1(X^*) ) }\sigma^2_j(\mathcal{M}_1(X^*))} \le \|\mathcal M_1(X^*)  -W  \|_{\mathrm{F}}.
    \end{align*}
    For any $A \in \mathcal T_{(r_1,r_2,r_3)}$, it holds that $\rank(\mathcal M_1(A))\le r_1$. Therefore for any  $A \in \mathcal T_{(r_1,r_2,r_3)}$, 
    \begin{align*}
        \sqrt{\sum_{j = r_1+1}^{  \rank( \mathcal M_1(X^*) ) }\sigma^2_j(\mathcal{M}_1(X^*))} \le \|\mathcal M_1(X^*)  -A  \|_{\mathrm{F}}.
    \end{align*} 
    Taking the inf  over all $A \in \mathcal T_{(r_1,r_2,r_3)}$, it follows that 
    $$ \sqrt{\sum_{j = r_1+1}^{  \rank( \mathcal M_1(X^*) ) }\sigma^2_j(\mathcal{M}_1(X^*))}  \le \xi_{(r_1,r_2,r_3)}. $$
\end{proof}

\begin{lemma}[Ky Fan-type Inequality for Sums of Matrices]\label{thm: KyFan}
   Let $A,B\in\mathbb{R}^{m\times n}$ and denote their singular values (in nonincreasing order) by $\{\sigma_i(A)\}$ and $\{\sigma_i(B)\}$  respectively. In addition denote the singular values of $A+B$ as $\{\sigma_i(A+B)\}$.  Then for any \( 1\le k  \le  \min\{m,n\} \), it holds that
   \[
\sqrt{\sum_{i=1}^k \sigma_i^2(A + B)} \le \sqrt{\sum_{i=1}^k \sigma_i^2(A)} + \sqrt{\sum_{i=1}^k \sigma_i^2(B)}.
    \] 
\end{lemma}
\begin{proof}
    For a symmetric matrix \( M \in \mathbb{R}^{n \times n} \),  by Ky Fan's maximum principle (see e.g.~II.1.13 in \cite{bhatia2013matrix}), for any \(1 \le  k \le   n \),
    \[
    \sum_{i=1}^k \lambda_i (M) = \sup_{ P\in \mathbb O^{n\times k}  } \tr ( P^\top MP  ).
    \]
    Therefore 
    $$  \sum_{i=1}^k \sigma_i^2(A) =\sum_{i=1}^k \lambda_i (A^\top A ) = \sup_{ P\in \mathbb O^{n\times k}  } \tr ( P^\top A^\top A  P  )  = \sup_{ P\in \mathbb O^{n\times k}  } \|AP \| _{\mathrm{F}} ^2,$$
    and so 
    $$ \sqrt { \sum_{i=1}^k \sigma_i^2(A)  }  = \sup_{ P\in \mathbb O^{n\times k}  } \|AP \| _{\mathrm{F}}  . $$
     Then 
    \begin{align*}
        \sqrt{\sum_{i=1}^k \sigma_i^2 \left(A+B \right)} 
        = & \max_{U \in \mathbb{O}_{n,k}}\|(A+B)U\|_{\mathrm{F}}
         \leq \max_{U \in \mathbb{O}_{n,k}}\|AU\|_{\mathrm{F}} + \max_{U \in \mathbb{O}_{n,k}}\|BU\|_{\mathrm{F}}\\
        &= \sqrt{\sum_{i=1}^k \sigma_i^2 \left(A  \right)} +\sqrt{\sum_{i=1}^k \sigma_i^2 \left( B \right)}  .
    \end{align*}
    
\end{proof} 

\subsection{Minimax lower bound for vectors}

\begin{lemma}\label{lemma:lower bound vector}
Let $\kappa >0$ and $d \in \mathbb Z^+$, and denote   
$$ \mathcal V= \{ V\in \mathbb R^d, \|V\|_{\infty} \le  \kappa \}.$$
 Consider the model  $ Y\sim \mathcal N(V , \kappa^2 I_d) $ where  $V\in \mathcal V $. Then there exists a universal constant $c$ such that 
$$ \inf_{\widehat V} \sup _{  V\in \mathcal V } \mathbb E \| \widehat V -V \| ^2  \ge c \kappa^2 d, $$
where infimum is taken over all the estimators ${\widehat V}$ based on $Y$.
\end{lemma}

\begin{proof} This is a standard minimax lower bound for bounded normal means; see for example \cite{berry1990minimax} and \cite{donoho1990minimax}. 
\end{proof}

\end{document}